\documentclass{article}

\usepackage{microtype}
\usepackage{graphicx}
\usepackage{subcaption}
\usepackage{booktabs} 
\usepackage{multirow}
\usepackage{amssymb}
\usepackage{hyperref}
\usepackage{tabularx}
\usepackage{xcolor}
\usepackage[table]{xcolor}
\usepackage{makecell}
\usepackage{xspace}
\usepackage{xspace}
\usepackage{caption}
\usepackage{pgf}
\usepackage{booktabs}
\usepackage{xfp}
\usepackage{subcaption}

\newcommand{\GetQA}[2]{%
\expandafter\GetValAux#1,\relax,#2%
}
\def\GetValAux#1,#2,#3,#4,#5,#6,#7,#8\relax,#9{%
\ifcase#9\or#1\or#2\or#3\or#4\or#5\or#6\or#7\fi
}

\newcommand{\QAMean}[1]{%
\fpeval{
round(
(\GetQA{#1}{1}+
 \GetQA{#1}{2}+
 \GetQA{#1}{3}+
 \GetQA{#1}{4}+
 \GetQA{#1}{5}+
 \GetQA{#1}{6}+
 \GetQA{#1}{7})/7,
2)
}
}

\newcommand{\GetGen}[2]{%
\expandafter\GetGenAux#1,\relax,#2%
}
\def\GetGenAux#1,#2,#3,#4,#5,#6\relax,#7{%
\ifcase#7\or#1\or#2\or#3\or#4\or#5\fi
}

\newcommand{\GENMean}[1]{%
\fpeval{
round(
(\GetGen{#1}{1}+
 \GetGen{#1}{2}+
 \GetGen{#1}{3}+
 \GetGen{#1}{4}+
 \GetGen{#1}{5})/5,
2)
}
}

\newcommand{\paperN}{POP\xspace}

\definecolor{fst}{rgb}{0.909, 0.504, 0.588}
\definecolor{sed}{rgb}{0.994, 0.806, 0.742}
\definecolor{thd}{rgb}{0.999, 0.966, 0.921}

\newcommand{\fst}{\cellcolor{fst}}
\newcommand{\sed}{\cellcolor{sed}}
\newcommand{\thd}{\cellcolor{thd}}

\usepackage[preprint]{icml2026}

\usepackage{amsmath}
\usepackage{mathtools}
\usepackage{amsthm}

\newcommand{\cmark}{\textcolor{green!70!black}{\checkmark}}
\newcommand{\xmark}{\textcolor{red}{\texttimes}}

\usepackage[capitalize,noabbrev]{cleveref}

\theoremstyle{plain}

\theoremstyle{definition}

\theoremstyle{remark}

\usepackage[textsize=tiny]{todonotes}

\icmltitlerunning{POP: Online Structural Pruning Enables Efficient Inference of Large Foundation Models}

\begin{document}

\twocolumn[
  \icmltitle{POP: Online Structural Pruning Enables \\ Efficient Inference of Large Foundation Models}



  \icmlsetsymbol{equal}{*}

  \begin{icmlauthorlist}
    \icmlauthor{Yi Chen}{KAIST}
    \icmlauthor{Wonjin Shin}{KAIST}
    \icmlauthor{Shuhong Liu}{UT}
    \icmlauthor{Tho Mai}{KAIST}
    \icmlauthor{Jeongmo Lee}{KAIST}\\
    \icmlauthor{Chuanbo Hua}{KAIST}
    \icmlauthor{Kun Wang}{KAIST}
    \icmlauthor{Jun Liu}{IST}
    \icmlauthor{Joo-Young Kim}{KAIST}
  \end{icmlauthorlist}

  \icmlaffiliation{KAIST}{Korea Advanced Institute of Science and Technology
(KAIST), Daejeon, South Korea}
  \icmlaffiliation{UT}{University of Tokyo, Tokyo, Japan}
    \icmlaffiliation{IST}{ Tokyo Institute of Technology, Tokyo, Japan}

  \icmlcorrespondingauthor{Joo-Young Kim}{jooyoung1203@kaist.ac.kr}

  \icmlkeywords{Machine Learning, ICML}

  \vskip 0.3in
]



\printAffiliationsAndNotice{}  

\begin{abstract}
Large foundation models (LFMs) achieve strong performance through scaling, yet current structural pruning methods derive fixed pruning decisions during inference, overlooking sparsity patterns that emerge in the autoregressive token generation.
In this paper, we propose POP (\textbf{P}artition-guided \textbf{O}nline \textbf{P}runing), an efficient online structural pruning framework that enables context-conditioned dynamic pruning with minimal computational overhead. POP partitions model channels into retained, candidate, and pruned regions, where prefilling defines a coarse pruning partition, and the decoding stage generates a fine-grained mask within the candidate region, avoiding full-channel re-evaluation. The coarse pruning partition preserves consistently important weights, while the fine-grained masking provides context-conditioned variation during decoding. Moreover, POP is a lightweight, plug-and-play method that requires no preprocessing, including offline calibration, retraining, or learning predictors. Extensive evaluations across diverse LFMs, including large language models (LLMs), mixture-of-experts models (MoEs), and vision–language models (VLMs), demonstrate that POP consistently delivers higher accuracy than existing pruning approaches while incurring smaller computational overhead and minimizing inference latency.

\end{abstract}
\section{Introduction}
\label{sec:Introduction}

\begin{figure}[t]
    \centering
    \includegraphics[width=\linewidth]{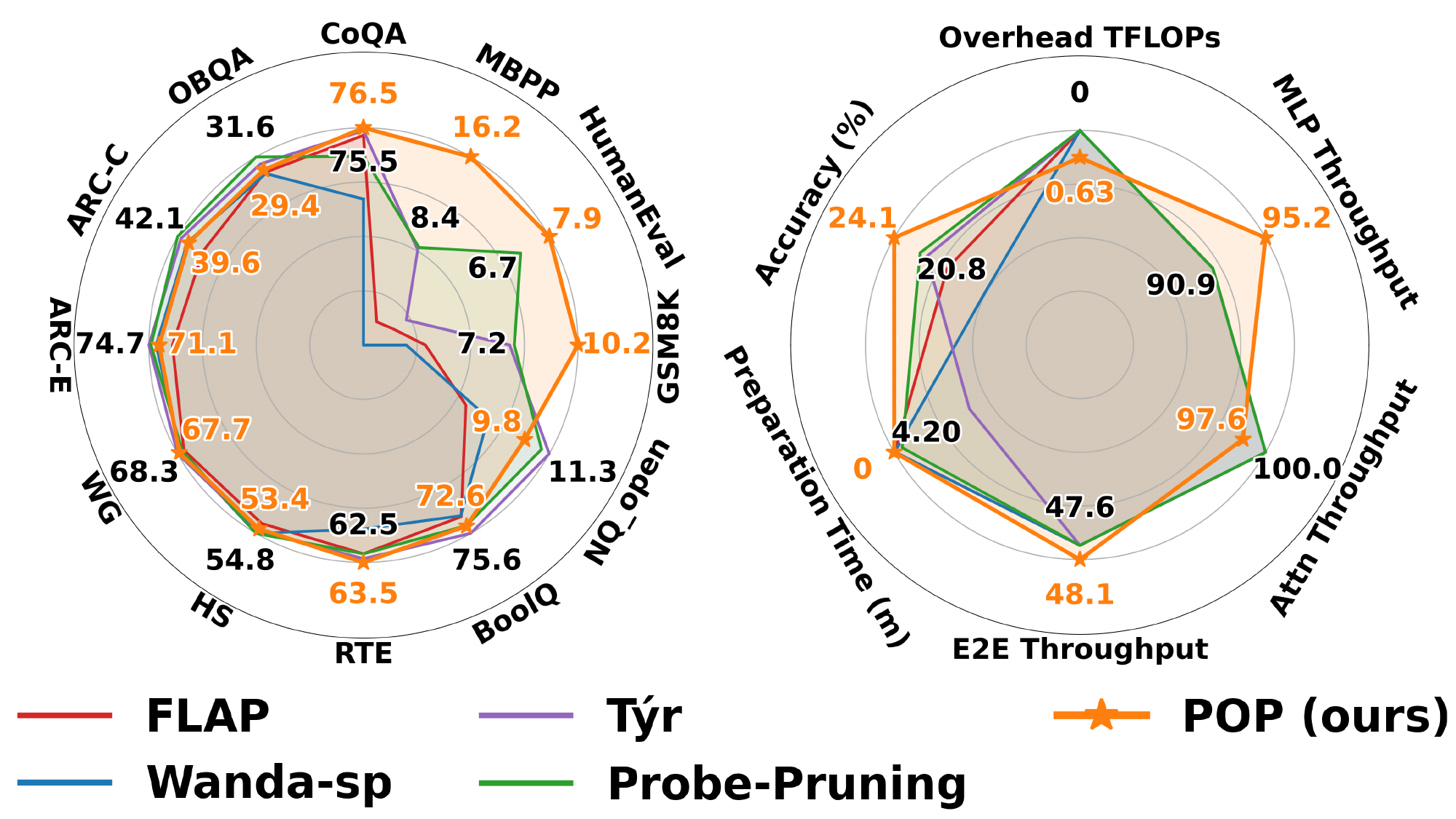}
    \caption{
Task-wise normalized performance across benchmarks on Llama2-7B, together with inference efficiency.
\textsc{POP} exhibits more consistent performance across tasks while maintaining higher efficiency than baseline pruning methods.}
    \label{fig:radar_taskwise}
    \vspace{-0.2in}
\end{figure}

Large foundation models (LFMs)~\cite{bommasani2021opportunities}, such as large language models (LLMs)~\cite{brown2020language,chowdhery2023palm}, vision–language models (VLMs)~\cite{radford2021learning,alayrac2022flamingo}, and mixture-of-experts models (MoE)~\cite{du2022glam,fedus2022switch,liu2024deepseek}, have achieved remarkable performance gains through continuous scaling of model size~\cite{kaplan2020scaling}.
However, this scaling trend comes with substantial computational overhead, posing a major challenge for efficient inference.
To mitigate this issue, various model compression techniques~\cite{han2015deep}, such as quantization~\cite{jacob2018quantization,dettmers2022gpt3}, low-rank decomposition~\cite{denton2014exploiting,hu2022lora}, and pruning~\cite{han2015learning,han2015deep,sun2024simple}, have been extensively studied.
Among them, structural pruning~\cite{an2024fluctuation,Li2025TyrThePruner,ma2023llm,qi2025probe} is particularly appealing, as it removes entire computational structures (e.g., weight channels) without relying on specialized hardware support~\cite{wang2020structured,xia2022structured,hu2025fasp}.

Early structural pruning methods for LLMs typically derive a static pruning mask through offline calibration~\cite{ma2023llm,an2024fluctuation,Li2025TyrThePruner} or pretrained predictors~\cite{liu2023deja}, and apply it uniformly throughout inference. Although effective at reducing model size and computational cost, such static strategies are input-agnostic and fail to capture the diverse sparsity patterns induced by different contexts and tasks. Recent studies have therefore explored context-conditioned~\cite{hou2025instruction,qi2025probe} or dynamic pruning~\cite{liu2025rap} approaches that adapt pruning decisions across input prompts, often without retraining. 

Despite this progress, online pruning remains challenging.
Most existing approaches make pruning decisions only once during the prefilling stage and apply the resulting decision throughout autoregressive generation.
Such static pruning strategies fail to adapt to evolving decoding contexts,
which can conflict with the inherently input-dependent computation behavior of LLMs
and result in substantial performance degradation on generation tasks compared to dense models. This issue is reflected in our empirical observations. For example, under a 20\% pruning ratio,
the model pruned by the static framework Tyr~\cite{Li2025TyrThePruner} retains 98\% accuracy on short-form QA tasks such as ARC-C~\cite{allenai:arc}, with LLaMA-2-7B~\cite{touvron2023llama}, but preserves only 35\% performance on long-form generation benchmarks such as MBPP~\cite{austin2021program}. These observations motivate the consideration of alternatives to static pruning, raising the following question:
\begin{center}
\emph{Can pruning decisions be conditioned on the generation context during online inference?}
\end{center}
A natural way to address this question is to update pruning decisions dynamically during decoding. However, most existing context-conditioned pruning methods are designed for multi-token settings. For example, Probe Pruning~\cite{qi2025probe} relies on probing-based importance estimation, while Instruction-Following Pruning~\cite{hou2025instruction} employs an input-conditioned predictor; both require aggregated activations across multiple tokens. Such designs are fundamentally incompatible with autoregressive decoding, where only a single token is available at each step. Even if one applies probing-based importance estimation to the generated token itself, it would further introduce an additional forward pass, incurring prohibitive runtime overhead. These limitations highlight the need for an online pruning framework that enables context-conditioned adaptation with minimal overhead and broad applicability across tasks and model architectures.

To narrow the performance gap between pruned and dense models
while maintaining minimal runtime overhead,
we propose \paperN (\textbf{P}artition-guided \textbf{O}nline \textbf{P}runing).
\paperN is an online structural pruning framework
that performs context-conditioned pruning decisions
during autoregressive inference with negligible additional computation.
In particular, \paperN is a lightweight and plug-and-play approach,
requiring no preprocessing such as offline calibration,
retraining, or predictor learning,
and can be readily applied to diverse large foundation models.
The main contributions of this work are summarized as follows:

\begin{itemize}

\item \textbf{We identify the necessity of online pruning during autoregressive generation.}
We show that pruning decisions fixed during prefilling introduce systematic bias in newly generated tokens because they remain determined by the initial prompt, highlighting the need for online, dynamic pruning to address contextual sparsity during decoding.

\item \textbf{We propose POP: a Partition-guided Online Pruning framework that balances accuracy and efficiency.}
\paperN introduces a tri-state channel partition based on prefilling-stage importance, dividing channels into retained, pruned, and candidate regions, where only candidate channels are dynamically selected during decoding, enabling fine-grained context adaptivity with minimal overhead.

\item \textbf{We develop \paperN as a model-agnostic and deployment-friendly framework.}
Without offline calibration, auxiliary predictors, or retraining, \paperN can be seamlessly applied to a wide range of LFMs, enabling practical, scalable inference-time acceleration.

\end{itemize}

We empirically evaluate \textsc{POP} on diverse LFMs, including representative LLMs, MoE models, and VLMs, and observe consistent accuracy gains and lower latency over state-of-the-art structural pruning methods. In particular, \textsc{POP} improves Llama3.1-8B by 5.27 on short-form QA and 17.71 on generation. Moreover, with only 2.85\% FLOPs overhead relative to the dense model, \textsc{POP} achieves a $1.29\times$ inference speedup on Llama2-7B.

\section{Related Work}
\label{sec:Related}

\subsection{Contextual Sparsity in LLMs}

LLMs exhibit contextual sparsity in internal computation, where only a subset of neurons is activated under a given input context~\cite{michel2019sixteen,geva2021transformer,bills2023language,luo2025icp}. This property has motivated conditional-execution methods that dynamically select computation structures via learned predictors or input-adaptive thresholding. Deja Vu~\cite{liu2023deja} employs per-layer predictors to activate a subset of attention heads and MLP neurons, with subsequent works improving the robustness of predictor-based sparsity
\cite{akhauri2024shadowllm}. In parallel, training-free thresholding approaches induce sparsity by suppressing low-utility activations in an input-dependent manner\cite{lee2024cats,zhou2024sirius,liu2025training}.· System- and architecture-level efforts further exploit this property for efficient inference and sparsity-aware design
\cite{song2024powerinfer,mirzadeh2023relu}.

MoE architectures instantiate contextual sparsity at a coarser granularity by replacing parts of the dense feed-forward computation with multiple experts and using a learned router to select only a small subset of experts for each token \cite{lepikhin2021gshard,zhou2022mixture}. This design yields conditional computation, enabling parameter scaling while keeping per-token FLOPs roughly bounded by the number of activated experts \cite{dikkala2023benefits}.

\subsection{Unstructured \& Structural Pruning}
Model weight pruning compresses LLMs by removing parameters while preserving accuracy.
Unstructured pruning achieves high sparsity with limited post-training recovery:
SparseGPT~\cite{frantar2023sparsegpt} performs one-shot reconstruction-based pruning,
while Wanda~\cite{sun2024simple} adopts activation-informed saliency,
with subsequent methods improving layer- or blockwise sparsity allocation
\cite{xu2024besa,yin2023outlier}.
However, irregular sparsity often yields limited latency gains on mainstream GPUs,
due to poor compatibility with dense kernels
in the absence of specialized sparse runtimes.

Structured pruning removes contiguous architectural components,
offering more predictable acceleration on standard hardware
\cite{wang2020structured,xia2022structured,hu2025fasp}.
LLM-Pruner~\cite{ma2023llm} prunes coupled substructures with lightweight tuning,
and LoRAPrune~\cite{zhang2023loraprune} integrates structured removal
with low-rank adaptation~\cite{hu2022lora}.
SliceGPT~\cite{ashkboos2024slicegpt} and ZipLM~\cite{kurtic2023ziplm}
further align pruning decisions with dense inference efficiency.
For stronger compression, Sheared-LLaMA~\cite{xia2023sheared}
combines structured pruning with continued training.
Beyond static policies, recent methods explore adaptive sparsity allocation
across layers or blocks
\cite{zhong2025blockpruner,zhang2405finercut,men2024shortgpt,yang2025let,he2025olica},
including FLAP~\cite{an2024fluctuation}, which uses fluctuation-based criteria.
Global optimization approaches ~\cite{Li2025TyrThePruner}
employ evolutionary search to identify model-wide sparsity distributions
under constrained budgets.

\subsection{Context-Conditioned Pruning}
Context-conditioned pruning adapts the pruning mask to each input, rather than applying a single context-agnostic mask after compression. EBERT \cite{liu2021ebert} is an early method that dynamically prunes structured components using a lightweight input-conditioned predictor. Recent work revisits this paradigm for modern LLMs and extends the conditioning signal and optimization targets. Instruction-Following Pruning \cite{hou2025instruction} learns an instruction-conditioned mask predictor and jointly optimizes it with the LLM so that different instructions activate different parameter subsets. Probe Pruning \cite{qi2025probe} estimates batch-specific importance via lightweight probing and then applies online structured pruning for the remaining inference. Subsequent studies introduce alignment-aware constraints to stabilize dynamic masking under distribution shift \cite{patel2025alignment}. RAP \cite{liu2025rap} formulates runtime pruning as a control problem and uses an RL agent to adapt pruning decisions under memory budgets that couple model weights and KV-cache. $\mu$-MoE \cite{koike2025mu} casts prompt-adaptive test-time pruning as micro-grained expert selection, enabling per-prompt structured sparsity without committing to a single mask.

\begin{table}[tb]
\footnotesize
\centering
\caption{Comparison of pruning methods. Our approach prunes based on contextual input at inference time, requiring neither calibration nor additional training overhead.}
\setlength{\tabcolsep}{2pt}
\resizebox{\columnwidth}{!}{
\begin{tabular}{lcccc}
\toprule
Method & w/o.SFT & w/o.Calib & w/o.Predictor & Context-Cond. \\
\midrule
LLM-Pruner & \xmark & \xmark & \xmark & \xmark \\
SparseGPT & \cmark & \xmark & \cmark & \xmark \\
Wanda & \cmark & \xmark & \cmark & \xmark \\
FLAP & \cmark & \xmark & \cmark & \xmark \\
T{\'y}r-the-Pruner & \cmark & \xmark & \cmark & \xmark \\
Deja vu & \cmark & \xmark & \xmark & \cmark \\
IF-Pruning & \xmark & \xmark & \xmark & \cmark \\
SkipGPT & \xmark & \cmark & \xmark & \cmark \\
Probe Pruning & \cmark & \xmark & \cmark & \cmark \\
\rowcolor{thd}
\textbf{Ours} & \cmark & \cmark & \cmark & \cmark \\
\bottomrule
\end{tabular}
}
\label{Tab:Comparisons}
\end{table}
\section{Method}
\label{sec:Method}
The objective of \paperN is to offer online structural pruning at inference time to improve computational efficiency and minimize latency while preserving accuracy. The comparisons between \paperN and previous works are listed in Table~\ref{Tab:Comparisons}. We begin with the preliminaries of structural pruning and LLM architectures in \Cref{sec:3.1}. Motivated by the contextual sparsity of LFMs, we empirically show the correlation of pruning partitions and formulate \paperN in \Cref{sec:3.2}. \Cref{sec:3.3} subsequently elaborates our pruning metric. \Cref{sec:3.4} analyzes pruning ratios and computation overhead.

\subsection{Preliminaries and Notations}
\label{sec:3.1}
\paragraph{Notation.}
We denote a linear transformation by a weight matrix
$\mathbf{W} \in \mathbb{R}^{C_{\mathrm{out}} \times C_{\mathrm{in}}}$,
where $C_{\mathrm{in}}$ and $C_{\mathrm{out}}$ indicate the input and output channel dimensions, respectively.
Let $\mathbf{X} \in \mathbb{R}^{B \times L \times C_{\mathrm{in}}}$ denote the input activations,
where $B$ is the batch size and $L$ is the sequence length.
During autoregressive decoding, the activation at step $t$ is denoted as
$\mathbf{X}_t \in \mathbb{R}^{B \times 1 \times C_{\mathrm{in}}}$. We use $i$ to index output channels and $k$ to index input channels.
Channel-wise importance scores are denoted by $I_i$,
and their step-wise counterparts during decoding are written as $I_i(t)$.
The target pruning ratio is denoted by $r$.

\paragraph{Weight-Activation Structural Pruning.} We consider the common activation-aware formulation for structural pruning, where importance is estimated jointly from model weights and input activations \cite{sun2024simple}. The element-wise importance weight can be defined as
\begin{equation}
I_{i,k} = |W_{i,k}| \cdot \|\mathbf{X}_{k}\|_2 ,
\end{equation}
where \(\|\mathbf{X}_{k}\|_2\) denotes the $\ell_2$ norm of activations of the \(k\)-th input channel aggregated across tokens and samples.

While importance can be computed for individual weights, structural pruning operates on channel-level structures rather than isolated parameters.
Accordingly, following \cite{an2024fluctuation}, element-wise importance scores are aggregated to obtain output-channel importance, which can be defined as
\begin{equation}
I^{\mathrm{out}}_i
=
\mathcal{A}
\left(
\{ I_{i,k} \}_{k=1}^{C_{\mathrm{in}}}
\right),
\quad i = 1, \dots, C_{\mathrm{out}},
\end{equation}
where  $\mathcal{A}(\cdot)$ denotes an aggregation operator that maps element-wise scores to a channel-level importance value.

The resulting output-channel importance vector \(\mathbf{I}^{\mathrm{out}}\) serves as the basic criterion for structural pruning, where pruning decisions are applied to entire channels as a unit.
In feed-forward networks (FFNs), each output channel corresponds to one dimension
of the intermediate representation, which is consistently used across consecutive
linear transformations.
Therefore, output-channel importance provides a principled basis for structural pruning of FFN blocks.

\subsection{Contextual Sparsity for Online Pruning} 
\label{sec:3.2}
As discussed in the preliminaries, importance estimation in LFMs depends jointly on fixed model weights and input-dependent activation statistics.
During inference, this coupling induces a structural mismatch: pruning targets remain fixed at the channel level, whereas the activation patterns used for importance estimation vary across inference stages.
In particular, while prefilling computes importance from full-sequence activations of the input prompt, decoding relies on step-wise activations $\mathbf{X}_t$, in which the effective token context collapses from length $L$ to a single token, leading to token-dependent fluctuations in channel importance.

\begin{figure}[t]
    \centering
    \includegraphics[width=1\linewidth]{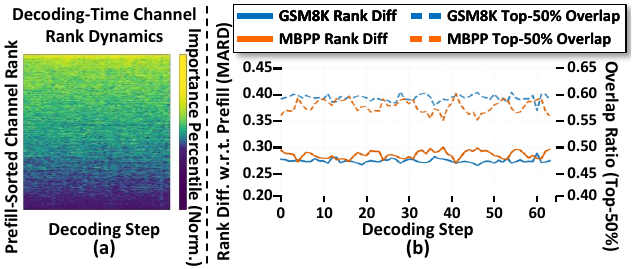}
    \caption{Decoding-time structural dynamics under online pruning.
(a) Channel-rank evolution across decoding steps, where channels are ordered
by their prefilling-stage importance and colors indicate normalized rank percentiles.
(b) Mean rank difference and Top-50\% channel overlap with respect to the prefilling ranking
on the generation benchmarks GSM8K and MBPP using Llama3.1-8B,
reflecting the degree of ranking variation
and the stability of highly important channels during decoding.}
    \label{fig:decoding_dynamics}
\end{figure}

Figure~\ref{fig:decoding_dynamics} provides a direct visualization of this phenomenon.
Channels are sorted according to their importance ranking obtained at the prefilling stage,
and their relative rankings at each decoding step are tracked thereafter.
As illustrated in Figure~\ref{fig:decoding_dynamics}(a),
channel rankings exhibit noticeable temporal variations during decoding,
yet the overall ordering remains strongly anchored to the prefilling reference.
This behavior is further quantified in Figure~\ref{fig:decoding_dynamics}(b).
The mean rank difference reflects the degree of fine-grained ranking variation across decoding steps,
while the Top-50\% channel overlap measures the stability of highly important channels.
Across both GSM8K and MBPP, we observe that although local ranking perturbations persist, a substantial portion of top-ranked channels remains consistent throughout decoding.

Together, these observations reveal a characteristic pattern of decoding-time behavior:
global importance structure remains largely stable,
whereas intermediate channels exhibit context-dependent fluctuations.
We refer to this coexistence of structural stability and localized variation as
\emph{contextual sparsity}.
As a result, pruning decisions determined solely at the prefilling stage
are insufficient to fully capture decoding-time computation,
motivating the need for online structural pruning
that adapts pruning decisions to the evolving generation context.


\begin{figure*}[t]
    \centering
    \includegraphics[width=\linewidth]{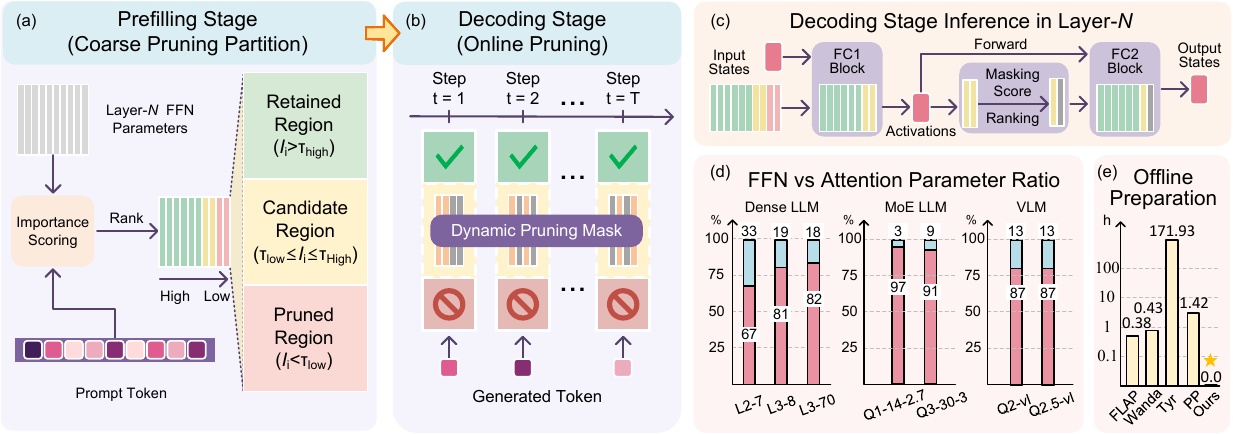}
    \caption{
    The overview of \textbf{P}artition-guided \textbf{O}nline \textbf{P}runing (POP). (a) During the prefilling stage, FFN channels in each layer are ranked by importance and partitioned into retained, candidate, and pruned regions, forming a coarse pruning partition.
(b) During decoding, \textsc{POP} performs online pruning by dynamically selecting candidate channels at each generation step using step-wise pruning masks.
(c) At each decoding step, intermediate activations are used to compute importance scores of candidates and update the pruning mask before the final FFN computation in layer $N$.
(d) The proportion of FFN parameters dominates over attention parameters across dense LLMs, MoE LLMs, and VLMs, motivating our FFN-focused pruning.
(e) Compared with prior methods, \textsc{POP} requires no offline preparation cost. 
    }
    \label{fig:overview}
\end{figure*}

\subsection{POP Architecture}
\label{sec:3.3}
Motivated by the contextual sparsity observed above, we design \paperN as a two-stage online pruning framework that separates
coarse-grained pruning decisions from fine-grained, context-conditioned pruning during autoregressive decoding. The overview of \paperN is illustrated in Figure \ref{fig:overview} (a) and (b). We focus on pruning FFNs, which dominate the model size as shown in Figure~\ref{fig:overview}(d), while the proposed framework can be extended to other components such as attention heads.

\paragraph{Calibration Stage.}
Unlike previous approaches \cite{sun2024simple,an2024fluctuation,qi2025probe,Li2025TyrThePruner} that require offline calibration, shown in Figure~\ref{fig:overview} (e), \paperN is a plug-and-play framework that requires no preparation or training, and is ready for out-of-the-box inference.

\paragraph{Prefilling Stage.}
Inference begins with the \emph{prefilling} stage,
which processes all prompt tokens in parallel
and produces full-sequence activations
$\mathbf{X} \in \mathbb{R}^{B \times L \times C_{\mathrm{in}}}$.
These activations offer a global view of channel utilization
under the given input prompt.
Leveraging this information,
\paperN derives an initial \emph{coarse pruning partition}.
Channel-wise importance scores $I_i^{\mathrm{out}}$ are computed from the full input prompt
$\mathbf{X}$, following the formulation in Section~\ref{sec:3.1}.

Given a target pruning ratio $r$ and a partition width $\delta$,
\paperN defines a narrow importance band around the pruning threshold.
Let $q_{\alpha}(\cdot)$ denote the $\alpha$-quantile of the importance scores
$\{ I_i^{\mathrm{out}} \}_{i=1}^{C_{\mathrm{out}}}$.
Two thresholds are defined as
$\tau_{\mathrm{low}} = q_{\,r-\delta}$ and
$\tau_{\mathrm{high}} = q_{\,r+\delta}$.
Channels are then partitioned into three disjoint regions:

\begin{itemize}
\item \textbf{Retained Region} ($\mathcal{R}$):
$\mathcal{R} = \{\, i \mid I_i^{\mathrm{out}} \ge \tau_{\mathrm{high}} \,\}$.
Channels in this region exhibit consistently high importance and are always preserved,
forming a stable computational backbone throughout inference.

\item \textbf{Pruned Region} ($\mathcal{P}$):
$\mathcal{P} = \{\, i \mid I_i^{\mathrm{out}} \le \tau_{\mathrm{low}} \,\}$.
These channels contribute marginally and are removed upfront
to improve inference efficiency for the given input prompt.

\item \textbf{Candidate Region} ($\mathcal{C}$):
$\mathcal{C} = \{\, i \mid \tau_{\mathrm{low}} < I_i^{\mathrm{out}} < \tau_{\mathrm{high}} \,\}$.
This region captures channels with context-conditioned variability and defines a bounded
\emph{dynamic search space} for online pruning.
\end{itemize}

\paragraph{Decoding Stage.}
Following the prefilling stage, autoregressive models proceed one token at a time in the decoding stage, with each step producing activations
$\mathbf{X}_t \in \mathbb{R}^{B \times 1 \times C_{\mathrm{in}}}$.  At each decoding step $t$, \paperN performs fine-grained, context-conditioned pruning
within the candidate region.
Given the step-wise activation $\mathbf{X}_t$,
channel importance scores $I_i^{\mathrm{out}}(t)$ are evaluated only for channels $i \in \mathcal{C}$,
while channels in $\mathcal{R}$ remain active and channels in $\mathcal{P}$ remain disabled.
By preserving the retained region and avoiding global importance re-evaluation,
\paperN reduces the online overhead of pruning without degrading inference performance.

Based on the step-wise importance scores $I_i^{\mathrm{out}}(t)$ evaluated for channels $i \in \mathcal{C}$,
\paperN selects a subset of candidate channels
such that the total number of active channels
matches the target pruning ratio $r$.
This step enables fine-grained, context-conditioned pruning,
as channel selection is adaptively determined by the decoding context. The selected channels are then combined with the retained region to form the effective computation graph for decoding step $t$.

\subsection{Pruning Metric and Inference Efficiency}
\label{sec:3.4}
In \paperN, we adopt activation-aware importance metrics
following prior structural pruning methods
such as Wanda~\cite{sun2024simple}
and Probe Pruning~\cite{qi2025probe}.
Rather than proposing new importance formulations, we focus on enabling their efficient online use during inference.

Given activations $\mathbf{X}$
and a linear transformation parameterized by
$\mathbf{W}$, the importance of the $i$-th output channel is computed as
\begin{equation}
I_i(\mathbf{X})
=
\left\|
\mathbf{W}_{:,i} \odot \mathbf{s}
\right\|_2,
\quad
\mathbf{s}_k = \|\mathbf{X}_{:,:,k}\|_2 ,
\end{equation}
which reflects the joint contribution of activation magnitude and associated weights.

The same importance formulation is used throughout inference.
During prefilling, importance is first estimated from full-sequence or partial probing prompt activations
to determine coarse pruning decisions.
The prefilling inference is then completed under the resulting pruning structure,
which is shared with the subsequent decoding stage.

To balance estimation accuracy and online efficiency,
\paperN performs decoding in two steps.
Retained and candidate channels first participate in the forward computation,
yielding intermediate activations
$\mathbf{H}_t = f(\mathbf{X}_t; \mathcal{R} \cup \mathcal{C})$.
Step-wise importance scores $I_i(t)$ are then evaluated from $\mathbf{H}_t$
for channels $i \in \mathcal{C}$. Based on $I_i(t)$, a subset of candidate channels is selected in a context-dependent manner.
Only the retained channels and the selected candidates are involved
in the final computation for the current decoding step.

\section{Experiment}
\label{sec:Experiment}

\paragraph{Large Foundation Models.}

Our evaluation spans dense LLMs, VLMs, and MoE models.
For dense large language models, we consider Llama-2 7B/13B/70B \cite{touvron2023llama}, Llama-3.1 8B/70B \cite{dubey2024llama}, and Qwen-3 8B/32B \cite{yang2025qwen3}.To examine MoE architectures, we include Qwen1.5-MoE-A2.7B \cite{qwen_moe}, Qwen2-57B-A14B \cite{yang2024qwen2technicalreport}, and Qwen3-30B-A3B \cite{yang2025qwen3}. VLMs are represented by Qwen2-VL \cite{Qwen2-VL} and Qwen2.5-VL \cite{Qwen2.5-VL}. Additional experiments on pure vision backbones from SAM \cite{kirillov2023segment} are reported in the Appendix~\ref{app:sam}.

\paragraph{Evaluation Benchmarks.}
We conduct our evaluations using the \texttt{LM-Evaluation-Harness} framework~\cite{eval-harness} for LLMs and MoE models, and the \texttt{LMMs-Eval} framework~\cite{zhang2024lmmsevalrealitycheckevaluation}
for VLMs.
The benchmarks are grouped into three categories according to their input modalities
and inference characteristics:
(i) \textbf{question answering (QA)} benchmarks for evaluating language understanding in LLMs and MoEs,
(ii) \textbf{generation} benchmarks that require extended autoregressive inference,
and (iii) \textbf{visual question answering (VQA)} benchmarks for assessing multimodal reasoning in VLMs.
The QA benchmarks include BoolQ~\cite{clark-etal-2019-boolq},
RTE~\cite{poliak2020surveyrecognizingtextualentailment},
HellaSwag~\cite{zellers2019hellaswag},
WinoGrande~\cite{sakaguchi2019winograndeadversarialwinogradschema},
ARC-Easy~\cite{allenai:arc},
ARC-Challenge~\cite{allenai:arc},
and OpenBookQA~\cite{OpenBookQA2018}.
These tasks primarily evaluate model understanding and reasoning capability
under limited-output inference settings.
The text generation benchmarks include CoQA~\cite{reddy2019coqaconversationalquestionanswering},
MBPP~\cite{austin2021program},
NQ-Open~\cite{10.1162/tacl_a_00276},
HumanEval~\cite{chen2021evaluating},
and GSM8K~\cite{cobbe2021gsm8k},
which require multi-step autoregressive generation and thus provide a more challenging
testbed for decoding-stage efficiency.
The VQA benchmarks include POPE~\cite{li2023evaluating},
OK-VQA~\cite{marino2019okvqavisualquestionanswering},
GQA~\cite{hudson2019gqanewdatasetrealworld},
ScienceQA~\cite{lu2022learn},
and MME~\cite{fu2025mmecomprehensiveevaluationbenchmark}.
The evaluation metrics used for benchmarks are detailed in Appendix~\ref{app:metrics},
and the complete results corresponding to each table are provided in Appendix~\ref{app:results}.
In addition, perplexity-based evaluations are reported in Appendix~\ref{app:perplexity}.

\paragraph{Baseline Methods.}
We evaluate our method against recent structural pruning baselines, including Wanda~\cite{sun2024simple}, FLAP~\cite{an2023fluctuationbased}, and T{\'y}r~\cite{Li2025TyrThePruner}, which rely on offline pruning, as well as Probe Pruning~\cite{qi2025probe}, which supports online pruning.
For Wanda, we adopt its structural pruning variant, denoted as Wanda-sp.
The detailed settings for the offline stage of these baselines are provided in the Appendix~\ref{app:calibration}.

\paragraph{Implementation Details.}
We implemented all baseline methods strictly following their original designs. $\gamma$ in POP for the partition fraction was set to 0.1. All experiments were conducted on RTX A6000 clusters, and the batch size is set to 10 for LLM evaluations and 1 for VLM evaluations. Since our method exclusively prunes FFN layers while preserving attention mechanisms, we adjust the pruning ratio within FFN blocks to match the target level of parameter reduction for fair comparison. More configuration details for pruning ratio and experimental settings are outlined in Appendix~\ref{app:PR}. Best results in experiments are shaded as \colorbox{fst}{first}, \colorbox{sed}{second}, and \colorbox{thd}{third}. \\

\subsection{Large Language Models}
\begin{table*}[t]
\centering
\footnotesize
\caption{Average results for Llama-2 \cite{touvron2023llama}, Llama-3.1 \cite{dubey2024llama}, and Qwen-3 \cite{yang2025qwen3} across different pruning ratios on 7 QA tasks and 5 Generation tasks. The dash symbol indicates the corresponding setting is not applicable.}
\setlength{\tabcolsep}{2.1pt}
\renewcommand{\arraystretch}{1.15}

\begin{tabular}{c l c | ccc | cc | cc | c || ccc | cc | cc | c }
\hline
\multirow{3}{*}{PR} & \multirow{3}{*}{Method} & \multirow{3}{*}{Online}
& \multicolumn{8}{c||}{Average QA tasks Accuracy (\%) $\uparrow$}
& \multicolumn{8}{c}{Average Generation tasks Accuracy (\%) $\uparrow$} \\
\hline
& & & \multicolumn{3}{c|}{Llama2} & \multicolumn{2}{c|}{Llama3.1} & \multicolumn{2}{c|}{Qwen3} & Avg. & \multicolumn{3}{c|}{Llama2} & \multicolumn{2}{c|}{Llama3.1} & \multicolumn{2}{c|}{Qwen3} & Avg. \\
& & & 7B & 13B & 70B & 8B & 70B & 8B & 32B
&  & 7B & 13B & 70B & 8B & 70B & 8B & 32B & \\\hline
0\% & N/A & N/A & 59.70 & 63.00 & 66.96 & 64.68 & 69.44 & 65.74 & 68.28 & 65.40 & 28.69 & 33.07 & 45.49 & 44.58 & 61.14 & 60.61 & 55.09 & 45.77
\\
\hline

\multirow{5}{*}{20\%}
& Wanda-sp
& \xmark & 55.02 & 53.25 & 32.61 & 34.01 & 32.44 & 44.82 & 51.80 & 43.42 & 12.20 & 10.96 & 0.84 & 0.79 & 0.05 & 8.97 & 14.91 & 6.96
\\
& FLAP
& \xmark & 54.24 & 57.87 & 35.34 & 53.02 & 38.55 & \thd 54.44 & \thd 53.50 & 49.56 & 17.23 & 19.04 & 2.99 & 16.48 & 1.68 & \thd 14.91 & \thd 17.97 & 12.90
\\
& PP
& \cmark & \sed 57.51 & \thd 60.36 & \fst 66.89 & \thd 54.35 & \thd 67.37 & - & - & \thd\underline{61.29} & \thd 19.91 & \sed 27.76 & \sed 41.83 & \thd 19.66 & \sed 51.51 & - & - & \sed\underline{32.13}
\\
& T{\'y}r
& \xmark & \fst 58.11 & \fst 62.38 & \thd 65.58 & \fst 59.85 & \fst 68.19 & \fst 63.81 & \fst 68.16 & \fst63.72 & \sed 20.75 & \thd 23.92 & \thd 38.80 & \sed 21.14 & \thd 46.81 & \sed 22.69 & \sed 35.68 & \thd29.97
\\
& Ours
& \cmark & \thd 56.78 & \sed 60.43 & \sed 65.93 & \sed 59.62 & \sed 67.90 & \sed 61.50 & \sed 64.24 & \sed 62.34 & \fst 24.12 & \fst 28.97 & \fst 42.04 & \fst 37.37 & \fst 55.67  & \fst 56.13 & \fst 53.04 & \fst 42.48
\\
\hline

\multirow{5}{*}{40\%}
& Wanda-sp
& \xmark & 41.39 & 37.05 & 32.77 & 33.65 & 31.91 & 34.55 & 40.74 & 36.01 & 4.93 & 2.36 & 0.33 & 0.06 & 0.03 & \thd 0.61 & 5.07 & 1.91
\\
& FLAP
& \xmark & \thd 45.65 & 47.59 & 33.00 & \thd 43.17 & 34.41 & \thd 36.66 & \thd 40.75 & 40.17 & \thd 9.08 & 14.19 & 0.19 & \thd 6.84 & 0.64 & 0.40 & \thd 6.08 & 5.35
\\
& PP
& \cmark & \sed 49.39 & \sed 54.14 & \sed 63.23 & 43.09 & \thd 60.08 & - & - & \sed \underline{53.99} & 8.68 & \thd 14.36 & \sed 33.06 & 1.53 & \thd 28.47 & - & - & \thd \underline{17.22}
\\
& T{\'y}r
& \xmark & \fst 52.01 & \fst 57.88 & \fst 63.73 & \fst 52.01 & \fst 65.90 & \fst 53.73 & \fst 67.21 & \fst 58.92 & \sed 11.27 & \sed 17.09 & \thd 24.92 & \sed 10.31 & \sed 30.24 & \sed 11.22 & \sed 29.60 & \sed 19.24
\\
& Ours
& \cmark & 43.95 & \thd 48.15 & \thd 62.58 & \sed 49.26 & \sed 62.53 & \sed 50.98 & \sed 55.98 & \thd 53.35 & \fst 13.84 & \fst 17.99 & \fst 34.47 & \fst 23.34 & \fst 43.14 & \fst 42.73 & \fst 47.78 & \fst 31.90
\\
\hline
\end{tabular}
\label{tab:main}
\end{table*}

\paragraph{QA Tasks.}
We report the QA performance comparisons with baselines in Table~\ref{tab:main}. 
Under a 20\% pruning ratio, \paperN achieves the second-best performance across the majority of evaluated models.
Under a 20\% pruning ratio, \paperN outperforms Wanda-sp, FLAP, and Probe Pruning by $18.92\%$, $12.78\%$, and $1.05\%$ on average, respectively.
Although \paperN shows a modest performance gap of $1.38\%$ relative to T{\'y}r, this difference can be attributed to T{\'y}r's reliance on offline evolutionary search with calibration data.
By contrast, \paperN performs pruning decisions entirely online.

\paragraph{Generation Tasks.}

We report the results on generation tasks in Table~\ref{tab:main}.
Existing baselines, which rely on a static pruning mask during decoding, suffer from substantial performance degradation on generation tasks compared to QA settings.
This degradation is particularly pronounced for long-form generation, where offline pruning methods like Wanda-sp and FLAP exhibit near-collapse even under moderate pruning ratios.
In contrast, \paperN consistently outperforms all baseline methods across all evaluated models and pruning ratios.
Under a 20\% pruning ratio, \paperN achieves average performance improvements of $35.52\%$, $29.58\%$, $10.35\%$, and $12.51\%$, respectively, over the competing baselines.
Moreover, the performance gap is markedly larger on generation tasks, underscoring the need for context-conditioned pruning during autoregressive decoding.

\subsection{Mixture-of-Expert Models}
\begin{table}[tb]
\centering
\scriptsize
\caption{Average results for Qwen1.5-MoE \cite{qwen_moe}, Qwen2-MoE \cite{yang2024qwen2technicalreport} and Qwen3-MoE~\cite{yang2025qwen3} models across different pruning ratios on 7 QA tasks and 5 Generation tasks. The dash symbol indicates the corresponding setting is not applicable.}
\label{tab:moemain}
\setlength{\tabcolsep}{5.5pt}
\renewcommand{\arraystretch}{1.15}
\begin{tabular}{c l | c | c | c || c | c | c}
\hline
\multirow{3}{*}{PR} & \multirow{3}{*}{Method}
& \multicolumn{3}{c||}{Avg. QA Acc. (\%) $\uparrow$}
& \multicolumn{3}{c}{Avg. Gen. Acc. (\%) $\uparrow$} \\
\cline{3-8}
& & Q1.5 & Q2 & Q3 & Q1.5 & Q2 & Q3 \\
& & A2.7B & A14B & A3B & A2.7B & A14B & A3B \\
\hline
0\% & N/A & 60.21 & 64.71 & 66.76 & 43.17 & 54.92 & 57.77 \\
\hline
\multirow{4}{*}{20\%} 
& Wanda-sp & \thd53.97 & \thd 40.61& \thd 52.37 & 13.21 & \thd 2.35 & \thd 15.71 \\
& FLAP     & 53.58 & \fst 65.04 & 32.34 & \thd 20.12 & \sed 37.53 & 0.00 \\
& T{\'y}r   & \fst 59.61 & - & \sed 63.80 & \sed 27.66 & - & \sed 43.07 \\
& Ours     & \sed 58.40 & \sed 64.64 & \fst 64.28 & \fst 42.10 & \fst 53.95 & \fst 57.96 \\
\hline
\multirow{4}{*}{40\%} 
& Wanda-sp & 38.04 & \thd 34.43 & \thd 39.01 & 2.71 & \thd 0.39 & \thd 1.54\\
& FLAP     & \sed 49.81 & \sed 58.12 & 32.34 & \sed 15.15 & \sed 17.26 & 0.00 \\
& T{\'y}r   & \thd 48.74 & - & \sed 53.74 & \thd 7.81 & - & \sed 15.87 \\
& Ours     & \fst 53.95 & \fst 64.24 & \fst 54.94 & \fst 31.11 & \fst 53.86 & \fst 38.65 \\
\hline
\end{tabular}
\end{table}

We evaluate \paperN on MoE models using the same protocol as for dense LLMs.
As shown in Table~\ref{tab:moemain}, \paperN consistently outperforms baselines across both QA and generation tasks under different pruning ratios.
At a pruning ratio of 20\%, \paperN preserves strong QA performance
while achieving substantially higher generation accuracy than baselines.
Under a more aggressive 40\% pruning ratio, baseline approaches exhibit severe degradation, whereas \paperN remains robust. These results suggest that online structural pruning is particularly beneficial for MoE architectures,
possibly due to the additional variability introduced by dynamic expert routing
between prefilling-stage importance estimation and decoding-stage computation.

\subsection{Vision Language Models}

\begin{table}[tb]
\centering
\scriptsize
\caption{Average results for Qwen2-VL \cite{Qwen2-VL} and Qwen2.5-VL \cite{Qwen2.5-VL} on 5 VQA tasks under different pruning ratios. }
\label{tab:vlm_main}

\setlength{\tabcolsep}{6.5pt}
\renewcommand{\arraystretch}{1.15}

\begin{tabular}{c | ccc | ccc}
\hline
PR 
& \multicolumn{3}{c|}{Qwen2-VL} 
& \multicolumn{3}{c}{Qwen2.5-VL} \\
\cline{2-7}
& Wanda-sp & FLAP & \textbf{Ours}
& Wanda-sp & FLAP & \textbf{Ours} \\
\hline
0\% & -- & -- & 62.14
& -- & -- & 60.68 \\
\hline
20\% & \thd 7.86 & \sed 46.74 & \fst 59.94 & \thd 3.71 & \sed 42.53 & \fst 59.46 \\
40\% & \thd 0.02 & \sed 0.01 & \fst 54.33 & 0.00 & 0.00 & \fst 54.63 \\
\hline
\end{tabular}
\end{table}


We further evaluate the generalizability of \paperN on VLMs.
We apply \paperN and baselines to Qwen2-VL and Qwen2.5-VL, and report their performance on five VQA benchmarks.
As shown in Table~\ref{tab:vlm_main}, existing baselines suffer severe performance degradation under pruning, with accuracy dropping to near zero at higher sparsity levels.
In contrast, \paperN consistently preserves strong VQA performance across both models.
Under a 20\% pruning ratio, \paperN achieves average improvements of $52.08\%$ and $13.20\%$ over Wanda-sp and FLAP on Qwen2-VL, and $55.75\%$ and $16.93\%$ on Qwen2.5-VL, respectively.
These results demonstrate that \paperN generalizes effectively to multimodal architectures and remains robust across different application domains.

\subsection{Inference Efficiency}
\begin{table}[t]
\centering
\caption{Comparison of inference latency between \paperN and Probe Pruning~\cite{qi2025probe} on Llama2-7B~\cite{touvron2023llama}.
Latency is reported for attention, MLP, and end-to-end inference, with speedup measured relative to the dense model.
Measurements are performed using CUDA event timing with input and output sequence lengths fixed to 128 tokens.}
\setlength{\tabcolsep}{4pt}
\renewcommand{\arraystretch}{1.15}
\resizebox{\columnwidth}{!}{
\begin{tabular}{c l | cc | cc | cc}
\hline
\multirow{2}{*}{PR} & \multirow{2}{*}{Method} & \multicolumn{2}{c|}{Attention} & \multicolumn{2}{c|}{MLP} & \multicolumn{2}{c}{E2E Latency} \\
\cline{3-8}
& & Lat. (ms) & Speedup & Lat. (ms) & Speedup & Lat. (s) & Speedup \\
\hline
0\% & Dense & 42 & - & 53 & - & 3.04 & - \\
\hline
& PP & 40 & 1.05$\times$ & 44 & 1.20$\times$ & 2.69 & 1.13$\times$ \\
\rowcolor{thd}
\cellcolor{white}\multirow{-2}{*}{20\%}
& \textbf{Ours} & 42 & 1.00$\times$ & 41 & 1.29$\times$ & 2.66 & 1.14$\times$ \\
\hline
& PP & 37 & 1.14$\times$ & 30 & 1.77$\times$ & 2.14 & 1.42$\times$ \\
\rowcolor{thd}
\cellcolor{white}\multirow{-2}{*}{40\%} 
& \textbf{Ours} & 42 & 1.00$\times$ & 27 & 1.96$\times$ & 2.21 & 1.38$\times$ \\
\hline
\end{tabular}
}
\label{tab:latency}
\end{table}

To evaluate the efficiency of \paperN, we profile the latency of attention and FFN modules using CUDA event timing, and report the resulting end-to-end inference latency in Table~\ref{tab:latency}.
Since \paperN does not prune attention weights, attention latency remains unchanged, unlike Probe Pruning, which directly prunes attention weights.
To achieve the same target reduction in model parameters, \paperN applies a higher pruning ratio to the FFN layers.
This design results in a more pronounced reduction in latency within the FFN blocks, partially compensating for the absence of attention pruning.
As a result, \paperN achieves end-to-end inference latency comparable to Probe Pruning, while avoiding attention pruning during decoding.
Overall, \paperN achieves inference speedups of $1.14\times$ and $1.38\times$ over the dense model under 20\% and 40\% pruning ratios, respectively.


\section{Ablation Study}
\label{sec:ablation}
\begin{table}[t]
\centering
\caption{
Ablation study of the accuracy-efficiency trade-off under different pruning strategies.
We report the mean accuracy over 5 generative tasks and the MLP-level FLOPs overhead.
FLOPs are reported in TFLOPs, and overhead (\%) denotes the relative increase in MLP computation compared to the dense FFN.
Measurements are conducted using DeepSpeed~\cite{10.1145/3394486.3406703}.
}
\footnotesize
\setlength{\tabcolsep}{4pt}
\renewcommand{\arraystretch}{1.15}

\begin{tabular}{llccc}
\toprule
Model & Method & Mean & FLOPs & Overhead (\%) \\
\midrule

\multirow{4}{*}{Llama2-7B}
& FFN              & --    & 22.09 & --  \\
& Variant (1) & 23.40 & 0.00  & 0.0 \\
& Variant (2)  & 26.56 & 7.32  & 33.1 \\
& \textbf{Ours}    & \textbf{24.12} & \textbf{0.63} & \textbf{2.85} \\

\midrule

\multirow{4}{*}{Llama3.1-8B}
& FFN              & --    & 28.75 & --  \\
& Variant (1)& 34.69 & 0.00  & 0.0 \\
&  Variant (2)    & 38.95 & 9.56  & 33.3 \\
& \textbf{Ours}    & \textbf{37.37} & \textbf{1.00} & \textbf{3.48} \\

\bottomrule
\end{tabular}
\label{tab:ablation_tradeoff}
\end{table}


\paragraph{Accuracy–Efficiency Trade-off under Different Pruning Strategies.}


We justify the design of \paperN by comparing it with two alternative implementations that employ the same pruning metric and FFN-level structure. 

As shown in Table~\ref{tab:ablation_tradeoff}, Variant (1) reuses the pruning mask obtained during the prefilling stage for all decoding steps. While this strategy introduces no additional overhead during decoding, the fixed pruning mask cannot adapt to the evolving decoding context, leading to a degradation in accuracy.

Variant (2) recomputes fine-grained pruning decisions over all channels at every decoding step. Although this approach achieves better performance, it incurs substantial computational overhead, introducing over 30\% additional FFN FLOPs during decoding and largely offsetting the efficiency gains of structural pruning.

In contrast, \paperN achieves a more balanced trade-off. By restricting online pruning to a small candidate subset, it avoids full-channel re-evaluation and incurs less than 4\% additional FFN cost with a partition fraction $\gamma=0.1$, while consistently improving accuracy over Variant (1). These results demonstrate the effectiveness of POP in enabling efficient, context-conditioned online pruning.

\begin{figure}[t]
    \centering
    \includegraphics[width=1\linewidth]{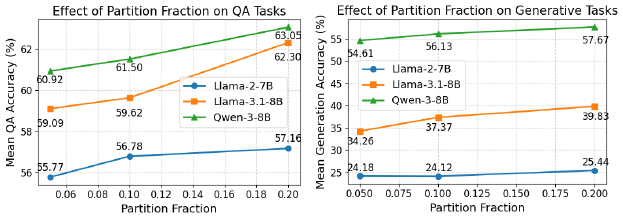}
    \caption{Ablation study on partition fraction $\gamma$ in \paperN.
Left: mean accuracy on zero-shot QA tasks.
Right: mean accuracy on generative tasks.}
    \label{fig:hyperparameter}
\end{figure}

\paragraph{Partition Fraction.}

To examine the effect of candidate region size, we vary the partition fraction $\gamma$ in \paperN and evaluate its impact on both QA and generation tasks, as shown in Figure~\ref{fig:hyperparameter}. As $\gamma$ increases, the candidate region expands, allowing more channels to participate in online selection, which consistently improves accuracy across all evaluated models. However, a larger candidate region also incurs higher decoding-time computational overhead. Balancing performance and efficiency, we set $\gamma=0.1$ as the default value in POP.

\paragraph{Offline Preparation Cost.}



\begin{table}[t]
\centering
\caption{Comparison of offline preparation time (hours) for baseline pruning methods on Llama2-7B~\cite{touvron2023llama}, Llama3.1-70B~\cite{dubey2024llama}, and Qwen1.5-30B-A3B~\cite{qwen_moe}. Zeros indicate that POP requires no offline preparation.}
\footnotesize
\renewcommand{\arraystretch}{1.05}
\setlength{\tabcolsep}{4pt}

\begin{tabular}{lccccc}
\toprule
Model & Wanda-sp & FLAP & PP & T{\'y}r & \textbf{POP} \\
\midrule
Llama2-7B       & 0.08 & 0.07 & 1.03 & 9.72  & \textbf{0.0} \\
Llama3.1-70B   & 0.43 & 0.38 & 1.42 & 171.93 & \textbf{0.0} \\
Qwen1.5-30B-A3B & 1.14 & 0.85 & --   & 45.44  & \textbf{0.0} \\
\bottomrule
\end{tabular}
\label{tab:offline_time}
\end{table}



Table~\ref{tab:offline_time} compares the offline preparation cost of different pruning methods.
While Wanda-sp~\cite{sun2024simple} and FLAP~\cite{an2024fluctuation} incur minimal calibration overhead,
Probe Pruning~\cite{qi2025probe} and T{\'y}r~\cite{Li2025TyrThePruner} require substantial offline preprocessing, while T{\'y}r introduces the highest offline cost, exceeding one day of preprocessing on large-scale models like Llama3.1-70B with four default iterations.
By contrast, \paperN is fully online and eliminates any offline preparation.

\section{Conclusion}
\label{sec:Conclusion}


In this work, we propose \paperN, an efficient online structural pruning pipeline that overcomes the limitations of static pruning strategies by enabling context-conditioned online pruning with minimal overhead. Extensive experiments across LLMs, MoEs, and VLMs demonstrate that \paperN consistently outperforms SOTA pruning methods across nearly all evaluated settings.
Beyond empirical improvements, POP highlights the importance of dynamic pruning decisions to evolving generation contexts, particularly during autoregressive decoding. We hope this work inspires further exploration of contextual sparsity and online pruning mechanisms and contributes to the development of more efficient and adaptive inference techniques for LFMs.

\section*{Impact Statement}

This paper presents work whose goal is to advance the field of Machine
Learning. There are many potential societal consequences of our work, none
which we feel must be specifically highlighted here.

\bibliography{ref}
\bibliographystyle{icml2026}

\newpage
\appendix
\onecolumn


\vspace{1em}
\section{Implementation Details}
\label{app:details}
In this section, we provide a detailed description of the experimental methodologies employed for both \paperN~and the baseline approaches. We elaborate on the specific configurations and implementation settings to ensure a rigorous and fair evaluation.

\subsection{Pruning Ratio Configurations}
\label{app:PR}
For the baseline methods Wanda-sp~\cite{sun2024simple} and Probe Pruning~\cite{qi2025probe}, a uniform pruning ratio is applied to both the self-attention and MLP layers, identical to the overall target sparsity.
In contrast, FLAP~\cite{an2023fluctuationbased}, which aggregates scores across all components, and T{\'y}r~\cite{Li2025TyrThePruner}, which utilizes evolutionary search, dynamically allocate pruning ratios between attention and MLP layers while satisfying the global target pruning ratio.

To ensure robust performance in generative tasks, \paperN~exclusively conducts MLP weight pruning and preserves attention mechanisms. We adjust the pruning ratio within the MLP blocks to match the total parameter reduction of these baselines for a fair comparison. 
Consequently, the effective sparsity applied to MLPs is inherently higher than the overall target pruning ratio. The specific effective MLP pruning ratios corresponding to each target FLOPs reduction are detailed in Table~\ref{tab:mlp-ratio}. All experiments were conducted on NVIDIA A6000 clusters.

\begin{table}[th]
\centering
\footnotesize
\caption{Proper MLP pruning ratios required to achieve the target FLOPs reduction. Since our method preserves attention layers, we apply higher sparsity to MLP blocks compared to the target pruning ratio. All values are rounded to the first decimal place.}
\setlength{\tabcolsep}{4.0pt}
\renewcommand{\arraystretch}{1.2}
\begin{tabular}{c | ccc | cc | cc || c | c | c || c | c}
\cline{1-13}
\multirow{4}{*}{Target PR} 
& \multicolumn{12}{c}{MLP Pruning Ratio (\%)} \\
\cline{2-13}
& \multicolumn{7}{c||}{LLM} & \multicolumn{3}{c||}{MoE} & \multicolumn{2}{c}{VLM} \\
\cline{2-13}
& \multicolumn{3}{c|}{Llama2} & \multicolumn{2}{c|}{Llama3.1} & \multicolumn{2}{c||}{Qwen3} & \multicolumn{1}{c|}{Qwen1.5} & \multicolumn{1}{c|}{Qwen2} & \multicolumn{1}{c||}{Qwen3} 
& \multicolumn{1}{c|}{Qwen2} & \multicolumn{1}{c}{Qwen2.5} \\

& 7B & 13B & 70B & 8B & 70B & 8B & 32B
& A2.7B & 57B-A14B & 30B-A3B
& VL & VL \\
\hline
\multirow{1}{*}{20\%} 
& 29.9 & 29.9 & 24.3 & 24.8 & 24.3 & 25.6 & 24.8 & 29.7 & 22.7 & 30.0  & 22.9 & 22.9
\\
\hline
\multirow{1}{*}{40\%} 
& 59.8 & 59.8 & 48.6 & 49.5 & 48.6 & 51.1 & 49.6 & 59.4 & 45.3 & 60.0 & 45.8 & 45.8
\\
\hline
\end{tabular}
\label{tab:mlp-ratio}
\end{table}


\subsection{Offline Calibration Stage}
\label{app:calibration}
To ensure a fair comparison, we utilized the C4 dataset~\cite{raffel2020exploring} with 2,000 samples for Wanda-sp~\cite{sun2024simple}, FLAP~\cite{an2023fluctuationbased}, and Probe Pruning~\cite{qi2025probe}. However, we distinguished the sequence length settings based on methodological requirements.
For Wanda-sp and FLAP, we set the sequence length to 1,024 tokens.
In contrast, for Probe Pruning, we employed a sequence length of 4,096 tokens. This setting matches the maximum position embedding of Llama2 and was necessary because Probe Pruning requires a historical state longer than the input sequence length to support generative tasks.

For T{\'y}r~\cite{Li2025TyrThePruner}, we adhered to the configuration specified in the original literature. We used the FineWeb dataset~\cite{penedo2024the} for calibration, conducting the offline stage with 1,000 samples and a maximum input length of 4k tokens.
Finally, \paperN~operates exclusively via online pruning, eliminating the need for a separate offline calibration stage.

\subsection{Perplexity}
\label{app:perplexity}
Table~\ref{tab:wikitext} presents the perplexity scores on the Wikitext~\cite{merity2016pointer} benchmark.
The primary goal of this experiment is to evaluate the robustness of the compressed models in general language understanding and generation scenarios.
While accuracy metrics on QA tasks demonstrate task-specific proficiency, perplexity provides a complementary view of the model's stability by quantifying the uncertainty in token prediction.
A lower perplexity indicates that the pruned model maintains a probability distribution close to that of the dense model.

\begin{table*}[t]
\centering
\footnotesize
\caption{Wikitext~\cite{merity2016pointer} perplexity results for LLM and MoE models across different pruning ratios. Lower perplexity ($\downarrow$) indicates better performance. Experiments are conducted using a batch size 10.}
\setlength{\tabcolsep}{6.0pt} 
\renewcommand{\arraystretch}{1.15}

\begin{tabular}{c l c | ccc | cc || c | c}
\hline
\multirow{3}{*}{PR} & \multirow{3}{*}{Method} & \multirow{3}{*}{Online}
& \multicolumn{5}{c||}{Wikitext Perplexity (LLMs) $\downarrow$}
& \multicolumn{2}{c}{Wikitext Perplexity (MoEs) $\downarrow$} \\
\hline
& & 
& \multicolumn{3}{c|}{Llama2} 
& \multicolumn{2}{c||}{Qwen3} 
& \multicolumn{1}{c|}{Qwen1.5-MoE} 
& \multicolumn{1}{c}{Qwen2} \\
& & 
& 7B & 13B & 70B 
& 8B & 32B
& A2.7B & 57B-A14B \\
\hline

0\% & N/A & N/A 
& 5.12 & 4.57 & 3.12 
& 9.00 & 7.02 
& 6.79 & 5.56
\\
\hline

\multirow{5}{*}{20\%} 
& Wanda-sp
& \xmark 
& 9.23 & 7.90 & 466.51 
& 785.16 & 11.55 
& 9.92 & 58.76
\\

& FLAP
& \xmark 
& 7.53 & 6.64 & 123.74 
& 13.99 & 8.23 
& 9.64 & 6.02
\\

& PP 
& \cmark 
& 6.17 & 5.25 & 3.61 
& - & - 
& - & -
\\

& T{\'y}r
& \xmark 
& 6.44 & 5.41 & 4.25 
& 11.66 & 7.20 
& 7.73 & -
\\

\rowcolor{yellow!15}
& Ours
& \cmark 
& 6.86 & 5.94 & 3.92 
& 10.95  &8.44 & 7.51 
&  5.75
\\
\hline

\multirow{5}{*}{40\%} 
& Wanda-sp
& \xmark 
& 23.26 & 50.55 & 629.09 
& 868.30 & 33.72 
& 75.00 & 822.18
\\

& FLAP
& \xmark 
& 20.06 & 11.48 & 505.42 
& 157.43 & 24.37 
& 16.53 & 58.76
\\

& PP
& \cmark 
& 9.82 & 7.34 & 4.62 
& - & - 
& - & -
\\

& T{\'y}r
& \xmark 
& 12.55 & 10.38 & 5.64 
& 27.47 & 9.22 
& 44.72 & -
\\

\rowcolor{yellow!15}
& Ours
& \cmark 
& 20.57 & 17.97 & 5.63 
& 18.40& 12.46 & 10.43 
&  6.21
\\
\hline

\end{tabular}
\label{tab:wikitext}
\end{table*}

\subsection{Evaluation Metrics}
\label{app:metrics}
Table~\ref{tab:metrics} presents the evaluation metrics for the tasks used in the experiments.
For LLM QA tasks, we consistently used Accuracy as the primary metric.
In the case of LLM Generation tasks, we employed F1 Score for CoQA, Pass@1 for MBPP and HumanEval, and Exact Match for NQ Open and GSM8K.
For VLM tasks, we utilized F1 Score for POPE, Exact Match for OK-VQA, GQA, and ScienceQA, and Perception Score for MME.
To calculate the mean value for each category, we standardized all metrics to a 100-point scale.
Since Accuracy, F1 Score, Pass@1, and Exact Match have a maximum value of 1, they were converted to percentages.
For MME, which uses Perception Score, we normalized the score based on the total possible score corresponding to the dataset size.
Finally, we computed the macro average of these normalized scores to derive the mean value for each category.
\begin{table}[h]
    \centering
    \caption{Evaluation Metrics and Tasks used in Experiments}
    \label{tab:metrics}
    \footnotesize
    \renewcommand{\arraystretch}{1.2}
    \begin{tabular}{l | l | c}
        \toprule
        \textbf{Category} & \textbf{Task} & \textbf{Metric} \\
        \midrule
        \multirow{7}{*}{LLM QA Tasks} 
        & BoolQ & Accuracy \\
        & RTE & Accuracy \\
        & HellaSwag & Accuracy \\
        & Winogrande & Accuracy \\
        & ARC-Easy & Accuracy \\
        & ARC-Challenge & Accuracy \\
        & OpenBookQA & Accuracy \\
        \midrule
        \multirow{5}{*}{LLM Generation Tasks} 
        & CoQA & F1 Score \\
        & MBPP & Pass@1 \\
        & NQ Open & Exact Match \\
        & HumanEval & Pass@1 \\
        & GSM8K & Exact Match \\
        \midrule
        \multirow{1}{*}{Language Modeling} 
        & WikiText & Perplexity \\
        \midrule
        \multirow{5}{*}{VLM Tasks} 
        & POPE & F1 Score \\
        & OK-VQA & Exact Match \\
        & GQA & Exact Match \\
        & ScienceQA & Exact Match \\
        & MME & Perception Score \\
        \bottomrule
    \end{tabular}
\end{table}

\section{Segment Anything Model}
\label{app:sam}

The Segment Anything Model (SAM)~\cite{kirillov2023segment} comprises a heavy image encoder, a prompt encoder, and a lightweight mask decoder. Notably, the image encoder accounts for over 90\% of the parameters and dominates inference costs, making it a prime candidate for pruning-based acceleration. To validate the extensibility of \paperN~to this pure vision architecture, we adapted our method to target the image encoder. Unlike autoregressive models that utilize a token decoding loop, SAM operates via a single-pass representation; consequently, we applied \paperN~exclusively using the strategy designed for the \textit{prefilling phase}. Given the lack of established structured pruning baselines for this architecture, we implemented a \textbf{Random structured pruning} baseline for comparison.

We evaluated zero-shot segmentation performance on the full validation sets to ensure statistical significance: 5,000 images for COCO~\cite{lin2014microsoft} and 19,809 for LVIS~\cite{gupta2019lvis}. All experiments were conducted with a batch size of 10. To generate box prompts, we adopted the \textbf{generators employed in HQ-SAM}~\cite{ke2023segment}: FocalNet-DINO~\cite{zhang2022dino} for COCO and ViTDet-H~\cite{li2022exploring} for LVIS. The detailed experimental results are presented in Tables~\ref{tab:appendix_coco} and \ref{tab:appendix_lvis}. These results demonstrate that \paperN~effectively maintains performance compared to dense and random baselines, confirming its efficacy in pure vision architectures.

\section{Additional Experimental Results}
\label{app:results}
In this section, we present detailed experimental results stratified by task. Our evaluation encompasses dense models, \paperN~models, and various baseline methods.

Tables~\ref{tab:llama2-QA}, \ref{tab:llama3-QA}, \ref{tab:qwen3-QA}, and \ref{tab:moe-QA} report the performance on zero-shot commonsense question-answering (QA) benchmarks for Large Language Models (LLMs) and Mixture-of-Experts (MoE) models. For text generation tasks, the results across LLMs and MoE models are provided in Tables~\ref{tab:llama2-GEN}, \ref{tab:llama3-GEN}, \ref{tab:qwen3-Gen}, and \ref{tab:moe-Gen}. We also evaluate the language modeling capability of the models. 

Moreover, Table~\ref{tab:vlm-QA} illustrates the experimental outcomes for Visual Question Answering (VQA) tasks using Vision-Language Models (VLMs).

Finally, Tables~\ref{tab:strategy}, \ref{tab:partition-frac-qa}, and \ref{tab:partition-frac-gen} present the detailed results corresponding to the ablation study discussed in Section~\ref{sec:ablation}.
Table~\ref{tab:strategy} compares the performance of models under different pruning strategies.
Tables~\ref{tab:partition-frac-qa} and \ref{tab:partition-frac-gen} demonstrate the impact of the partition fraction, which is a hyperparameter of \paperN, on model accuracy for QA tasks and generative tasks, respectively.
\begin{table*}[t]
\centering
\caption{Detailed Accuracies (\%) on COCO dataset~\cite{lin2014microsoft} using SAM models with different levels of structured sparsity. We compare the dense model with POP and Random structured pruning at 20\% sparsity. For the COCO dataset, we use a SOTA detector FocalNet-DINO~\cite{zhang2022dino} trained on the COCO dataset as our box prompt generator. Bold indicates the better performance between POP and Random.}
\label{tab:appendix_coco}
\vspace{0.1cm}

\small
\renewcommand{\arraystretch}{1.2}
\setlength{\tabcolsep}{8pt}

\definecolor{avg_gray}{gray}{0.95}

\begin{tabular}{l|l c| cccccc >{\columncolor{avg_gray}}c}
\toprule
\textbf{Model} & \textbf{Method} & \textbf{PR (\%)} & \textbf{AP} & \textbf{AP$_{50}$} & \textbf{AP$_{75}$} & \textbf{AP$_s$} & \textbf{AP$_m$} & \textbf{AP$_l$} & \textbf{AP$_{avg}$} \\
\midrule

\multirow{3}{*}{\textbf{ViT-B}}
  & Dense & -- & 46.16 & 73.17 & 49.21 & 34.61 & 50.52 & 59.32 & 52.16 \\ 
  \cline{2-10}
  & Random & 20 & 44.31 & 71.76 & 46.36 & 32.02 & 48.66 & 57.86 & 50.16 \\
  & \textbf{POP (Ours)} & 20 & \textbf{45.38} & \textbf{72.79} & \textbf{47.94} & \textbf{34.06} & \textbf{49.57} & \textbf{58.32} & \textbf{51.34} \\
\midrule

\multirow{3}{*}{\textbf{ViT-L}}
  & Dense & -- & 48.72 & 74.97 & 52.94 & 36.85 & 53.47 & 61.90 & 54.81 \\ 
  \cline{2-10}
  & Random & 20 & 47.18 & 73.66 & \textbf{50.70} & 35.01 & \textbf{51.96} & \textbf{60.55} & 53.18 \\
  & \textbf{POP (Ours)} & 20 & \textbf{47.23} & \textbf{74.14} & 50.68 & \textbf{35.88} & 51.71 & 59.93 & \textbf{53.26} \\
\midrule

\multirow{3}{*}{\textbf{ViT-H}}
  & Dense & -- & 49.03 & 75.20 & 53.23 & 37.13 & 54.14 & 61.78 & 55.09 \\ 
  \cline{2-10}
  & Random & 20 & \textbf{47.70} & 73.98 & \textbf{51.68} & 35.52 & \textbf{52.82} & \textbf{61.25} & \textbf{53.82} \\
  & \textbf{POP (Ours)} & 20 & 47.56 & \textbf{74.23} & 50.99 & \textbf{36.09} & 52.08 & 60.32 & 53.54 \\

\bottomrule
\end{tabular}
\end{table*}
\begin{table*}[t]
\centering
\caption{Detailed Accuracies (\%) on LVIS dataset~\cite{gupta2019lvis} using SAM models with different levels of structured sparsity. We compare the dense model with POP and Random structured pruning at 20\% sparsity. For LVIS, we adopt ViTDet-H~\cite{li2022exploring} trained on the LVIS dataset as our box prompt generator. Bold indicates the better performance between POP and Random.}
\label{tab:appendix_lvis}
\vspace{0.1cm}

\small
\renewcommand{\arraystretch}{1.2}
\setlength{\tabcolsep}{5pt}

\definecolor{avg_gray}{gray}{0.95}

\begin{tabular}{l|l c| cccccc ccc >{\columncolor{avg_gray}}c}
\toprule
\textbf{Model} & \textbf{Method} & \textbf{PR (\%)} & \textbf{AP} & \textbf{AP$_{50}$} & \textbf{AP$_{75}$} & \textbf{AP$_s$} & \textbf{AP$_m$} & \textbf{AP$_l$} & \textbf{AP$_r$} & \textbf{AP$_c$} & \textbf{AP$_f$} & \textbf{AP$_{avg}$} \\
\midrule

\multirow{3}{*}{\textbf{ViT-B}}
  & Dense & -- & 41.41 & 59.61 & 43.68 & 29.47 & 53.49 & 61.39 & 32.96 & 42.27 & 44.18 & 45.38 \\ 
  \cline{2-13}
  & Random & 20 & 39.20 & 58.00 & 40.90 & 27.61 & 50.65 & 59.10 & 31.05 & 40.50 & 41.34 & 43.15 \\
  & \textbf{POP (Ours)} & 20 & \textbf{41.07} & \textbf{59.28} & \textbf{43.51} & \textbf{29.56} & \textbf{52.87} & \textbf{60.49} & \textbf{32.60} & \textbf{41.99} & \textbf{43.76} & \textbf{45.02} \\
\midrule

\multirow{3}{*}{\textbf{ViT-L}}
  & Dense & -- & 44.01 & 60.78 & 46.63 & 31.48 & 56.77 & 65.46 & 33.84 & 44.94 & 47.44 & 47.93 \\ 
  \cline{2-13}
  & Random & 20 & 42.26 & 59.72 & 44.56 & 29.70 & 54.79 & \textbf{64.15} & 32.75 & 43.31 & 45.26 & 46.28 \\
  & \textbf{POP (Ours)} & 20 & \textbf{43.34} & \textbf{60.58} & \textbf{46.03} & \textbf{31.83} & \textbf{55.43} & 63.01 & \textbf{33.87} & \textbf{44.20} & \textbf{46.54} & \textbf{47.20} \\
\midrule

\multirow{3}{*}{\textbf{ViT-H}}
  & Dense & -- & 44.65 & 61.15 & 47.38 & 32.09 & 57.54 & 65.68 & 34.61 & 45.43 & 48.19 & 48.53 \\ 
  \cline{2-13}
  & Random & 20 & 42.82 & 60.06 & 45.37 & 30.23 & 55.26 & \textbf{64.07} & 34.10 & 43.55 & 45.86 & 46.81 \\
  & \textbf{POP (Ours)} & 20 & \textbf{43.66} & \textbf{60.68} & \textbf{46.37} & \textbf{31.76} & \textbf{55.77} & 63.87 & \textbf{34.14} & \textbf{44.31} & \textbf{47.13} & \textbf{47.52} \\

\bottomrule
\end{tabular}
\end{table*}

\def\LTSDenseQA{77.71,62.82,57.17,69.14,76.26,43.43,31.40}
\def\LTSWandaTwentyQA{68.53,53.79,54.74,67.25,72.18,39.85,28.80}
\def\LTSFLAPTwentyQA{68.90,61.01,51.89,66.06,66.50,36.35,29.00}
\def\LTSPPTwentyQA{72.51,61.01,54.75,66.61,73.99,42.07,31.60}
\def\LTSTyrTwentyQA{75.60,62.46,53.94,68.35,74.71,41.30,30.40}
\def\LTSOursTwentyQA{72.63,63.54,53.42,67.72,71.13,39.59,29.40}
\def\LTSWandaFortyQA{61.74,54.51,35.08,51.14,49.37,23.12,14.80}
\def\LTSFLAPFortyQA{63.39,50.54,42.92,62.59,49.03,27.05,24.00}
\def\LTSPPFortyQA{64.43,45.85,46.68,58.72,66.25,34.98,28.80}
\def\LTSTyrFortyQA{69.27,49.46,46.44,64.01,68.98,37.12,28.80}
\def\LTSOursFortyQA{64.04,50.18,37.71,57.46,50.76,27.73,19.80}
\def\LTOSDenseQA{80.55,65.34,60.05,72.14,79.42,48.29,35.20}
\def\LTOSWandaTwentyQA{67.46,54.15,48.56,67.48,69.74,37.54,27.80}
\def\LTOSFLAPTwentyQA{73.49,60.65,55.98,71.90,71.13,42.15,29.80}
\def\LTOSPPTwentyQA{78.78,59.21,58.45,70.32,77.32,44.63,33.80}
\def\LTOSTyrTwentyQA{79.82,68.59,58.01,73.01,77.65,46.16,33.40}
\def\LTOSOursTwentyQA{79.94,67.87,57.68,69.61,74.29,43.00,30.60}
\def\LTOSWandaFortyQA{62.17,52.71,30.07,51.46,31.52,20.05,11.40}
\def\LTOSFLAPFortyQA{63.43,53.07,42.00,61.64,55.14,31.06,26.80}
\def\LTOSPPFortyQA{70.40,49.82,52.77,62.51,71.25,40.44,31.80}
\def\LTOSTyrFortyQA{76.91,61.37,53.03,68.82,73.86,40.36,30.80}
\def\LTOSOursFortyQA{68.23,56.32,42.32,60.62,55.51,32.42,21.60}

\def\LTSZDenseQA{83.76,67.87,64.78,77.98,82.70,54.44,37.20}
\def\LTSZWandaTwentyQA{37.83,51.99,26.84,49.49,29.42,19.71,13.00}
\def\LTSZFLAPTwentyQA{50.55,49.46,30.01,48.38,34.55,20.05,14.40}
\def\LTSZPPTwentyQA{82.26,69.68,64.04,78.14,82.41,55.29,36.40}
\def\LTSZTyrTwentyQA{77.10,69.68,63.80,75.85,81.31,54.35,37.00}
\def\LTSZOursTwentyQA{82.94,66.07,63.92,76.48,81.69,54.18,36.20}
\def\LTSZWandaFortyQA{38.23,54.87,26.24,49.80,27.23,21.42,11.60}
\def\LTSZFLAPFortyQA{39.63,54.87,26.49,49.80,26.05,20.99,13.20}
\def\LTSZPPFortyQA{80.58,63.90,61.42,72.77,78.70,51.62,33.60}
\def\LTSZTyrFortyQA{76.15,69.31,60.63,76.56,78.49,49.15,35.80}
\def\LTSZOursFortyQA{82.60,64.98,60.31,73.72,77.74,46.08,32.60}

\begin{table}[t]
\centering
\caption{Detailed Accuracies (\%) of Llama-2 \cite{touvron2023llama} for 7 QA tasks with different levels of structured sparsity.}
\label{tab:llama2-QA}

\setlength{\tabcolsep}{6pt}
\fontsize{8}{9}\selectfont

\begin{tabular}{l|lc|*{8}{c}}
\toprule
Params & Method & PR(\%) & BoolQ & RTE & HS & WG & ARC-e & ARC-c & OBQA & Mean \\
\midrule

\multirow{11}{*}{Llama2-7B}

& Dense & -
& \GetQA{\LTSDenseQA}{1}
& \GetQA{\LTSDenseQA}{2}
& \GetQA{\LTSDenseQA}{3}
& \GetQA{\LTSDenseQA}{4}
& \GetQA{\LTSDenseQA}{5}
& \GetQA{\LTSDenseQA}{6}
& \GetQA{\LTSDenseQA}{7}
& \QAMean{\LTSDenseQA} \\
\cline{2-11}

\cline{2-11}

& Wanda-sp & 20
& \GetQA{\LTSWandaTwentyQA}{1}
& \GetQA{\LTSWandaTwentyQA}{2}
& \sed \GetQA{\LTSWandaTwentyQA}{3}
& \thd \GetQA{\LTSWandaTwentyQA}{4}
& \thd \GetQA{\LTSWandaTwentyQA}{5}
& \thd \GetQA{\LTSWandaTwentyQA}{6}
& \GetQA{\LTSWandaTwentyQA}{7}
& \QAMean{\LTSWandaTwentyQA} \\

& FLAP & 20
& \GetQA{\LTSFLAPTwentyQA}{1}
& \thd \GetQA{\LTSFLAPTwentyQA}{2}
& \GetQA{\LTSFLAPTwentyQA}{3}
& \GetQA{\LTSFLAPTwentyQA}{4}
& \GetQA{\LTSFLAPTwentyQA}{5}
& \GetQA{\LTSFLAPTwentyQA}{6}
& \GetQA{\LTSFLAPTwentyQA}{7}
& \QAMean{\LTSFLAPTwentyQA} \\

& Probe Pruning & 20
& \thd \GetQA{\LTSPPTwentyQA}{1}
& \thd \GetQA{\LTSPPTwentyQA}{2}
& \fst \GetQA{\LTSPPTwentyQA}{3}
& \GetQA{\LTSPPTwentyQA}{4}
& \sed \GetQA{\LTSPPTwentyQA}{5}
& \fst \GetQA{\LTSPPTwentyQA}{6}
& \fst \GetQA{\LTSPPTwentyQA}{7}
& \sed \QAMean{\LTSPPTwentyQA} \\

& T{\'y}r & 20
& \fst \GetQA{\LTSTyrTwentyQA}{1}
& \sed \GetQA{\LTSTyrTwentyQA}{2}
& \thd \GetQA{\LTSTyrTwentyQA}{3}
& \fst \GetQA{\LTSTyrTwentyQA}{4}
& \fst \GetQA{\LTSTyrTwentyQA}{5}
& \sed \GetQA{\LTSTyrTwentyQA}{6}
& \sed \GetQA{\LTSTyrTwentyQA}{7}
& \fst \QAMean{\LTSTyrTwentyQA} \\

& \textbf{\paperN(Ours)} & 20
& \sed \GetQA{\LTSOursTwentyQA}{1}
& \fst \GetQA{\LTSOursTwentyQA}{2}
& \GetQA{\LTSOursTwentyQA}{3}
& \sed \GetQA{\LTSOursTwentyQA}{4}
& \GetQA{\LTSOursTwentyQA}{5}
& \GetQA{\LTSOursTwentyQA}{6}
& \thd \GetQA{\LTSOursTwentyQA}{7}
& \thd \QAMean{\LTSOursTwentyQA} \\

\cline{2-11}

& Wanda-sp & 40
& \GetQA{\LTSWandaFortyQA}{1}
& \fst \GetQA{\LTSWandaFortyQA}{2}
& \GetQA{\LTSWandaFortyQA}{3}
& \GetQA{\LTSWandaFortyQA}{4}
& \GetQA{\LTSWandaFortyQA}{5}
& \GetQA{\LTSWandaFortyQA}{6}
& \GetQA{\LTSWandaFortyQA}{7}
& \QAMean{\LTSWandaFortyQA} \\

& FLAP & 40
& \GetQA{\LTSFLAPFortyQA}{1}
& \sed \GetQA{\LTSFLAPFortyQA}{2}
& \thd \GetQA{\LTSFLAPFortyQA}{3}
& \sed \GetQA{\LTSFLAPFortyQA}{4}
& \GetQA{\LTSFLAPFortyQA}{5}
& \GetQA{\LTSFLAPFortyQA}{6}
& \sed \GetQA{\LTSFLAPFortyQA}{7}
& \thd \QAMean{\LTSFLAPFortyQA} \\

& Probe Pruning & 40
& \sed \GetQA{\LTSPPFortyQA}{1}
& \GetQA{\LTSPPFortyQA}{2}
& \fst \GetQA{\LTSPPFortyQA}{3}
& \thd \GetQA{\LTSPPFortyQA}{4}
& \sed \GetQA{\LTSPPFortyQA}{5}
& \sed \GetQA{\LTSPPFortyQA}{6}
& \fst \GetQA{\LTSPPFortyQA}{7}
& \sed \QAMean{\LTSPPFortyQA} \\

& T{\'y}r & 40
& \fst \GetQA{\LTSTyrFortyQA}{1}
& \GetQA{\LTSTyrFortyQA}{2}
& \sed \GetQA{\LTSTyrFortyQA}{3}
& \fst \GetQA{\LTSTyrFortyQA}{4}
& \fst \GetQA{\LTSTyrFortyQA}{5}
& \fst \GetQA{\LTSTyrFortyQA}{6}
& \fst \GetQA{\LTSTyrFortyQA}{7}
& \fst \QAMean{\LTSTyrFortyQA} \\

& \textbf{\paperN(Ours)} & 40
& \thd \GetQA{\LTSOursFortyQA}{1}
& \thd \GetQA{\LTSOursFortyQA}{2}
& \GetQA{\LTSOursFortyQA}{3}
& \GetQA{\LTSOursFortyQA}{4}
& \thd \GetQA{\LTSOursFortyQA}{5}
& \thd \GetQA{\LTSOursFortyQA}{6}
& \thd \GetQA{\LTSOursFortyQA}{7}
& \QAMean{\LTSOursFortyQA} \\

\midrule

\multirow{11}{*}{Llama2-13B}

& Dense & -
& \GetQA{\LTOSDenseQA}{1}
& \GetQA{\LTOSDenseQA}{2}
& \GetQA{\LTOSDenseQA}{3}
& \GetQA{\LTOSDenseQA}{4}
& \GetQA{\LTOSDenseQA}{5}
& \GetQA{\LTOSDenseQA}{6}
& \GetQA{\LTOSDenseQA}{7}
& \QAMean{\LTOSDenseQA} \\

\cline{2-11}

& Wanda-sp & 20
& \GetQA{\LTOSWandaTwentyQA}{1}
& \GetQA{\LTOSWandaTwentyQA}{2}
& \GetQA{\LTOSWandaTwentyQA}{3}
& \GetQA{\LTOSWandaTwentyQA}{4}
& \GetQA{\LTOSWandaTwentyQA}{5}
& \GetQA{\LTOSWandaTwentyQA}{6}
& \GetQA{\LTOSWandaTwentyQA}{7}
& \QAMean{\LTOSWandaTwentyQA} \\

& FLAP & 20
& \GetQA{\LTOSFLAPTwentyQA}{1}
& \thd \GetQA{\LTOSFLAPTwentyQA}{2}
& \GetQA{\LTOSFLAPTwentyQA}{3}
& \sed \GetQA{\LTOSFLAPTwentyQA}{4}
& \GetQA{\LTOSFLAPTwentyQA}{5}
& \GetQA{\LTOSFLAPTwentyQA}{6}
& \GetQA{\LTOSFLAPTwentyQA}{7}
& \QAMean{\LTOSFLAPTwentyQA} \\

& Probe Pruning & 20
& \thd \GetQA{\LTOSPPTwentyQA}{1}
& \GetQA{\LTOSPPTwentyQA}{2}
& \fst \GetQA{\LTOSPPTwentyQA}{3}
& \thd \GetQA{\LTOSPPTwentyQA}{4}
& \sed \GetQA{\LTOSPPTwentyQA}{5}
& \sed \GetQA{\LTOSPPTwentyQA}{6}
& \fst \GetQA{\LTOSPPTwentyQA}{7}
& \thd \QAMean{\LTOSPPTwentyQA} \\

& T{\'y}r & 20
& \sed \GetQA{\LTOSTyrTwentyQA}{1}
& \fst \GetQA{\LTOSTyrTwentyQA}{2}
& \sed \GetQA{\LTOSTyrTwentyQA}{3}
& \fst \GetQA{\LTOSTyrTwentyQA}{4}
& \fst \GetQA{\LTOSTyrTwentyQA}{5}
& \fst \GetQA{\LTOSTyrTwentyQA}{6}
& \sed \GetQA{\LTOSTyrTwentyQA}{7}
& \fst \QAMean{\LTOSTyrTwentyQA} \\

& \textbf{\paperN(Ours)} & 20
& \fst \GetQA{\LTOSOursTwentyQA}{1}
& \sed \GetQA{\LTOSOursTwentyQA}{2}
& \thd \GetQA{\LTOSOursTwentyQA}{3}
& \GetQA{\LTOSOursTwentyQA}{4}
& \thd \GetQA{\LTOSOursTwentyQA}{5}
& \thd \GetQA{\LTOSOursTwentyQA}{6}
& \thd \GetQA{\LTOSOursTwentyQA}{7}
& \sed \QAMean{\LTOSOursTwentyQA} \\

\cline{2-11}

& Wanda-sp & 40
& \GetQA{\LTOSWandaFortyQA}{1}
& \GetQA{\LTOSWandaFortyQA}{2}
& \GetQA{\LTOSWandaFortyQA}{3}
& \GetQA{\LTOSWandaFortyQA}{4}
& \GetQA{\LTOSWandaFortyQA}{5}
& \GetQA{\LTOSWandaFortyQA}{6}
& \GetQA{\LTOSWandaFortyQA}{7}
& \QAMean{\LTOSWandaFortyQA} \\

& FLAP & 40
& \GetQA{\LTOSFLAPFortyQA}{1}
& \thd \GetQA{\LTOSFLAPFortyQA}{2}
& \GetQA{\LTOSFLAPFortyQA}{3}
& \thd \GetQA{\LTOSFLAPFortyQA}{4}
& \GetQA{\LTOSFLAPFortyQA}{5}
& \GetQA{\LTOSFLAPFortyQA}{6}
& \thd \GetQA{\LTOSFLAPFortyQA}{7}
& \QAMean{\LTOSFLAPFortyQA} \\

& Probe Pruning & 40
& \sed \GetQA{\LTOSPPFortyQA}{1}
& \GetQA{\LTOSPPFortyQA}{2}
& \sed \GetQA{\LTOSPPFortyQA}{3}
& \sed \GetQA{\LTOSPPFortyQA}{4}
& \sed \GetQA{\LTOSPPFortyQA}{5}
& \fst \GetQA{\LTOSPPFortyQA}{6}
& \fst \GetQA{\LTOSPPFortyQA}{7}
& \sed \QAMean{\LTOSPPFortyQA} \\

& T{\'y}r & 40
& \fst \GetQA{\LTOSTyrFortyQA}{1}
& \fst \GetQA{\LTOSTyrFortyQA}{2}
& \fst \GetQA{\LTOSTyrFortyQA}{3}
& \fst \GetQA{\LTOSTyrFortyQA}{4}
& \fst \GetQA{\LTOSTyrFortyQA}{5}
& \sed \GetQA{\LTOSTyrFortyQA}{6}
& \sed \GetQA{\LTOSTyrFortyQA}{7}
& \fst \QAMean{\LTOSTyrFortyQA} \\

& \textbf{\paperN(Ours)} & 40
& \thd \GetQA{\LTOSOursFortyQA}{1}
& \sed \GetQA{\LTOSOursFortyQA}{2}
& \thd \GetQA{\LTOSOursFortyQA}{3}
& \GetQA{\LTOSOursFortyQA}{4}
& \thd \GetQA{\LTOSOursFortyQA}{5}
& \thd \GetQA{\LTOSOursFortyQA}{6}
& \GetQA{\LTOSOursFortyQA}{7}
& \thd \QAMean{\LTOSOursFortyQA} \\

\midrule

\multirow{11}{*}{Llama2-70B}

& Dense & -
& \GetQA{\LTSZDenseQA}{1}
& \GetQA{\LTSZDenseQA}{2}
& \GetQA{\LTSZDenseQA}{3}
& \GetQA{\LTSZDenseQA}{4}
& \GetQA{\LTSZDenseQA}{5}
& \GetQA{\LTSZDenseQA}{6}
& \GetQA{\LTSZDenseQA}{7}
& \QAMean{\LTSZDenseQA} \\

\cline{2-11}

& Wanda-sp & 20
& \GetQA{\LTSZWandaTwentyQA}{1}
& \thd \GetQA{\LTSZWandaTwentyQA}{2}
& \GetQA{\LTSZWandaTwentyQA}{3}
& \GetQA{\LTSZWandaTwentyQA}{4}
& \GetQA{\LTSZWandaTwentyQA}{5}
& \GetQA{\LTSZWandaTwentyQA}{6}
& \GetQA{\LTSZWandaTwentyQA}{7}
& \QAMean{\LTSZWandaTwentyQA} \\

& FLAP & 20
& \GetQA{\LTSZFLAPTwentyQA}{1}
& \GetQA{\LTSZFLAPTwentyQA}{2}
& \GetQA{\LTSZFLAPTwentyQA}{3}
& \GetQA{\LTSZFLAPTwentyQA}{4}
& \GetQA{\LTSZFLAPTwentyQA}{5}
& \GetQA{\LTSZFLAPTwentyQA}{6}
& \GetQA{\LTSZFLAPTwentyQA}{7}
& \QAMean{\LTSZFLAPTwentyQA} \\

& Probe Pruning & 20
& \sed \GetQA{\LTSZPPTwentyQA}{1}
& \fst \GetQA{\LTSZPPTwentyQA}{2}
& \fst \GetQA{\LTSZPPTwentyQA}{3}
& \fst \GetQA{\LTSZPPTwentyQA}{4}
& \fst \GetQA{\LTSZPPTwentyQA}{5}
& \fst \GetQA{\LTSZPPTwentyQA}{6}
& \sed \GetQA{\LTSZPPTwentyQA}{7}
& \fst \QAMean{\LTSZPPTwentyQA} \\

& T{\'y}r & 20
& \thd \GetQA{\LTSZTyrTwentyQA}{1}
& \fst \GetQA{\LTSZTyrTwentyQA}{2}
& \thd \GetQA{\LTSZTyrTwentyQA}{3}
& \thd \GetQA{\LTSZTyrTwentyQA}{4}
& \thd \GetQA{\LTSZTyrTwentyQA}{5}
& \sed \GetQA{\LTSZTyrTwentyQA}{6}
& \fst \GetQA{\LTSZTyrTwentyQA}{7}
& \thd \QAMean{\LTSZTyrTwentyQA} \\

& \textbf{\paperN(Ours)} & 20
& \fst \GetQA{\LTSZOursTwentyQA}{1}
& \sed \GetQA{\LTSZOursTwentyQA}{2}
& \sed \GetQA{\LTSZOursTwentyQA}{3}
& \sed \GetQA{\LTSZOursTwentyQA}{4}
& \sed \GetQA{\LTSZOursTwentyQA}{5}
& \thd \GetQA{\LTSZOursTwentyQA}{6}
& \thd \GetQA{\LTSZOursTwentyQA}{7}
& \sed \QAMean{\LTSZOursTwentyQA} \\

\cline{2-11}

& Wanda-sp & 40
& \GetQA{\LTSZWandaFortyQA}{1}
& \GetQA{\LTSZWandaFortyQA}{2}
& \GetQA{\LTSZWandaFortyQA}{3}
& \GetQA{\LTSZWandaFortyQA}{4}
& \GetQA{\LTSZWandaFortyQA}{5}
& \GetQA{\LTSZWandaFortyQA}{6}
& \GetQA{\LTSZWandaFortyQA}{7}
& \QAMean{\LTSZWandaFortyQA} \\

& FLAP & 40
& \GetQA{\LTSZFLAPFortyQA}{1}
& \GetQA{\LTSZFLAPFortyQA}{2}
& \GetQA{\LTSZFLAPFortyQA}{3}
& \GetQA{\LTSZFLAPFortyQA}{4}
& \GetQA{\LTSZFLAPFortyQA}{5}
& \GetQA{\LTSZFLAPFortyQA}{6}
& \GetQA{\LTSZFLAPFortyQA}{7}
& \QAMean{\LTSZFLAPFortyQA} \\

& Probe Pruning & 40
& \sed \GetQA{\LTSZPPFortyQA}{1}
& \thd \GetQA{\LTSZPPFortyQA}{2}
& \fst \GetQA{\LTSZPPFortyQA}{3}
& \thd \GetQA{\LTSZPPFortyQA}{4}
& \fst \GetQA{\LTSZPPFortyQA}{5}
& \fst \GetQA{\LTSZPPFortyQA}{6}
& \sed \GetQA{\LTSZPPFortyQA}{7}
& \sed \QAMean{\LTSZPPFortyQA} \\

& T{\'y}r & 40
& \thd \GetQA{\LTSZTyrFortyQA}{1}
& \fst \GetQA{\LTSZTyrFortyQA}{2}
& \sed \GetQA{\LTSZTyrFortyQA}{3}
& \fst \GetQA{\LTSZTyrFortyQA}{4}
& \sed \GetQA{\LTSZTyrFortyQA}{5}
& \sed \GetQA{\LTSZTyrFortyQA}{6}
& \fst \GetQA{\LTSZTyrFortyQA}{7}
& \fst \QAMean{\LTSZTyrFortyQA} \\

& \textbf{\paperN(Ours)} & 40
& \fst \GetQA{\LTSZOursFortyQA}{1}
& \sed \GetQA{\LTSZOursFortyQA}{2}
& \thd \GetQA{\LTSZOursFortyQA}{3}
& \sed \GetQA{\LTSZOursFortyQA}{4}
& \thd \GetQA{\LTSZOursFortyQA}{5}
& \thd \GetQA{\LTSZOursFortyQA}{6}
& \thd \GetQA{\LTSZOursFortyQA}{7}
& \thd \QAMean{\LTSZOursFortyQA} \\

\bottomrule
\end{tabular}
\end{table}

\def\LTOEDense@QA{81.96,70.76,60.01,73.72,81.48,51.45,33.40}
\def\LTOEWandaTwenty@QA{46.12,54.15,26.62,49.41,29.29,19.11,13.40}
\def\LTOEFLAPTwenty@QA{68.62,61.37,48.26,68.27,65.15,32.85,26.60}
\def\LTOEPPTwenty@QA{67.92,53.79,53.02,63.30,73.36,40.44,28.60}
\def\LTOETyrTwenty@QA{78.20,65.34,54.31,68.43,76.18,44.71,31.80}
\def\LTOEOursTwenty@QA{77.28,63.90,54.68,70.88,76.22,44.20,30.20}

\def\LTOEWandaForty@QA{45.75,50.54,25.63,49.96,28.20,20.65,14.80}
\def\LTOEFLAPForty@QA{62.05,57.76,37.35,58.41,44.91,20.90,20.80}
\def\LTOEPPForty@QA{58.62,53.07,33.78,54.85,55.98,25.51,19.80}
\def\LTOETyrForty@QA{71.35,59.93,44.27,61.80,68.01,33.11,25.60}
\def\LTOEOursForty@QA{66.30,56.68,40.44,62.75,61.95,35.07,21.60}

\def\LTOSZDense@QA{85.38,69.68,66.43,79.24,87.21,60.92,37.20}
\def\LTOSZWandaTwenty@QA{38.59,51.99,25.88,51.14,25.46,20.22,13.80}
\def\LTOSZFLAPTwenty@QA{57.03,53.07,29.53,53.67,36.28,19.88,20.40}
\def\LTOSZPPTwenty@QA{85.60,66.07,63.82,78.30,84.72,58.28,34.80}
\def\LTOSZTyrTwenty@QA{85.72,69.31,64.67,79.16,85.10,58.36,35.00}
\def\LTOSZOursTwenty@QA{84.25,68.95,65.32,78.14,85.86,58.36,34.40}

\def\LTOSZWandaForty@QA{37.86,53.79,26.00,47.51,24.66,21.93,11.60}
\def\LTOSZFLAPForty@QA{44.92,52.71,27.93,49.57,31.73,18.60,15.40}
\def\LTOSZTyrForty@QA{84.07,70.76,60.64,77.03,81.48,53.75,33.60}
\def\LTOSZPPForty@QA{77.98,62.46,57.10,70.48,78.24,45.48,28.80}
\def\LTOSZOursForty@QA{78.78,67.15,58.81,72.46,80.09,50.85,29.60}


\begin{table}[t]
\centering
\caption{Detailed Accuracies (\%) of Llama-3 \cite{dubey2024llama} for 7 QA tasks with different levels of structured sparsity.}
\label{tab:llama3-QA}

\setlength{\tabcolsep}{6pt}
\fontsize{8}{9}\selectfont

\begin{tabular}{l|lc|*{8}{c}}
\toprule
Params & Method & PR(\%) & BoolQ & RTE & HS & WG & ARC-e & ARC-c & OBQA & Mean \\
\midrule

\multirow{11}{*}{Llama3.1-8B}

& Dense & -
& \GetQA{\LTOEDense@QA}{1}
& \GetQA{\LTOEDense@QA}{2}
& \GetQA{\LTOEDense@QA}{3}
& \GetQA{\LTOEDense@QA}{4}
& \GetQA{\LTOEDense@QA}{5}
& \GetQA{\LTOEDense@QA}{6}
& \GetQA{\LTOEDense@QA}{7}
& \QAMean{\LTOEDense@QA} \\

\cline{2-11}

& Wanda-sp & 20
& \GetQA{\LTOEWandaTwenty@QA}{1}
& \GetQA{\LTOEWandaTwenty@QA}{2}
& \GetQA{\LTOEWandaTwenty@QA}{3}
& \GetQA{\LTOEWandaTwenty@QA}{4}
& \GetQA{\LTOEWandaTwenty@QA}{5}
& \GetQA{\LTOEWandaTwenty@QA}{6}
& \GetQA{\LTOEWandaTwenty@QA}{7}
& \QAMean{\LTOEWandaTwenty@QA} \\

& FLAP & 20
& \thd\GetQA{\LTOEFLAPTwenty@QA}{1}
& \thd\GetQA{\LTOEFLAPTwenty@QA}{2}
& \GetQA{\LTOEFLAPTwenty@QA}{3}
& \thd\GetQA{\LTOEFLAPTwenty@QA}{4}
& \GetQA{\LTOEFLAPTwenty@QA}{5}
& \GetQA{\LTOEFLAPTwenty@QA}{6}
& \GetQA{\LTOEFLAPTwenty@QA}{7}
& \QAMean{\LTOEFLAPTwenty@QA} \\

& Probe Pruning & 20
& \GetQA{\LTOEPPTwenty@QA}{1}
& \GetQA{\LTOEPPTwenty@QA}{2}
& \thd\GetQA{\LTOEPPTwenty@QA}{3}
& \GetQA{\LTOEPPTwenty@QA}{4}
& \thd\GetQA{\LTOEPPTwenty@QA}{5}
& \thd\GetQA{\LTOEPPTwenty@QA}{6}
& \thd\GetQA{\LTOEPPTwenty@QA}{7}
& \thd\QAMean{\LTOEPPTwenty@QA} \\

& T{\'y}r & 20
& \fst\GetQA{\LTOETyrTwenty@QA}{1}
& \fst\GetQA{\LTOETyrTwenty@QA}{2}
& \sed\GetQA{\LTOETyrTwenty@QA}{3}
& \sed\GetQA{\LTOETyrTwenty@QA}{4}
& \sed\GetQA{\LTOETyrTwenty@QA}{5}
& \fst\GetQA{\LTOETyrTwenty@QA}{6}
& \fst\GetQA{\LTOETyrTwenty@QA}{7}
& \fst\QAMean{\LTOETyrTwenty@QA} \\

& \textbf{\paperN(Ours)} & 20
& \sed\GetQA{\LTOEOursTwenty@QA}{1}
& \sed\GetQA{\LTOEOursTwenty@QA}{2}
& \fst\GetQA{\LTOEOursTwenty@QA}{3}
& \fst\GetQA{\LTOEOursTwenty@QA}{4}
& \fst\GetQA{\LTOEOursTwenty@QA}{5}
& \sed\GetQA{\LTOEOursTwenty@QA}{6}
& \sed\GetQA{\LTOEOursTwenty@QA}{7}
& \sed\QAMean{\LTOEOursTwenty@QA} \\

\cline{2-11}

& Wanda-sp & 40
& \GetQA{\LTOEWandaForty@QA}{1}
& \GetQA{\LTOEWandaForty@QA}{2}
& \GetQA{\LTOEWandaForty@QA}{3}
& \GetQA{\LTOEWandaForty@QA}{4}
& \GetQA{\LTOEWandaForty@QA}{5}
& \GetQA{\LTOEWandaForty@QA}{6}
& \GetQA{\LTOEWandaForty@QA}{7}
& \QAMean{\LTOEWandaForty@QA} \\

& FLAP & 40
& \thd\GetQA{\LTOEFLAPForty@QA}{1}
& \sed\GetQA{\LTOEFLAPForty@QA}{2}
& \thd\GetQA{\LTOEFLAPForty@QA}{3}
& \thd\GetQA{\LTOEFLAPForty@QA}{4}
& \GetQA{\LTOEFLAPForty@QA}{5}
& \GetQA{\LTOEFLAPForty@QA}{6}
& \thd\GetQA{\LTOEFLAPForty@QA}{7}
& \thd\QAMean{\LTOEFLAPForty@QA} \\

& Probe Pruning & 40
& \GetQA{\LTOEPPForty@QA}{1}
& \GetQA{\LTOEPPForty@QA}{2}
& \GetQA{\LTOEPPForty@QA}{3}
& \GetQA{\LTOEPPForty@QA}{4}
& \thd\GetQA{\LTOEPPForty@QA}{5}
& \thd\GetQA{\LTOEPPForty@QA}{6}
& \GetQA{\LTOEPPForty@QA}{7}
& \QAMean{\LTOEPPForty@QA} \\

& T{\'y}r & 40
& \fst\GetQA{\LTOETyrForty@QA}{1}
& \fst\GetQA{\LTOETyrForty@QA}{2}
& \fst\GetQA{\LTOETyrForty@QA}{3}
& \sed\GetQA{\LTOETyrForty@QA}{4}
& \fst\GetQA{\LTOETyrForty@QA}{5}
& \sed\GetQA{\LTOETyrForty@QA}{6}
& \fst\GetQA{\LTOETyrForty@QA}{7}
& \fst\QAMean{\LTOETyrForty@QA} \\

& \textbf{\paperN(Ours)} & 40 
& \sed\GetQA{\LTOEOursForty@QA}{1}
& \thd\GetQA{\LTOEOursForty@QA}{2}
& \sed\GetQA{\LTOEOursForty@QA}{3}
& \fst\GetQA{\LTOEOursForty@QA}{4}
& \sed\GetQA{\LTOEOursForty@QA}{5}
& \fst\GetQA{\LTOEOursForty@QA}{6}
& \sed\GetQA{\LTOEOursForty@QA}{7}
& \sed\QAMean{\LTOEOursForty@QA} \\

\midrule

\multirow{11}{*}{Llama3.1-70B}

& Dense & -
& \GetQA{\LTOSZDense@QA}{1}
& \GetQA{\LTOSZDense@QA}{2}
& \GetQA{\LTOSZDense@QA}{3}
& \GetQA{\LTOSZDense@QA}{4}
& \GetQA{\LTOSZDense@QA}{5}
& \GetQA{\LTOSZDense@QA}{6}
& \GetQA{\LTOSZDense@QA}{7}
& \QAMean{\LTOSZDense@QA} \\

\cline{2-11}

& Wanda-sp & 20
& \GetQA{\LTOSZWandaTwenty@QA}{1}
& \GetQA{\LTOSZWandaTwenty@QA}{2}
& \GetQA{\LTOSZWandaTwenty@QA}{3}
& \GetQA{\LTOSZWandaTwenty@QA}{4}
& \GetQA{\LTOSZWandaTwenty@QA}{5}
& \GetQA{\LTOSZWandaTwenty@QA}{6}
& \GetQA{\LTOSZWandaTwenty@QA}{7}
& \QAMean{\LTOSZWandaTwenty@QA} \\

& FLAP & 20
& \GetQA{\LTOSZFLAPTwenty@QA}{1}
& \GetQA{\LTOSZFLAPTwenty@QA}{2}
& \GetQA{\LTOSZFLAPTwenty@QA}{3}
& \GetQA{\LTOSZFLAPTwenty@QA}{4}
& \GetQA{\LTOSZFLAPTwenty@QA}{5}
& \GetQA{\LTOSZFLAPTwenty@QA}{6}
& \GetQA{\LTOSZFLAPTwenty@QA}{7}
& \QAMean{\LTOSZFLAPTwenty@QA} \\

& Probe Pruning & 20
& \sed\GetQA{\LTOSZPPTwenty@QA}{1}
& \thd\GetQA{\LTOSZPPTwenty@QA}{2}
& \thd\GetQA{\LTOSZPPTwenty@QA}{3}
& \sed\GetQA{\LTOSZPPTwenty@QA}{4}
& \thd\GetQA{\LTOSZPPTwenty@QA}{5}
& \sed\GetQA{\LTOSZPPTwenty@QA}{6}
& \sed\GetQA{\LTOSZPPTwenty@QA}{7}
& \thd\QAMean{\LTOSZPPTwenty@QA} \\

& T{\'y}r & 20
& \fst\GetQA{\LTOSZTyrTwenty@QA}{1}
& \fst\GetQA{\LTOSZTyrTwenty@QA}{2}
& \sed\GetQA{\LTOSZTyrTwenty@QA}{3}
& \fst\GetQA{\LTOSZTyrTwenty@QA}{4}
& \sed\GetQA{\LTOSZTyrTwenty@QA}{5}
& \fst\GetQA{\LTOSZTyrTwenty@QA}{6}
& \fst\GetQA{\LTOSZTyrTwenty@QA}{7}
& \fst\QAMean{\LTOSZTyrTwenty@QA} \\

& \textbf{\paperN(Ours)} & 20
& \thd\GetQA{\LTOSZOursTwenty@QA}{1}
& \sed\GetQA{\LTOSZOursTwenty@QA}{2}
& \fst\GetQA{\LTOSZOursTwenty@QA}{3}
& \thd\GetQA{\LTOSZOursTwenty@QA}{4}
& \fst\GetQA{\LTOSZOursTwenty@QA}{5}
& \fst\GetQA{\LTOSZOursTwenty@QA}{6}
& \thd\GetQA{\LTOSZOursTwenty@QA}{7}
& \sed\QAMean{\LTOSZOursTwenty@QA} \\

\cline{2-11}

& Wanda-sp & 40
& \GetQA{\LTOSZWandaForty@QA}{1}
& \GetQA{\LTOSZWandaForty@QA}{2}
& \GetQA{\LTOSZWandaForty@QA}{3}
& \GetQA{\LTOSZWandaForty@QA}{4}
& \GetQA{\LTOSZWandaForty@QA}{5}
& \GetQA{\LTOSZWandaForty@QA}{6}
& \GetQA{\LTOSZWandaForty@QA}{7}
& \QAMean{\LTOSZWandaForty@QA} \\

& FLAP & 40
& \GetQA{\LTOSZFLAPForty@QA}{1}
& \GetQA{\LTOSZFLAPForty@QA}{2}
& \GetQA{\LTOSZFLAPForty@QA}{3}
& \GetQA{\LTOSZFLAPForty@QA}{4}
& \GetQA{\LTOSZFLAPForty@QA}{5}
& \GetQA{\LTOSZFLAPForty@QA}{6}
& \GetQA{\LTOSZFLAPForty@QA}{7}
& \QAMean{\LTOSZFLAPForty@QA} \\

& Probe Pruning & 40
& \thd\GetQA{\LTOSZPPForty@QA}{1}
& \thd\GetQA{\LTOSZPPForty@QA}{2}
& \thd\GetQA{\LTOSZPPForty@QA}{3}
& \thd\GetQA{\LTOSZPPForty@QA}{4}
& \thd\GetQA{\LTOSZPPForty@QA}{5}
& \thd\GetQA{\LTOSZPPForty@QA}{6}
& \thd\GetQA{\LTOSZPPForty@QA}{7}
& \thd\QAMean{\LTOSZPPForty@QA} \\

& T{\'y}r & 40
& \fst\GetQA{\LTOSZTyrForty@QA}{1}
& \fst\GetQA{\LTOSZTyrForty@QA}{2}
& \fst\GetQA{\LTOSZTyrForty@QA}{3}
& \fst\GetQA{\LTOSZTyrForty@QA}{4}
& \fst\GetQA{\LTOSZTyrForty@QA}{5}
& \fst\GetQA{\LTOSZTyrForty@QA}{6}
& \fst\GetQA{\LTOSZTyrForty@QA}{7}
& \fst\QAMean{\LTOSZTyrForty@QA} \\

& \textbf{\paperN(Ours)} & 40 
& \sed\GetQA{\LTOSZOursForty@QA}{1}
& \sed\GetQA{\LTOSZOursForty@QA}{2}
& \sed\GetQA{\LTOSZOursForty@QA}{3}
& \sed\GetQA{\LTOSZOursForty@QA}{4}
& \sed\GetQA{\LTOSZOursForty@QA}{5}
& \sed\GetQA{\LTOSZOursForty@QA}{6}
& \sed\GetQA{\LTOSZOursForty@QA}{7}
& \sed\QAMean{\LTOSZOursForty@QA} \\

\bottomrule
\end{tabular}
\end{table}

\def\QTEDense@QA{86.61,77.98,57.10,68.11,83.29,55.46,31.60}

\def\QTEWandaTwenty@QA{64.59,53.07,37.64,52.57,54.46,26.62,24.80}
\def\QTEFLAPTwenty@QA{77.52,66.07,50.05,62.90,64.39,35.32,24.80}
\def\QTEPPTwenty@QA{,,,,,,}
\def\QTETyrTwenty@QA{84.62,74.73,54.27,67.56,81.61,52.47,31.40}
\def\QTEOursTwenty@QA{84.77,74.37,51.76,64.80,77.78,48.21,28.80}

\def\QTEWandaForty@QA{48.41,48.74,27.65,50.20,34.39,18.86,13.60}
\def\QTEFLAPForty@QA{57.62,51.63,32.31,51.07,32.28,19.54,12.20}
\def\QTEPPForty@QA{,,,,,,}
\def\QTETyrForty@QA{62.39,67.15,42.98,59.35,73.70,39.51,31.00}
\def\QTEOursForty@QA{76.67,65.70,39.52,57.14,60.31,32.34,25.20}

\def\QTTTDense@QA{86.30,76.17,63.91,73.32,84.43,57.86,36.00}

\def\QTTTWandaTwenty@QA{72.39,59.57,49.21,59.12,63.47,36.86,22.00}
\def\QTTTFLAPTwenty@QA{85.96,79.06,61.13,67.40,25.08,22.70,33.20}
\def\QTTTPPTwenty@QA{,,,,,,}
\def\QTTTTyrTwenty@QA{86.97,78.34,63.94,71.90,83.38,57.00,35.60}
\def\QTTTOursTwenty@QA{85.72,76.53,60.94,65.43,76.85,52.22,32.00}

\def\QTTTWandaForty@QA{59.51,51.26,36.80,51.54,42.72,25.34,18.00}
\def\QTTTFLAPForty@QA{58.10,53.79,39.37,51.22,42.76,23.04,17.00}
\def\QTTTTyrForty@QA{85.75,75.45,61.08,70.64,83.71,57.85,36.00}
\def\QTTTOursForty@QA{80.67,70.40,51.23,58.33,64.52,40.10,26.60}

\begin{table}[t]
\centering
\caption{Detailed Accuracies (\%) of Qwen3~\cite{yang2025qwen3} for 7 QA tasks with different levels of structured sparsity.}
\label{tab:qwen3-QA}

\setlength{\tabcolsep}{6pt}
\fontsize{8}{9}\selectfont

\begin{tabular}{l|lc|*{8}{c}}
\toprule
Params & Method & PR(\%) & BoolQ & RTE & HS & WG & ARC-e & ARC-c & OBQA & Mean \\
\midrule

\multirow{9}{*}{Qwen3-8B}

& Dense & -
& \GetQA{\QTEDense@QA}{1}
& \GetQA{\QTEDense@QA}{2}
& \GetQA{\QTEDense@QA}{3}
& \GetQA{\QTEDense@QA}{4}
& \GetQA{\QTEDense@QA}{5}
& \GetQA{\QTEDense@QA}{6}
& \GetQA{\QTEDense@QA}{7}
& \QAMean{\QTEDense@QA} \\

\cline{2-11}

& Wanda-sp & 20
& \GetQA{\QTEWandaTwenty@QA}{1}
& \GetQA{\QTEWandaTwenty@QA}{2}
& \GetQA{\QTEWandaTwenty@QA}{3}
& \GetQA{\QTEWandaTwenty@QA}{4}
& \GetQA{\QTEWandaTwenty@QA}{5}
& \GetQA{\QTEWandaTwenty@QA}{6}
& \thd\GetQA{\QTEWandaTwenty@QA}{7}
& \QAMean{\QTEWandaTwenty@QA} \\

& FLAP & 20
& \thd\GetQA{\QTEFLAPTwenty@QA}{1}
& \thd\GetQA{\QTEFLAPTwenty@QA}{2}
& \thd\GetQA{\QTEFLAPTwenty@QA}{3}
& \thd\GetQA{\QTEFLAPTwenty@QA}{4}
& \thd\GetQA{\QTEFLAPTwenty@QA}{5}
& \thd\GetQA{\QTEFLAPTwenty@QA}{6}
& \thd\GetQA{\QTEFLAPTwenty@QA}{7}
& \thd\QAMean{\QTEFLAPTwenty@QA} \\

& T{\'y}r & 20
& \sed\GetQA{\QTETyrTwenty@QA}{1}
& \fst\GetQA{\QTETyrTwenty@QA}{2}
& \fst\GetQA{\QTETyrTwenty@QA}{3}
& \fst\GetQA{\QTETyrTwenty@QA}{4}
& \fst\GetQA{\QTETyrTwenty@QA}{5}
& \fst\GetQA{\QTETyrTwenty@QA}{6}
& \fst\GetQA{\QTETyrTwenty@QA}{7}
& \fst\QAMean{\QTETyrTwenty@QA} \\

& \textbf{\paperN(Ours)} & 20 
& \fst\GetQA{\QTEOursTwenty@QA}{1}
& \sed\GetQA{\QTEOursTwenty@QA}{2}
& \sed\GetQA{\QTEOursTwenty@QA}{3}
& \sed\GetQA{\QTEOursTwenty@QA}{4}
& \sed\GetQA{\QTEOursTwenty@QA}{5}
& \sed\GetQA{\QTEOursTwenty@QA}{6}
& \sed\GetQA{\QTEOursTwenty@QA}{7}
& \sed\QAMean{\QTEOursTwenty@QA} \\

\cline{2-11}

& Wanda-sp & 40
& \GetQA{\QTEWandaForty@QA}{1}
& \GetQA{\QTEWandaForty@QA}{2}
& \GetQA{\QTEWandaForty@QA}{3}
& \GetQA{\QTEWandaForty@QA}{4}
& \thd\GetQA{\QTEWandaForty@QA}{5}
& \GetQA{\QTEWandaForty@QA}{6}
& \thd\GetQA{\QTEWandaForty@QA}{7}
& \QAMean{\QTEWandaForty@QA} \\

& FLAP & 40
& \thd\GetQA{\QTEFLAPForty@QA}{1}
& \thd\GetQA{\QTEFLAPForty@QA}{2}
& \thd\GetQA{\QTEFLAPForty@QA}{3}
& \thd\GetQA{\QTEFLAPForty@QA}{4}
& \GetQA{\QTEFLAPForty@QA}{5}
& \thd\GetQA{\QTEFLAPForty@QA}{6}
& \GetQA{\QTEFLAPForty@QA}{7}
& \thd\QAMean{\QTEFLAPForty@QA} \\

& T{\'y}r & 40
& \sed\GetQA{\QTETyrForty@QA}{1}
& \fst\GetQA{\QTETyrForty@QA}{2}
& \fst\GetQA{\QTETyrForty@QA}{3}
& \fst\GetQA{\QTETyrForty@QA}{4}
& \fst\GetQA{\QTETyrForty@QA}{5}
& \fst\GetQA{\QTETyrForty@QA}{6}
& \fst\GetQA{\QTETyrForty@QA}{7}
& \fst\QAMean{\QTETyrForty@QA} \\

& \textbf{\paperN(Ours)} & 40
& \fst\GetQA{\QTEOursForty@QA}{1}
& \sed\GetQA{\QTEOursForty@QA}{2}
& \sed\GetQA{\QTEOursForty@QA}{3}
& \sed\GetQA{\QTEOursForty@QA}{4}
& \sed\GetQA{\QTEOursForty@QA}{5}
& \sed\GetQA{\QTEOursForty@QA}{6}
& \sed\GetQA{\QTEOursForty@QA}{7}
& \sed\QAMean{\QTEOursForty@QA} \\

\midrule

\multirow{9}{*}{Qwen3-32B}

& Dense & -
& \GetQA{\QTTTDense@QA}{1}
& \GetQA{\QTTTDense@QA}{2}
& \GetQA{\QTTTDense@QA}{3}
& \GetQA{\QTTTDense@QA}{4}
& \GetQA{\QTTTDense@QA}{5}
& \GetQA{\QTTTDense@QA}{6}
& \GetQA{\QTTTDense@QA}{7}
& \QAMean{\QTTTDense@QA} \\

\cline{2-11}

& Wanda-sp & 20
& \GetQA{\QTTTWandaTwenty@QA}{1}
& \GetQA{\QTTTWandaTwenty@QA}{2}
& \GetQA{\QTTTWandaTwenty@QA}{3}
& \GetQA{\QTTTWandaTwenty@QA}{4}
& \thd\GetQA{\QTTTWandaTwenty@QA}{5}
& \thd\GetQA{\QTTTWandaTwenty@QA}{6}
& \GetQA{\QTTTWandaTwenty@QA}{7}
& \QAMean{\QTTTWandaTwenty@QA} \\

& FLAP & 20
& \sed\GetQA{\QTTTFLAPTwenty@QA}{1}
& \fst\GetQA{\QTTTFLAPTwenty@QA}{2}
& \sed\GetQA{\QTTTFLAPTwenty@QA}{3}
& \sed\GetQA{\QTTTFLAPTwenty@QA}{4}
& \GetQA{\QTTTFLAPTwenty@QA}{5}
& \GetQA{\QTTTFLAPTwenty@QA}{6}
& \sed\GetQA{\QTTTFLAPTwenty@QA}{7}
& \thd\QAMean{\QTTTFLAPTwenty@QA} \\

& T{\'y}r & 20
& \fst\GetQA{\QTTTTyrTwenty@QA}{1}
& \sed\GetQA{\QTTTTyrTwenty@QA}{2}
& \fst\GetQA{\QTTTTyrTwenty@QA}{3}
& \fst\GetQA{\QTTTTyrTwenty@QA}{4}
& \fst\GetQA{\QTTTTyrTwenty@QA}{5}
& \fst\GetQA{\QTTTTyrTwenty@QA}{6}
& \fst\GetQA{\QTTTTyrTwenty@QA}{7}
& \fst\QAMean{\QTTTTyrTwenty@QA} \\

& \textbf{\paperN(Ours)} & 20 
& \thd\GetQA{\QTTTOursTwenty@QA}{1}
& \thd\GetQA{\QTTTOursTwenty@QA}{2}
& \thd\GetQA{\QTTTOursTwenty@QA}{3}
& \thd\GetQA{\QTTTOursTwenty@QA}{4}
& \sed\GetQA{\QTTTOursTwenty@QA}{5}
& \sed\GetQA{\QTTTOursTwenty@QA}{6}
& \thd\GetQA{\QTTTOursTwenty@QA}{7}
& \sed\QAMean{\QTTTOursTwenty@QA} \\

\cline{2-11}

& Wanda-sp & 40
& \thd\GetQA{\QTTTWandaForty@QA}{1}
& \GetQA{\QTTTWandaForty@QA}{2}
& \GetQA{\QTTTWandaForty@QA}{3}
& \GetQA{\QTTTWandaForty@QA}{4}
& \GetQA{\QTTTWandaForty@QA}{5}
& \thd\GetQA{\QTTTWandaForty@QA}{6}
& \thd\GetQA{\QTTTWandaForty@QA}{7}
& \QAMean{\QTTTWandaForty@QA} \\

& FLAP & 40
& \GetQA{\QTTTFLAPForty@QA}{1}
& \thd\GetQA{\QTTTFLAPForty@QA}{2}
& \thd\GetQA{\QTTTFLAPForty@QA}{3}
& \GetQA{\QTTTFLAPForty@QA}{4}
& \thd\GetQA{\QTTTFLAPForty@QA}{5}
& \GetQA{\QTTTFLAPForty@QA}{6}
& \GetQA{\QTTTFLAPForty@QA}{7}
& \thd\QAMean{\QTTTFLAPForty@QA} \\

& T{\'y}r & 40
& \fst\GetQA{\QTTTTyrForty@QA}{1}
& \fst\GetQA{\QTTTTyrForty@QA}{2}
& \fst\GetQA{\QTTTTyrForty@QA}{3}
& \fst\GetQA{\QTTTTyrForty@QA}{4}
& \fst\GetQA{\QTTTTyrForty@QA}{5}
& \fst\GetQA{\QTTTTyrForty@QA}{6}
& \fst\GetQA{\QTTTTyrForty@QA}{7}
& \fst\QAMean{\QTTTTyrForty@QA} \\

& \textbf{\paperN(Ours)} & 40 
& \sed\GetQA{\QTTTOursForty@QA}{1}
& \sed\GetQA{\QTTTOursForty@QA}{2}
& \sed\GetQA{\QTTTOursForty@QA}{3}
& \sed\GetQA{\QTTTOursForty@QA}{4}
& \sed\GetQA{\QTTTOursForty@QA}{5}
& \sed\GetQA{\QTTTOursForty@QA}{6}
& \sed\GetQA{\QTTTOursForty@QA}{7}
& \sed\QAMean{\QTTTOursForty@QA} \\

\bottomrule
\end{tabular}
\end{table}

\def\LTSGENDense{77.35,23.40,18.89,9.76,14.03}
\def\LTSGENWandaTwenty{51.37,0.00,7.56,0.00,2.05}
\def\LTSGENFLAPTwenty{73.77,2.00,6.21,1.22,2.96}
\def\LTSGENPPTwenty{66.46,8.40,10.80,6.71,7.20}
\def\LTSGENTyrTwenty{75.46,8.20,11.27,1.83,6.98}
\def\LTSGENOursTwenty{76.46,16.20,9.78,7.93,10.24}

\def\LTSGENWandaForty{22.45,0.00,0.28,0.00,1.90}
\def\LTSGENFLAPForty{43.07,0.00,0.42,0.00,1.90}
\def\LTSGENPPForty{38.18,1.40,2.38,0.00,1.44}
\def\LTSGENTyrForty{51.69,0.00,4.64,0.00,0.00}
\def\LTSGENOursForty{56.44,3.80,2.02,3.05,3.87}

\def\LTOTGENDense{79.18,27.20,23.63,12.20,23.12}
\def\LTOTGENWandaTwenty{49.96,0.60,1.66,0.00,2.58}
\def\LTOTGENFLAPTwenty{76.94,3.40,6.54,0.61,7.73}
\def\LTOTGENPPTwenty{76.05,20.80,15.43,10.37,16.15}
\def\LTOTGENTyrTwenty{78.05,11.40,11.63,1.83,16.68}
\def\LTOTGENOursTwenty{79.00,21.60,13.85,9.76,20.62}

\def\LTOTGENWandaForty{10.65,0.00,0.00,0.00,1.14}
\def\LTOTGENFLAPForty{68.49,0.00,1.11,0.00,1.36}
\def\LTOTGENPPForty{52.36,5.20,5.18,3.66,5.38}
\def\LTOTGENTyrForty{74.79,2.00,5.15,0.00,3.49}
\def\LTOTGENOursForty{59.51,10.00,2.91,6.10,11.45}

\def\LTSZGENDense{83.71,45.00,29.00,16.46,53.30}
\def\LTSZGENWandaTwenty{2.68,0.00,0.00,0.00,1.52}
\def\LTSZGENFLAPTwenty{14.13,0.00,0.00,0.00,0.83}
\def\LTSZGENPPTwenty{82.24,39.00,23.30,14.02,50.57}
\def\LTSZGENTyrTwenty{81.66,30.20,24.32,10.98,46.85}
\def\LTSZGENOursTwenty{83.09,38.00,23.82,14.63,50.64}

\def\LTSZGENWandaForty{0.99,0.00,0.00,0.00,0.68}
\def\LTSZGENFLAPForty{0.67,0.00,0.00,0.00,0.30}
\def\LTSZGENPPForty{79.98,25.60,14.07,10.98,34.65}
\def\LTSZGENTyrForty{75.85,0.00,13.24,5.49,30.02}
\def\LTSZGENOursForty{81.71,24.00,11.52,14.02,41.09}

\begin{table}[t]
\centering
\caption{Accuracies (\%) of Llama-2 \cite{touvron2023llama} for 5 Generative tasks with different levels of structured sparsity.}
\label{tab:llama2-GEN}

\setlength{\tabcolsep}{7.5pt}
\renewcommand{\arraystretch}{1.1}
\footnotesize

\begin{tabular}{l|lc|l*{6}{c}}
\toprule
Params & Method & PR
& coqa & mbpp & nq\_open & humaneval & gsm8k
& Mean \\
\midrule

\multirow{11}{*}{Llama2-7B}

& Dense & -
& \GetGen{\LTSGENDense}{1}
& \GetGen{\LTSGENDense}{2}
& \GetGen{\LTSGENDense}{3}
& \GetGen{\LTSGENDense}{4}
& \GetGen{\LTSGENDense}{5}
& \GENMean{\LTSGENDense} \\

\cline{2-9}

& Wanda-sp & 20
& \GetGen{\LTSGENWandaTwenty}{1}
& \GetGen{\LTSGENWandaTwenty}{2}
& \GetGen{\LTSGENWandaTwenty}{3}
& \GetGen{\LTSGENWandaTwenty}{4}
& \GetGen{\LTSGENWandaTwenty}{5}
& \GENMean{\LTSGENWandaTwenty} \\

& FLAP & 20
& \thd\GetGen{\LTSGENFLAPTwenty}{1}
& \GetGen{\LTSGENFLAPTwenty}{2}
& \GetGen{\LTSGENFLAPTwenty}{3}
& \GetGen{\LTSGENFLAPTwenty}{4}
& \GetGen{\LTSGENFLAPTwenty}{5}
& \GENMean{\LTSGENFLAPTwenty} \\

& Probe Pruning & 20
& \GetGen{\LTSGENPPTwenty}{1}
& \sed\GetGen{\LTSGENPPTwenty}{2}
& \sed\GetGen{\LTSGENPPTwenty}{3}
& \sed\GetGen{\LTSGENPPTwenty}{4}
& \sed\GetGen{\LTSGENPPTwenty}{5}
& \thd\GENMean{\LTSGENPPTwenty} \\

& T{\'y}r & 20
& \sed\GetGen{\LTSGENTyrTwenty}{1}
& \thd\GetGen{\LTSGENTyrTwenty}{2}
& \fst\GetGen{\LTSGENTyrTwenty}{3}
& \thd\GetGen{\LTSGENTyrTwenty}{4}
& \thd\GetGen{\LTSGENTyrTwenty}{5}
& \sed\GENMean{\LTSGENTyrTwenty} \\

& \textbf{\paperN(Ours)} & 20
& \fst\GetGen{\LTSGENOursTwenty}{1}
& \fst\GetGen{\LTSGENOursTwenty}{2}
& \thd\GetGen{\LTSGENOursTwenty}{3}
& \fst\GetGen{\LTSGENOursTwenty}{4}
& \fst\GetGen{\LTSGENOursTwenty}{5}
& \fst\GENMean{\LTSGENOursTwenty} \\

\cline{2-9}

& Wanda-sp & 40
& \GetGen{\LTSGENWandaForty}{1}
& \GetGen{\LTSGENWandaForty}{2}
& \GetGen{\LTSGENWandaForty}{3}
& \GetGen{\LTSGENWandaForty}{4}
& \sed\GetGen{\LTSGENWandaForty}{5}
& \GENMean{\LTSGENWandaForty} \\

& FLAP & 40
& \thd\GetGen{\LTSGENFLAPForty}{1}
& \GetGen{\LTSGENFLAPForty}{2}
& \GetGen{\LTSGENFLAPForty}{3}
& \GetGen{\LTSGENFLAPForty}{4}
& \sed\GetGen{\LTSGENFLAPForty}{5}
& \thd\GENMean{\LTSGENFLAPForty} \\

& Probe Pruning & 40
& \GetGen{\LTSGENPPForty}{1}
& \sed\GetGen{\LTSGENPPForty}{2}
& \sed\GetGen{\LTSGENPPForty}{3}
& \GetGen{\LTSGENPPForty}{4}
& \GetGen{\LTSGENPPForty}{5}
& \GENMean{\LTSGENPPForty} \\

& T{\'y}r & 40
& \sed\GetGen{\LTSGENTyrForty}{1}
& \GetGen{\LTSGENTyrForty}{2}
& \fst\GetGen{\LTSGENTyrForty}{3}
& \GetGen{\LTSGENTyrForty}{4}
& \GetGen{\LTSGENTyrForty}{5}
& \sed\GENMean{\LTSGENTyrForty} \\

& \textbf{\paperN(Ours)} & 40 
& \fst\GetGen{\LTSGENOursForty}{1}
& \fst\GetGen{\LTSGENOursForty}{2}
& \thd\GetGen{\LTSGENOursForty}{3}
& \fst\GetGen{\LTSGENOursForty}{4}
& \fst\GetGen{\LTSGENOursForty}{5}
& \fst\GENMean{\LTSGENOursForty} \\

\midrule

\multirow{11}{*}{Llama2-13B}

& Dense & -
& \GetGen{\LTOTGENDense}{1}
& \GetGen{\LTOTGENDense}{2}
& \GetGen{\LTOTGENDense}{3}
& \GetGen{\LTOTGENDense}{4}
& \GetGen{\LTOTGENDense}{5}
& \GENMean{\LTOTGENDense} \\

\cline{2-9}

& Wanda-sp & 20
& \GetGen{\LTOTGENWandaTwenty}{1}
& \GetGen{\LTOTGENWandaTwenty}{2}
& \GetGen{\LTOTGENWandaTwenty}{3}
& \GetGen{\LTOTGENWandaTwenty}{4}
& \GetGen{\LTOTGENWandaTwenty}{5}
& \GENMean{\LTOTGENWandaTwenty} \\

& FLAP & 20
& \thd\GetGen{\LTOTGENFLAPTwenty}{1}
& \GetGen{\LTOTGENFLAPTwenty}{2}
& \GetGen{\LTOTGENFLAPTwenty}{3}
& \GetGen{\LTOTGENFLAPTwenty}{4}
& \GetGen{\LTOTGENFLAPTwenty}{5}
& \GENMean{\LTOTGENFLAPTwenty} \\

& Probe Pruning & 20
& \GetGen{\LTOTGENPPTwenty}{1}
& \sed\GetGen{\LTOTGENPPTwenty}{2}
& \fst\GetGen{\LTOTGENPPTwenty}{3}
& \fst\GetGen{\LTOTGENPPTwenty}{4}
& \thd\GetGen{\LTOTGENPPTwenty}{5}
& \sed\GENMean{\LTOTGENPPTwenty} \\

& T{\'y}r & 20
& \sed\GetGen{\LTOTGENTyrTwenty}{1}
& \thd\GetGen{\LTOTGENTyrTwenty}{2}
& \thd\GetGen{\LTOTGENTyrTwenty}{3}
& \thd\GetGen{\LTOTGENTyrTwenty}{4}
& \sed\GetGen{\LTOTGENTyrTwenty}{5}
& \thd\GENMean{\LTOTGENTyrTwenty} \\

& \textbf{\paperN(Ours)} & 20
& \fst\GetGen{\LTOTGENOursTwenty}{1}
& \fst\GetGen{\LTOTGENOursTwenty}{2}
& \sed\GetGen{\LTOTGENOursTwenty}{3}
& \sed\GetGen{\LTOTGENOursTwenty}{4}
& \fst\GetGen{\LTOTGENOursTwenty}{5}
& \fst\GENMean{\LTOTGENOursTwenty} \\

\cline{2-9}

& Wanda-sp & 40
& \GetGen{\LTOTGENWandaForty}{1}
& \GetGen{\LTOTGENWandaForty}{2}
& \GetGen{\LTOTGENWandaForty}{3}
& \GetGen{\LTOTGENWandaForty}{4}
& \GetGen{\LTOTGENWandaForty}{5}
& \GENMean{\LTOTGENWandaForty} \\

& FLAP & 40
& \sed\GetGen{\LTOTGENFLAPForty}{1}
& \GetGen{\LTOTGENFLAPForty}{2}
& \GetGen{\LTOTGENFLAPForty}{3}
& \GetGen{\LTOTGENFLAPForty}{4}
& \GetGen{\LTOTGENFLAPForty}{5}
& \GENMean{\LTOTGENFLAPForty} \\

& Probe Pruning & 40
& \GetGen{\LTOTGENPPForty}{1}
& \sed\GetGen{\LTOTGENPPForty}{2}
& \fst\GetGen{\LTOTGENPPForty}{3}
& \sed\GetGen{\LTOTGENPPForty}{4}
& \sed\GetGen{\LTOTGENPPForty}{5}
& \thd\GENMean{\LTOTGENPPForty} \\

& T{\'y}r & 40
& \fst\GetGen{\LTOTGENTyrForty}{1}
& \thd\GetGen{\LTOTGENTyrForty}{2}
& \sed\GetGen{\LTOTGENTyrForty}{3}
& \GetGen{\LTOTGENTyrForty}{4}
& \thd\GetGen{\LTOTGENTyrForty}{5}
& \sed\GENMean{\LTOTGENTyrForty} \\

& \textbf{\paperN(Ours)} & 40 
& \thd\GetGen{\LTOTGENOursForty}{1}
& \fst\GetGen{\LTOTGENOursForty}{2}
& \thd\GetGen{\LTOTGENOursForty}{3}
& \fst\GetGen{\LTOTGENOursForty}{4}
& \fst\GetGen{\LTOTGENOursForty}{5}
& \fst\GENMean{\LTOTGENOursForty} \\

\midrule

\multirow{11}{*}{Llama2-70B}

& Dense & -
& \GetGen{\LTSZGENDense}{1}
& \GetGen{\LTSZGENDense}{2}
& \GetGen{\LTSZGENDense}{3}
& \GetGen{\LTSZGENDense}{4}
& \GetGen{\LTSZGENDense}{5}
& \GENMean{\LTSZGENDense} \\

\cline{2-9}

& Wanda-sp & 20
& \GetGen{\LTSZGENWandaTwenty}{1}
& \GetGen{\LTSZGENWandaTwenty}{2}
& \GetGen{\LTSZGENWandaTwenty}{3}
& \GetGen{\LTSZGENWandaTwenty}{4}
& \GetGen{\LTSZGENWandaTwenty}{5}
& \GENMean{\LTSZGENWandaTwenty} \\

& FLAP & 20
& \GetGen{\LTSZGENFLAPTwenty}{1}
& \GetGen{\LTSZGENFLAPTwenty}{2}
& \GetGen{\LTSZGENFLAPTwenty}{3}
& \GetGen{\LTSZGENFLAPTwenty}{4}
& \GetGen{\LTSZGENFLAPTwenty}{5}
& \GENMean{\LTSZGENFLAPTwenty} \\

& Probe Pruning & 20
& \sed\GetGen{\LTSZGENPPTwenty}{1}
& \fst\GetGen{\LTSZGENPPTwenty}{2}
& \thd\GetGen{\LTSZGENPPTwenty}{3}
& \sed\GetGen{\LTSZGENPPTwenty}{4}
& \sed\GetGen{\LTSZGENPPTwenty}{5}
& \sed\GENMean{\LTSZGENPPTwenty} \\

& T{\'y}r & 20
& \thd\GetGen{\LTSZGENTyrTwenty}{1}
& \thd\GetGen{\LTSZGENTyrTwenty}{2}
& \fst\GetGen{\LTSZGENTyrTwenty}{3}
& \thd\GetGen{\LTSZGENTyrTwenty}{4}
& \thd\GetGen{\LTSZGENTyrTwenty}{5}
& \thd\GENMean{\LTSZGENTyrTwenty} \\

& \textbf{\paperN(Ours)} & 20
& \fst\GetGen{\LTSZGENOursTwenty}{1}
& \sed\GetGen{\LTSZGENOursTwenty}{2}
& \sed\GetGen{\LTSZGENOursTwenty}{3}
& \fst\GetGen{\LTSZGENOursTwenty}{4}
& \fst\GetGen{\LTSZGENOursTwenty}{5}
& \fst\GENMean{\LTSZGENOursTwenty} \\

\cline{2-9}

& Wanda-sp & 40
& \GetGen{\LTSZGENWandaForty}{1}
& \GetGen{\LTSZGENWandaForty}{2}
& \GetGen{\LTSZGENWandaForty}{3}
& \GetGen{\LTSZGENWandaForty}{4}
& \GetGen{\LTSZGENWandaForty}{5}
& \GENMean{\LTSZGENWandaForty} \\

& FLAP & 40
& \GetGen{\LTSZGENFLAPForty}{1}
& \GetGen{\LTSZGENFLAPForty}{2}
& \GetGen{\LTSZGENFLAPForty}{3}
& \GetGen{\LTSZGENFLAPForty}{4}
& \GetGen{\LTSZGENFLAPForty}{5}
& \GENMean{\LTSZGENFLAPForty} \\

& Probe Pruning & 40
& \sed\GetGen{\LTSZGENPPForty}{1}
& \fst\GetGen{\LTSZGENPPForty}{2}
& \fst\GetGen{\LTSZGENPPForty}{3}
& \sed\GetGen{\LTSZGENPPForty}{4}
& \sed\GetGen{\LTSZGENPPForty}{5}
& \sed\GENMean{\LTSZGENPPForty} \\

& T{\'y}r & 40
& \thd\GetGen{\LTSZGENTyrForty}{1}
& \GetGen{\LTSZGENTyrForty}{2}
& \sed\GetGen{\LTSZGENTyrForty}{3}
& \thd\GetGen{\LTSZGENTyrForty}{4}
& \thd\GetGen{\LTSZGENTyrForty}{5}
& \thd\GENMean{\LTSZGENTyrForty} \\

& \textbf{\paperN(Ours)} & 40 
& \fst\GetGen{\LTSZGENOursForty}{1}
& \sed\GetGen{\LTSZGENOursForty}{2}
& \thd\GetGen{\LTSZGENOursForty}{3}
& \fst\GetGen{\LTSZGENOursForty}{4}
& \fst\GetGen{\LTSZGENOursForty}{5}
& \fst\GENMean{\LTSZGENOursForty} \\

\bottomrule
\end{tabular}
\end{table}

\def\LTOEGENDense{80.88,48.40,7.70,35.37,50.57}

\def\LTOEGENWandaTwenty{3.12,0.00,0.00,0.00,0.83}
\def\LTOEGENFLAPTwenty{73.01,1.80,2.83,1.22,3.56}
\def\LTOEGENPPTwenty{54.45,17.60,8.20,11.59,6.44}
\def\LTOEGENTyrTwenty{75.52,8.60,4.85,7.93,8.79}
\def\LTOEGENOursTwenty{79.70,39.20,3.82,23.78,40.33}

\def\LTOEGENWandaForty{0.00,0.00,0.06,0.00,0.23}
\def\LTOEGENFLAPForty{32.16,0.00,0.00,0.00,2.05}
\def\LTOEGENPPForty{5.38,0.00,0.30,0.61,1.36}
\def\LTOEGENTyrForty{45.51,0.00,4.15,0.00,1.90}
\def\LTOEGENOursForty{69.20,14.40,2.63,12.20,18.27}
\def\LTOSZGENDense{83.70,64.60,24.96,51.22,81.20}

\def\LTOSZGENWandaTwenty{0.19,0.00,0.06,0.00,0.00}
\def\LTOSZGENFLAPTwenty{7.04,0.00,0.06,0.00,1.29}
\def\LTOSZGENPPTwenty{81.27,54.60,20.19,26.83,74.68}
\def\LTOSZGENTyrTwenty{82.51,47.20,20.64,15.85,67.85}
\def\LTOSZGENOursTwenty{83.11,59.20,17.40,40.85,77.79}

\def\LTOSZGENWandaForty{0.07,0.00,0.00,0.00,0.08}
\def\LTOSZGENFLAPForty{2.58,0.00,0.03,0.00,0.61}
\def\LTOSZGENPPForty{68.80,25.40,9.36,7.93,30.86}
\def\LTOSZGENTyrForty{78.85,15.40,8.25,5.49,43.22}
\def\LTOSZGENOursForty{80.24,40.60,10.25,25.00,59.59}

\begin{table}[t]
\centering
\caption{Accuracies (\%) of Llama-3 \cite{dubey2024llama} for 5 Generative tasks with different levels of structured sparsity.}
\label{tab:llama3-GEN}

\setlength{\tabcolsep}{7.5pt}
\renewcommand{\arraystretch}{1.1}
\footnotesize

\begin{tabular}{l|lc|l*{6}{c}}
\toprule
Params & Method & PR
& coqa & mbpp & nq\_open & humaneval & gsm8k
& Mean \\
\midrule

\multirow{11}{*}{Llama3.1-8B}

& Dense & -
& \GetGen{\LTOEGENDense}{1}
& \GetGen{\LTOEGENDense}{2}
& \GetGen{\LTOEGENDense}{3}
& \GetGen{\LTOEGENDense}{4}
& \GetGen{\LTOEGENDense}{5}
& \GENMean{\LTOEGENDense} \\

\cline{2-9}

& Wanda-sp & 20
& \GetGen{\LTOEGENWandaTwenty}{1}
& \GetGen{\LTOEGENWandaTwenty}{2}
& \GetGen{\LTOEGENWandaTwenty}{3}
& \GetGen{\LTOEGENWandaTwenty}{4}
& \GetGen{\LTOEGENWandaTwenty}{5}
& \GENMean{\LTOEGENWandaTwenty} \\

& FLAP & 20
& \thd\GetGen{\LTOEGENFLAPTwenty}{1}
& \GetGen{\LTOEGENFLAPTwenty}{2}
& \GetGen{\LTOEGENFLAPTwenty}{3}
& \GetGen{\LTOEGENFLAPTwenty}{4}
& \GetGen{\LTOEGENFLAPTwenty}{5}
& \GENMean{\LTOEGENFLAPTwenty} \\

& Probe Pruning & 20
& \GetGen{\LTOEGENPPTwenty}{1}
& \sed\GetGen{\LTOEGENPPTwenty}{2}
& \fst\GetGen{\LTOEGENPPTwenty}{3}
& \sed\GetGen{\LTOEGENPPTwenty}{4}
& \thd\GetGen{\LTOEGENPPTwenty}{5}
& \thd\GENMean{\LTOEGENPPTwenty} \\

& T{\'y}r & 20
& \sed\GetGen{\LTOEGENTyrTwenty}{1}
& \thd\GetGen{\LTOEGENTyrTwenty}{2}
& \sed\GetGen{\LTOEGENTyrTwenty}{3}
& \thd\GetGen{\LTOEGENTyrTwenty}{4}
& \sed\GetGen{\LTOEGENTyrTwenty}{5}
& \sed\GENMean{\LTOEGENTyrTwenty} \\

& \textbf{\paperN(Ours)} & 20 
& \fst\GetGen{\LTOEGENOursTwenty}{1}
& \fst\GetGen{\LTOEGENOursTwenty}{2}
& \thd\GetGen{\LTOEGENOursTwenty}{3}
& \fst\GetGen{\LTOEGENOursTwenty}{4}
& \fst\GetGen{\LTOEGENOursTwenty}{5}
& \fst\GENMean{\LTOEGENOursTwenty} \\

\cline{2-9}

& Wanda-sp & 40
& \GetGen{\LTOEGENWandaForty}{1}
& \GetGen{\LTOEGENWandaForty}{2}
& \GetGen{\LTOEGENWandaForty}{3}
& \GetGen{\LTOEGENWandaForty}{4}
& \GetGen{\LTOEGENWandaForty}{5}
& \GENMean{\LTOEGENWandaForty} \\

& FLAP & 40
& \thd\GetGen{\LTOEGENFLAPForty}{1}
& \GetGen{\LTOEGENFLAPForty}{2}
& \GetGen{\LTOEGENFLAPForty}{3}
& \GetGen{\LTOEGENFLAPForty}{4}
& \sed\GetGen{\LTOEGENFLAPForty}{5}
& \thd\GENMean{\LTOEGENFLAPForty} \\

& Probe Pruning & 40
& \GetGen{\LTOEGENPPForty}{1}
& \GetGen{\LTOEGENPPForty}{2}
& \thd\GetGen{\LTOEGENPPForty}{3}
& \sed\GetGen{\LTOEGENPPForty}{4}
& \GetGen{\LTOEGENPPForty}{5}
& \GENMean{\LTOEGENPPForty} \\

& T{\'y}r & 40
& \sed\GetGen{\LTOEGENTyrForty}{1}
& \GetGen{\LTOEGENTyrForty}{2}
& \fst\GetGen{\LTOEGENTyrForty}{3}
& \GetGen{\LTOEGENTyrForty}{4}
& \thd\GetGen{\LTOEGENTyrForty}{5}
& \sed\GENMean{\LTOEGENTyrForty} \\

& \textbf{\paperN(Ours)} & 40
& \fst\GetGen{\LTOEGENOursForty}{1}
& \fst\GetGen{\LTOEGENOursForty}{2}
& \sed\GetGen{\LTOEGENOursForty}{3}
& \fst\GetGen{\LTOEGENOursForty}{4}
& \fst\GetGen{\LTOEGENOursForty}{5}
& \fst\GENMean{\LTOEGENOursForty} \\

\midrule

\multirow{11}{*}{Llama3.1-70B}

& Dense & -
& \GetGen{\LTOSZGENDense}{1}
& \GetGen{\LTOSZGENDense}{2}
& \GetGen{\LTOSZGENDense}{3}
& \GetGen{\LTOSZGENDense}{4}
& \GetGen{\LTOSZGENDense}{5}
& \GENMean{\LTOSZGENDense} \\

\cline{2-9}

& Wanda-sp & 20
& \GetGen{\LTOSZGENWandaTwenty}{1}
& \GetGen{\LTOSZGENWandaTwenty}{2}
& \GetGen{\LTOSZGENWandaTwenty}{3}
& \GetGen{\LTOSZGENWandaTwenty}{4}
& \GetGen{\LTOSZGENWandaTwenty}{5}
& \GENMean{\LTOSZGENWandaTwenty} \\

& FLAP & 20
& \GetGen{\LTOSZGENFLAPTwenty}{1}
& \GetGen{\LTOSZGENFLAPTwenty}{2}
& \GetGen{\LTOSZGENFLAPTwenty}{3}
& \GetGen{\LTOSZGENFLAPTwenty}{4}
& \GetGen{\LTOSZGENFLAPTwenty}{5}
& \GENMean{\LTOSZGENFLAPTwenty} \\

& Probe Pruning & 20
& \thd\GetGen{\LTOSZGENPPTwenty}{1}
& \sed\GetGen{\LTOSZGENPPTwenty}{2}
& \sed\GetGen{\LTOSZGENPPTwenty}{3}
& \sed\GetGen{\LTOSZGENPPTwenty}{4}
& \sed\GetGen{\LTOSZGENPPTwenty}{5}
& \sed\GENMean{\LTOSZGENPPTwenty} \\

& T{\'y}r & 20
& \sed\GetGen{\LTOSZGENTyrTwenty}{1}
& \thd\GetGen{\LTOSZGENTyrTwenty}{2}
& \fst\GetGen{\LTOSZGENTyrTwenty}{3}
& \thd\GetGen{\LTOSZGENTyrTwenty}{4}
& \thd\GetGen{\LTOSZGENTyrTwenty}{5}
& \thd\GENMean{\LTOSZGENTyrTwenty} \\

& \textbf{\paperN(Ours)} & 20 
& \fst\GetGen{\LTOSZGENOursTwenty}{1}
& \fst\GetGen{\LTOSZGENOursTwenty}{2}
& \thd\GetGen{\LTOSZGENOursTwenty}{3}
& \fst\GetGen{\LTOSZGENOursTwenty}{4}
& \fst\GetGen{\LTOSZGENOursTwenty}{5}
& \fst\GENMean{\LTOSZGENOursTwenty} \\

\cline{2-9}

& Wanda-sp & 40
& \GetGen{\LTOSZGENWandaForty}{1}
& \GetGen{\LTOSZGENWandaForty}{2}
& \GetGen{\LTOSZGENWandaForty}{3}
& \GetGen{\LTOSZGENWandaForty}{4}
& \GetGen{\LTOSZGENWandaForty}{5}
& \GENMean{\LTOSZGENWandaForty} \\

& FLAP & 40
& \GetGen{\LTOSZGENFLAPForty}{1}
& \GetGen{\LTOSZGENFLAPForty}{2}
& \GetGen{\LTOSZGENFLAPForty}{3}
& \GetGen{\LTOSZGENFLAPForty}{4}
& \GetGen{\LTOSZGENFLAPForty}{5}
& \GENMean{\LTOSZGENFLAPForty} \\

& Probe Pruning & 40
& \thd\GetGen{\LTOSZGENPPForty}{1}
& \sed\GetGen{\LTOSZGENPPForty}{2}
& \sed\GetGen{\LTOSZGENPPForty}{3}
& \sed\GetGen{\LTOSZGENPPForty}{4}
& \thd\GetGen{\LTOSZGENPPForty}{5}
& \thd\GENMean{\LTOSZGENPPForty} \\

& T{\'y}r & 40
& \sed\GetGen{\LTOSZGENTyrForty}{1}
& \thd\GetGen{\LTOSZGENTyrForty}{2}
& \thd\GetGen{\LTOSZGENTyrForty}{3}
& \thd\GetGen{\LTOSZGENTyrForty}{4}
& \sed\GetGen{\LTOSZGENTyrForty}{5}
& \sed\GENMean{\LTOSZGENTyrForty} \\

& \textbf{\paperN(Ours)} & 40 
& \fst\GetGen{\LTOSZGENOursForty}{1}
& \fst\GetGen{\LTOSZGENOursForty}{2}
& \fst\GetGen{\LTOSZGENOursForty}{3}
& \fst\GetGen{\LTOSZGENOursForty}{4}
& \fst\GetGen{\LTOSZGENOursForty}{5}
& \fst\GENMean{\LTOSZGENOursForty} \\

\bottomrule
\end{tabular}
\end{table}
\def\QTEGENDense{81.97,65.60,7.31,60.98,87.19}

\def\QTEGENWandaTwenty{43.16,0.00,0.11,0.00,1.59}
\def\QTEGENFLAPTwenty{72.11,0.00,0.55,0.00,1.90}
\def\QTEGENTyrTwenty{75.17,3.20,3.32,0.61,31.16}
\def\QTEGENOursTwenty{81.05,56.00,2.35,56.10,85.14}

\def\QTEGENWandaForty{1.71,0.00,0.00,0.00,1.36}
\def\QTEGENFLAPForty{1.64,0.00,0.14,0.00,0.23}
\def\QTEGENTyrForty{48.71,0.00,5.32,0.00,2.05}
\def\QTEGENOursForty{73.73,31.80,0.89,35.37,71.87}

\def\QTTTGENDense{82.45,78.20,2.52,37.81,74.45}

\def\QTTTGENWandaTwenty{63.39,0.20,1.11,0.61,9.25}
\def\QTTTGENFLAPTwenty{81.01,3.40,3.63,1.83,0.00}
\def\QTTTGENTyrTwenty{81.31,44.00,1.14,0.00,51.93}
\def\QTTTGENOursTwenty{80.75,70.00,3.16,39.63,71.65}

\def\QTTTGENWandaForty{24.21,0.00,0.08,0.00,1.06}
\def\QTTTGENFLAPForty{29.31,0.00,0.03,0.00,1.06}
\def\QTTTGENTyrForty{79.55,0.00,2.69,1.22,64.52}
\def\QTTTGENOursForty{75.98,49.20,1.63,37.81,74.30}

\begin{table}[t]
\centering
\caption{Accuracies (\%) of Qwen3~\cite{yang2025qwen3} for 5 Generative tasks with different levels of structured sparsity.}
\label{tab:qwen3-Gen}

\setlength{\tabcolsep}{8.1pt}
\renewcommand{\arraystretch}{1.1}
\footnotesize

\begin{tabular}{l|lc|l*{6}{c}}
\toprule
Params & Method & PR
& coqa & mbpp & nq\_open & humaneval & gsm8k
& Mean \\
\midrule

\multirow{9}{*}{Qwen3-8B}

& Dense & -
& \GetGen{\QTEGENDense}{1}
& \GetGen{\QTEGENDense}{2}
& \GetGen{\QTEGENDense}{3}
& \GetGen{\QTEGENDense}{4}
& \GetGen{\QTEGENDense}{5}
& \GENMean{\QTEGENDense} \\

\cline{2-9}

& Wanda-sp & 20
& \GetGen{\QTEGENWandaTwenty}{1}
& \GetGen{\QTEGENWandaTwenty}{2}
& \GetGen{\QTEGENWandaTwenty}{3}
& \GetGen{\QTEGENWandaTwenty}{4}
& \GetGen{\QTEGENWandaTwenty}{5}
& \GENMean{\QTEGENWandaTwenty} \\

& FLAP & 20
& \thd\GetGen{\QTEGENFLAPTwenty}{1}
& \GetGen{\QTEGENFLAPTwenty}{2}
& \thd\GetGen{\QTEGENFLAPTwenty}{3}
& \GetGen{\QTEGENFLAPTwenty}{4}
& \thd\GetGen{\QTEGENFLAPTwenty}{5}
& \thd\GENMean{\QTEGENFLAPTwenty} \\


& T{\'y}r & 20
& \sed\GetGen{\QTEGENTyrTwenty}{1}
& \sed\GetGen{\QTEGENTyrTwenty}{2}
& \fst\GetGen{\QTEGENTyrTwenty}{3}
& \sed\GetGen{\QTEGENTyrTwenty}{4}
& \sed\GetGen{\QTEGENTyrTwenty}{5}
& \sed\GENMean{\QTEGENTyrTwenty} \\

& \textbf{\paperN(Ours)} & 20 
& \fst\GetGen{\QTEGENOursTwenty}{1}
& \fst\GetGen{\QTEGENOursTwenty}{2}
& \sed\GetGen{\QTEGENOursTwenty}{3}
& \fst\GetGen{\QTEGENOursTwenty}{4}
& \fst\GetGen{\QTEGENOursTwenty}{5}
& \fst\GENMean{\QTEGENOursTwenty} \\

\cline{2-9}

& Wanda-sp & 40
& \thd\GetGen{\QTEGENWandaForty}{1}
& \GetGen{\QTEGENWandaForty}{2}
& \GetGen{\QTEGENWandaForty}{3}
& \GetGen{\QTEGENWandaForty}{4}
& \thd\GetGen{\QTEGENWandaForty}{5}
& \thd\GENMean{\QTEGENWandaForty} \\

& FLAP & 40
& \GetGen{\QTEGENFLAPForty}{1}
& \GetGen{\QTEGENFLAPForty}{2}
& \thd\GetGen{\QTEGENFLAPForty}{3}
& \GetGen{\QTEGENFLAPForty}{4}
& \GetGen{\QTEGENFLAPForty}{5}
& \GENMean{\QTEGENFLAPForty} \\


& T{\'y}r & 40
& \sed\GetGen{\QTEGENTyrForty}{1}
& \GetGen{\QTEGENTyrForty}{2}
& \fst\GetGen{\QTEGENTyrForty}{3}
& \GetGen{\QTEGENTyrForty}{4}
& \sed\GetGen{\QTEGENTyrForty}{5}
& \sed\GENMean{\QTEGENTyrForty} \\

& \textbf{\paperN(Ours)} & 40
& \fst\GetGen{\QTEGENOursForty}{1}
& \fst\GetGen{\QTEGENOursForty}{2}
& \sed\GetGen{\QTEGENOursForty}{3}
& \fst\GetGen{\QTEGENOursForty}{4}
& \fst\GetGen{\QTEGENOursForty}{5}
& \fst\GENMean{\QTEGENOursForty} \\

\midrule

\multirow{9}{*}{Qwen3-32B}

& Dense & -
& \GetGen{\QTTTGENDense}{1}
& \GetGen{\QTTTGENDense}{2}
& \GetGen{\QTTTGENDense}{3}
& \GetGen{\QTTTGENDense}{4}
& \GetGen{\QTTTGENDense}{5}
& \GENMean{\QTTTGENDense} \\

\cline{2-9}

& Wanda-sp & 20
& \GetGen{\QTTTGENWandaTwenty}{1}
& \GetGen{\QTTTGENWandaTwenty}{2}
& \GetGen{\QTTTGENWandaTwenty}{3}
& \thd\GetGen{\QTTTGENWandaTwenty}{4}
& \thd\GetGen{\QTTTGENWandaTwenty}{5}
& \GENMean{\QTTTGENWandaTwenty} \\

& FLAP & 20
& \sed\GetGen{\QTTTGENFLAPTwenty}{1}
& \thd\GetGen{\QTTTGENFLAPTwenty}{2}
& \fst\GetGen{\QTTTGENFLAPTwenty}{3}
& \sed\GetGen{\QTTTGENFLAPTwenty}{4}
& \GetGen{\QTTTGENFLAPTwenty}{5}
& \thd\GENMean{\QTTTGENFLAPTwenty} \\


& T{\'y}r & 20
& \fst\GetGen{\QTTTGENTyrTwenty}{1}
& \sed\GetGen{\QTTTGENTyrTwenty}{2}
& \thd\GetGen{\QTTTGENTyrTwenty}{3}
& \GetGen{\QTTTGENTyrTwenty}{4}
& \sed\GetGen{\QTTTGENTyrTwenty}{5}
& \sed\GENMean{\QTTTGENTyrTwenty} \\

& \textbf{\paperN(Ours)} & 20
& \thd\GetGen{\QTTTGENOursTwenty}{1}
& \fst\GetGen{\QTTTGENOursTwenty}{2}
& \sed\GetGen{\QTTTGENOursTwenty}{3}
& \fst\GetGen{\QTTTGENOursTwenty}{4}
& \fst\GetGen{\QTTTGENOursTwenty}{5}
& \fst\GENMean{\QTTTGENOursTwenty} \\

\cline{2-9}

& Wanda-sp & 40
& \GetGen{\QTTTGENWandaForty}{1}
& \GetGen{\QTTTGENWandaForty}{2}
& \thd\GetGen{\QTTTGENWandaForty}{3}
& \GetGen{\QTTTGENWandaForty}{4}
& \thd\GetGen{\QTTTGENWandaForty}{5}
& \GENMean{\QTTTGENWandaForty} \\

& FLAP & 40
& \thd\GetGen{\QTTTGENFLAPForty}{1}
& \GetGen{\QTTTGENFLAPForty}{2}
& \GetGen{\QTTTGENFLAPForty}{3}
& \GetGen{\QTTTGENFLAPForty}{4}
& \thd\GetGen{\QTTTGENFLAPForty}{5}
& \thd\GENMean{\QTTTGENFLAPForty} \\


& T{\'y}r & 40
& \fst\GetGen{\QTTTGENTyrForty}{1}
& \GetGen{\QTTTGENTyrForty}{2}
& \fst\GetGen{\QTTTGENTyrForty}{3}
& \sed\GetGen{\QTTTGENTyrForty}{4}
& \sed\GetGen{\QTTTGENTyrForty}{5}
& \sed\GENMean{\QTTTGENTyrForty} \\

& \textbf{\paperN(Ours)} & 40 
& \sed\GetGen{\QTTTGENOursForty}{1}
& \fst\GetGen{\QTTTGENOursForty}{2}
& \sed\GetGen{\QTTTGENOursForty}{3}
& \fst\GetGen{\QTTTGENOursForty}{4}
& \fst\GetGen{\QTTTGENOursForty}{5}
& \fst\GENMean{\QTTTGENOursForty} \\

\bottomrule
\end{tabular}
\end{table}

\def\QOFDense@QA{79.73,68.59,57.98,69.53,73.06,41.81,30.80}
\def\QOFWandaTwenty@QA{68.32,59.93,52.99,58.64,72.35,39.93,25.60}
\def\QOFFLAPTwenty@QA{72.66,64.62,47.43,67.32,63.68,32.17,27.20}
\def\QOFTyrTwenty@QA{76.64,69.31,53.56,69.77,74.96,41.81,31.20}
\def\QOFOursTwenty@QA{77.65,68.95,55.67,68.04,70.96,39.16,28.40}

\def\QOFWandaForty@QA{62.17,52.71,29.56,53.20,36.79,19.28,12.60}
\def\QOFFLAPForty@QA{64.04,62.09,42.89,64.17,59.85,30.46,25.20}
\def\QOFTyrForty@QA{64.45,53.43,36.34,56.28,70.71,35.58,24.40}
\def\QOFOursForty@QA{74.89,62.09,47.21,66.69,66.50,35.24,25.00}

\def\QWTFDense@QA{86.36,75.09,62.93,74.03,75.08,46.67,32.80}
\def\QWTFWandaTwenty@QA{59.17,53.07,32.63,53.51,46.97,23.55,15.40}
\def\QWTFFLAPTwenty@QA{87.25,75.45,62.78,75.14,76.60,46.08,32.00}
\def\QWTFOursTwenty@QA{87.00,75.45,62.37,73.95,74.75,47.18,31.80}
\def\QWTFWandaForty@QA{48.20,52.71,27.15,49.65,27.19,21.93,14.20}
\def\QWTFFLAPForty@QA{73.73,67.51,55.80,68.27,70.92,41.38,29.20}
\def\QWTFOursForty@QA{85.60,75.09,60.28,74.67,75.30,47.53,31.20}

\def\QTTZDense@QA{88.75,82.31,59.56,70.56,79.29,52.82,34.00}
\def\QTTZWandaTwenty@QA{73.30,63.54,44.73,60.62,62.21,37.37,24.80}
\def\QTTZFLAPTwenty@QA{37.83,47.29,25.59,50.12,24.33,21.42,19.80}
\def\QTTZTyrTwenty@QA{86.33,77.26,53.10,68.90,77.36,48.04,35.60}
\def\QTTZOursTwenty@QA{87.28,81.59,55.00,68.43,75.25,48.04,34.40}

\def\QTTZWandaForty@QA{56.70,50.18,34.49,49.49,41.71,23.72,16.80}
\def\QTTZFLAPForty@QA{37.83,47.29,25.59,50.12,24.33,21.42,19.80}
\def\QTTZTyrForty@QA{65.26,60.29,43.03,61.48,73.78,41.72,30.60}
\def\QTTZOursForty@QA{79.63,74.73,41.72,63.54,63.13,36.43,25.40}



\begin{table}[t]
\centering
\caption{Detailed Accuracies (\%) of Qwen1.5-MoE \cite{qwen_moe}, Qwen2-MoE \cite{yang2024qwen2technicalreport} and Qwen3-MoE~\cite{yang2025qwen3} models for 7 QA tasks with different levels of structured sparsity.}
\label{tab:moe-QA}

\setlength{\tabcolsep}{4pt}
\footnotesize

\begin{tabular}{l|lc|*{8}{c}}
\toprule
Params & Method & PR(\%) & BoolQ & RTE & HS & WG & ARC-e & ARC-c & OBQA & Mean \\
\midrule

\multirow{11}{*}{\shortstack[l]{Qwen1.5-MoE\\A2.7B}}

& Dense & -
& \GetQA{\QOFDense@QA}{1}
& \GetQA{\QOFDense@QA}{2}
& \GetQA{\QOFDense@QA}{3}
& \GetQA{\QOFDense@QA}{4}
& \GetQA{\QOFDense@QA}{5}
& \GetQA{\QOFDense@QA}{6}
& \GetQA{\QOFDense@QA}{7}
& \QAMean{\QOFDense@QA} \\

\cline{2-11}

& Wanda-sp & 20
& \GetQA{\QOFWandaTwenty@QA}{1}
& \GetQA{\QOFWandaTwenty@QA}{2}
& \thd\GetQA{\QOFWandaTwenty@QA}{3}
& \GetQA{\QOFWandaTwenty@QA}{4}
& \sed\GetQA{\QOFWandaTwenty@QA}{5}
& \sed\GetQA{\QOFWandaTwenty@QA}{6}
& \GetQA{\QOFWandaTwenty@QA}{7}
& \thd\QAMean{\QOFWandaTwenty@QA} \\

& FLAP & 20
& \thd\GetQA{\QOFFLAPTwenty@QA}{1}
& \thd\GetQA{\QOFFLAPTwenty@QA}{2}
& \GetQA{\QOFFLAPTwenty@QA}{3}
& \thd\GetQA{\QOFFLAPTwenty@QA}{4}
& \GetQA{\QOFFLAPTwenty@QA}{5}
& \GetQA{\QOFFLAPTwenty@QA}{6}
& \thd\GetQA{\QOFFLAPTwenty@QA}{7}
& \QAMean{\QOFFLAPTwenty@QA} \\

& T{\'y}r & 20
& \sed\GetQA{\QOFTyrTwenty@QA}{1}
& \fst\GetQA{\QOFTyrTwenty@QA}{2}
& \sed\GetQA{\QOFTyrTwenty@QA}{3}
& \fst\GetQA{\QOFTyrTwenty@QA}{4}
& \fst\GetQA{\QOFTyrTwenty@QA}{5}
& \fst\GetQA{\QOFTyrTwenty@QA}{6}
& \fst\GetQA{\QOFTyrTwenty@QA}{7}
& \fst\QAMean{\QOFTyrTwenty@QA} \\

& \textbf{\paperN(Ours)} & 20
& \fst\GetQA{\QOFOursTwenty@QA}{1}
& \sed\GetQA{\QOFOursTwenty@QA}{2}
& \fst\GetQA{\QOFOursTwenty@QA}{3}
& \sed\GetQA{\QOFOursTwenty@QA}{4}
& \thd\GetQA{\QOFOursTwenty@QA}{5}
& \thd\GetQA{\QOFOursTwenty@QA}{6}
& \sed\GetQA{\QOFOursTwenty@QA}{7}
& \sed\QAMean{\QOFOursTwenty@QA} \\

\cline{2-11}

& Wanda-sp & 40
& \GetQA{\QOFWandaForty@QA}{1}
& \GetQA{\QOFWandaForty@QA}{2}
& \GetQA{\QOFWandaForty@QA}{3}
& \GetQA{\QOFWandaForty@QA}{4}
& \GetQA{\QOFWandaForty@QA}{5}
& \GetQA{\QOFWandaForty@QA}{6}
& \GetQA{\QOFWandaForty@QA}{7}
& \QAMean{\QOFWandaForty@QA} \\

& FLAP & 40
& \thd\GetQA{\QOFFLAPForty@QA}{1}
& \fst\GetQA{\QOFFLAPForty@QA}{2}
& \sed\GetQA{\QOFFLAPForty@QA}{3}
& \sed\GetQA{\QOFFLAPForty@QA}{4}
& \thd\GetQA{\QOFFLAPForty@QA}{5}
& \thd\GetQA{\QOFFLAPForty@QA}{6}
& \fst\GetQA{\QOFFLAPForty@QA}{7}
& \sed\QAMean{\QOFFLAPForty@QA} \\

& T{\'y}r & 40
& \sed\GetQA{\QOFTyrForty@QA}{1}
& \sed\GetQA{\QOFTyrForty@QA}{2}
& \thd\GetQA{\QOFTyrForty@QA}{3}
& \thd\GetQA{\QOFTyrForty@QA}{4}
& \fst\GetQA{\QOFTyrForty@QA}{5}
& \fst\GetQA{\QOFTyrForty@QA}{6}
& \thd\GetQA{\QOFTyrForty@QA}{7}
& \thd\QAMean{\QOFTyrForty@QA} \\

& \textbf{\paperN(Ours)} & 40 
& \fst\GetQA{\QOFOursForty@QA}{1}
& \fst\GetQA{\QOFOursForty@QA}{2}
& \fst\GetQA{\QOFOursForty@QA}{3}
& \fst\GetQA{\QOFOursForty@QA}{4}
& \sed\GetQA{\QOFOursForty@QA}{5}
& \sed\GetQA{\QOFOursForty@QA}{6}
& \sed\GetQA{\QOFOursForty@QA}{7}
& \fst\QAMean{\QOFOursForty@QA} \\

\midrule

\multirow{7}{*}{\shortstack[l]{Qwen2\\57B-A14B}}

& Dense & -
& \GetQA{\QWTFDense@QA}{1}
& \GetQA{\QWTFDense@QA}{2}
& \GetQA{\QWTFDense@QA}{3}
& \GetQA{\QWTFDense@QA}{4}
& \GetQA{\QWTFDense@QA}{5}
& \GetQA{\QWTFDense@QA}{6}
& \GetQA{\QWTFDense@QA}{7}
& \QAMean{\QWTFDense@QA} \\

\cline{2-11}

& Wanda-sp & 20
& \thd\GetQA{\QWTFWandaTwenty@QA}{1}
& \thd\GetQA{\QWTFWandaTwenty@QA}{2}
& \thd\GetQA{\QWTFWandaTwenty@QA}{3}
& \thd\GetQA{\QWTFWandaTwenty@QA}{4}
& \thd\GetQA{\QWTFWandaTwenty@QA}{5}
& \thd\GetQA{\QWTFWandaTwenty@QA}{6}
& \thd\GetQA{\QWTFWandaTwenty@QA}{7}
& \thd\QAMean{\QWTFWandaTwenty@QA} \\

& FLAP & 20
& \fst\GetQA{\QWTFFLAPTwenty@QA}{1}
& \fst\GetQA{\QWTFFLAPTwenty@QA}{2}
& \fst\GetQA{\QWTFFLAPTwenty@QA}{3}
& \fst\GetQA{\QWTFFLAPTwenty@QA}{4}
& \fst\GetQA{\QWTFFLAPTwenty@QA}{5}
& \sed\GetQA{\QWTFFLAPTwenty@QA}{6}
& \fst\GetQA{\QWTFFLAPTwenty@QA}{7}
& \fst\QAMean{\QWTFFLAPTwenty@QA} \\

& \textbf{\paperN(Ours)} & 20 
& \sed\GetQA{\QWTFOursTwenty@QA}{1}
& \sed\GetQA{\QWTFOursTwenty@QA}{2}
& \sed\GetQA{\QWTFOursTwenty@QA}{3}
& \sed\GetQA{\QWTFOursTwenty@QA}{4}
& \sed\GetQA{\QWTFOursTwenty@QA}{5}
& \fst\GetQA{\QWTFOursTwenty@QA}{6}
& \sed\GetQA{\QWTFOursTwenty@QA}{7}
& \sed\QAMean{\QWTFOursTwenty@QA} \\

\cline{2-11}

& Wanda-sp & 40
& \thd\GetQA{\QWTFWandaForty@QA}{1}
& \thd\GetQA{\QWTFWandaForty@QA}{2}
& \thd\GetQA{\QWTFWandaForty@QA}{3}
& \thd\GetQA{\QWTFWandaForty@QA}{4}
& \thd\GetQA{\QWTFWandaForty@QA}{5}
& \thd\GetQA{\QWTFWandaForty@QA}{6}
& \thd\GetQA{\QWTFWandaForty@QA}{7}
& \thd\QAMean{\QWTFWandaForty@QA} \\

& FLAP & 40
& \sed\GetQA{\QWTFFLAPForty@QA}{1}
& \sed\GetQA{\QWTFFLAPForty@QA}{2}
& \sed\GetQA{\QWTFFLAPForty@QA}{3}
& \sed\GetQA{\QWTFFLAPForty@QA}{4}
& \sed\GetQA{\QWTFFLAPForty@QA}{5}
& \sed\GetQA{\QWTFFLAPForty@QA}{6}
& \sed\GetQA{\QWTFFLAPForty@QA}{7}
& \sed\QAMean{\QWTFFLAPForty@QA} \\

& \textbf{\paperN(Ours)} & 40 
& \fst\GetQA{\QWTFOursForty@QA}{1}
& \fst\GetQA{\QWTFOursForty@QA}{2}
& \fst\GetQA{\QWTFOursForty@QA}{3}
& \fst\GetQA{\QWTFOursForty@QA}{4}
& \fst\GetQA{\QWTFOursForty@QA}{5}
& \fst\GetQA{\QWTFOursForty@QA}{6}
& \fst\GetQA{\QWTFOursForty@QA}{7}
& \fst\QAMean{\QWTFOursForty@QA} \\

\midrule

\multirow{11}{*}{\shortstack[l]{Qwen3\\30B-A3B}}

& Dense & -
& \GetQA{\QTTZDense@QA}{1}
& \GetQA{\QTTZDense@QA}{2}
& \GetQA{\QTTZDense@QA}{3}
& \GetQA{\QTTZDense@QA}{4}
& \GetQA{\QTTZDense@QA}{5}
& \GetQA{\QTTZDense@QA}{6}
& \GetQA{\QTTZDense@QA}{7}
& \QAMean{\QTTZDense@QA} \\

\cline{2-11}

& Wanda-sp & 20
& \thd\GetQA{\QTTZWandaTwenty@QA}{1}
& \thd\GetQA{\QTTZWandaTwenty@QA}{2}
& \thd\GetQA{\QTTZWandaTwenty@QA}{3}
& \thd\GetQA{\QTTZWandaTwenty@QA}{4}
& \thd\GetQA{\QTTZWandaTwenty@QA}{5}
& \sed\GetQA{\QTTZWandaTwenty@QA}{6}
& \thd\GetQA{\QTTZWandaTwenty@QA}{7}
& \thd\QAMean{\QTTZWandaTwenty@QA} \\

& FLAP & 20
& \GetQA{\QTTZFLAPTwenty@QA}{1}
& \GetQA{\QTTZFLAPTwenty@QA}{2}
& \GetQA{\QTTZFLAPTwenty@QA}{3}
& \GetQA{\QTTZFLAPTwenty@QA}{4}
& \GetQA{\QTTZFLAPTwenty@QA}{5}
& \GetQA{\QTTZFLAPTwenty@QA}{6}
& \GetQA{\QTTZFLAPTwenty@QA}{7}
& \QAMean{\QTTZFLAPTwenty@QA} \\

& T{\'y}r & 20
& \sed\GetQA{\QTTZTyrTwenty@QA}{1}
& \sed\GetQA{\QTTZTyrTwenty@QA}{2}
& \sed\GetQA{\QTTZTyrTwenty@QA}{3}
& \fst\GetQA{\QTTZTyrTwenty@QA}{4}
& \fst\GetQA{\QTTZTyrTwenty@QA}{5}
& \fst\GetQA{\QTTZTyrTwenty@QA}{6}
& \fst\GetQA{\QTTZTyrTwenty@QA}{7}
& \sed\QAMean{\QTTZTyrTwenty@QA} \\

& \textbf{\paperN(Ours)} & 20 
& \fst\GetQA{\QTTZOursTwenty@QA}{1}
& \fst\GetQA{\QTTZOursTwenty@QA}{2}
& \fst\GetQA{\QTTZOursTwenty@QA}{3}
& \sed\GetQA{\QTTZOursTwenty@QA}{4}
& \sed\GetQA{\QTTZOursTwenty@QA}{5}
& \fst\GetQA{\QTTZOursTwenty@QA}{6}
& \sed\GetQA{\QTTZOursTwenty@QA}{7}
& \fst\QAMean{\QTTZOursTwenty@QA} \\

\cline{2-11}

& Wanda-sp & 40
& \thd\GetQA{\QTTZWandaForty@QA}{1}
& \thd\GetQA{\QTTZWandaForty@QA}{2}
& \thd\GetQA{\QTTZWandaForty@QA}{3}
& \GetQA{\QTTZWandaForty@QA}{4}
& \thd\GetQA{\QTTZWandaForty@QA}{5}
& \thd\GetQA{\QTTZWandaForty@QA}{6}
& \GetQA{\QTTZWandaForty@QA}{7}
& \thd\QAMean{\QTTZWandaForty@QA} \\

& FLAP & 40
& \GetQA{\QTTZFLAPForty@QA}{1}
& \GetQA{\QTTZFLAPForty@QA}{2}
& \GetQA{\QTTZFLAPForty@QA}{3}
& \thd\GetQA{\QTTZFLAPForty@QA}{4}
& \GetQA{\QTTZFLAPForty@QA}{5}
& \GetQA{\QTTZFLAPForty@QA}{6}
& \thd\GetQA{\QTTZFLAPForty@QA}{7}
& \QAMean{\QTTZFLAPForty@QA} \\

& T{\'y}r & 40
& \sed\GetQA{\QTTZTyrForty@QA}{1}
& \sed\GetQA{\QTTZTyrForty@QA}{2}
& \fst\GetQA{\QTTZTyrForty@QA}{3}
& \sed\GetQA{\QTTZTyrForty@QA}{4}
& \fst\GetQA{\QTTZTyrForty@QA}{5}
& \fst\GetQA{\QTTZTyrForty@QA}{6}
& \fst\GetQA{\QTTZTyrForty@QA}{7}
& \sed\QAMean{\QTTZTyrForty@QA} \\

& \textbf{\paperN(Ours)} & 40 
& \fst\GetQA{\QTTZOursForty@QA}{1}
& \fst\GetQA{\QTTZOursForty@QA}{2}
& \sed\GetQA{\QTTZOursForty@QA}{3}
& \fst\GetQA{\QTTZOursForty@QA}{4}
& \sed\GetQA{\QTTZOursForty@QA}{5}
& \sed\GetQA{\QTTZOursForty@QA}{6}
& \sed\GetQA{\QTTZOursForty@QA}{7}
& \fst\QAMean{\QTTZOursForty@QA} \\

\bottomrule
\end{tabular}
\end{table}

\def\QOFDense@Gen{78.61,38.80,4.57,32.93,60.96}
\def\QOFWandaTwenty@Gen{57.46,0.00,5.87,0.61,2.12}
\def\QOFFLAPTwenty@Gen{75.76,5.20,5.04,5.49,9.10}
\def\QOFTyrTwenty@Gen{73.77,14.00,3.24,7.32,39.96}
\def\QOFOursTwenty@Gen{79.49,32.60,5.26,32.93,60.20}

\def\QOFWandaForty@Gen{12.53,0.00,0.03,0.00,0.99}
\def\QOFFLAPForty@Gen{67.51,1.40,3.66,0.61,2.58}
\def\QOFTyrForty@Gen{33.56,0.00,3.57,0.00,1.90}
\def\QOFOursForty@Gen{76.99,21.20,2.24,12.20,42.91}

\def\QWTFDense@Gen{82.13,60.00,0.64,51.83,79.99}
\def\QWTFWandaTwenty@Gen{9.87,0.00,0.30,0.00,1.59}
\def\QWTFFLAPTwenty@Gen{81.76,22.60,1.25,11.59,70.43}
\def\QWTFOursTwenty@Gen{81.81,59.60,0.72,48.17,79.45}
\def\QWTFWandaForty@Gen{0.18,0.00,0.03,0.00,1.74}
\def\QWTFFLAPForty@Gen{66.15,1.00,1.77,0.61,16.76}
\def\QWTFOursForty@Gen{81.77,56.80,1.61,47.56,81.58}

\def\QTTZDense@Gen{81.70,73.00,12.08,32.93,89.16}
\def\QTTZWandaTwenty@Gen{53.71,0.40,1.14,3.66,19.64}
\def\QTTZFLAPTwenty@Gen{0.00,0.00,0.00,0.00,0.00}
\def\QTTZTyrTwenty@Gen{79.75,45.60,4.96,20.73,64.29}
\def\QTTZOursTwenty@Gen{80.84,66.40,8.37,45.73,88.48}

\def\QTTZWandaForty@Gen{6.82,0.00,0.06,0.61,0.23}
\def\QTTZFLAPForty@Gen{0.00,0.00,0.00,0.00,0.00}
\def\QTTZTyrForty@Gen{58.65,7.00,2.16,6.10,5.46}
\def\QTTZOursForty@Gen{79.85,28.60,1.77,15.24,67.78}



\begin{table}[t]
\centering
\caption{Detailed Accuracies (\%) of Qwen1.5-MoE \cite{qwen_moe}, Qwen2-MoE \cite{yang2024qwen2technicalreport} and Qwen3-MoE~\cite{yang2025qwen3} models for 5 Generative tasks with different levels of structured sparsity.}
\label{tab:moe-Gen}

\setlength{\tabcolsep}{4.2pt}
\renewcommand{\arraystretch}{1.1}
\footnotesize

\begin{tabular}{l|lc|l*{6}{c}}
\toprule
Params & Method & PR
& coqa & mbpp & nq\_open & humaneval & gsm8k
& Mean \\
\midrule

\multirow{11}{*}{\shortstack[l]{Qwen1.5-MoE\\A2.7B}}

& Dense & -
& \GetGen{\QOFDense@Gen}{1}
& \GetGen{\QOFDense@Gen}{2}
& \GetGen{\QOFDense@Gen}{3}
& \GetGen{\QOFDense@Gen}{4}
& \GetGen{\QOFDense@Gen}{5}
& \GENMean{\QOFDense@Gen} \\

\cline{2-9}

& Wanda-sp & 20
& \GetGen{\QOFWandaTwenty@Gen}{1}
& \GetGen{\QOFWandaTwenty@Gen}{2}
& \fst\GetGen{\QOFWandaTwenty@Gen}{3}
& \GetGen{\QOFWandaTwenty@Gen}{4}
& \GetGen{\QOFWandaTwenty@Gen}{5}
& \GENMean{\QOFWandaTwenty@Gen} \\

& FLAP & 20
& \sed\GetGen{\QOFFLAPTwenty@Gen}{1}
& \thd\GetGen{\QOFFLAPTwenty@Gen}{2}
& \thd\GetGen{\QOFFLAPTwenty@Gen}{3}
& \thd\GetGen{\QOFFLAPTwenty@Gen}{4}
& \thd\GetGen{\QOFFLAPTwenty@Gen}{5}
& \thd\GENMean{\QOFFLAPTwenty@Gen} \\

& T{\'y}r & 20
& \thd\GetGen{\QOFTyrTwenty@Gen}{1}
& \sed\GetGen{\QOFTyrTwenty@Gen}{2}
& \GetGen{\QOFTyrTwenty@Gen}{3}
& \sed\GetGen{\QOFTyrTwenty@Gen}{4}
& \sed\GetGen{\QOFTyrTwenty@Gen}{5}
& \sed\GENMean{\QOFTyrTwenty@Gen} \\

& \textbf{\paperN(Ours)} & 20
& \fst\GetGen{\QOFOursTwenty@Gen}{1}
& \fst\GetGen{\QOFOursTwenty@Gen}{2}
& \sed\GetGen{\QOFOursTwenty@Gen}{3}
& \fst\GetGen{\QOFOursTwenty@Gen}{4}
& \fst\GetGen{\QOFOursTwenty@Gen}{5}
& \fst\GENMean{\QOFOursTwenty@Gen} \\

\cline{2-9}

& Wanda-sp & 40
& \GetGen{\QOFWandaForty@Gen}{1}
& \GetGen{\QOFWandaForty@Gen}{2}
& \GetGen{\QOFWandaForty@Gen}{3}
& \GetGen{\QOFWandaForty@Gen}{4}
& \GetGen{\QOFWandaForty@Gen}{5}
& \GENMean{\QOFWandaForty@Gen} \\

& FLAP & 40
& \sed\GetGen{\QOFFLAPForty@Gen}{1}
& \sed\GetGen{\QOFFLAPForty@Gen}{2}
& \fst\GetGen{\QOFFLAPForty@Gen}{3}
& \sed\GetGen{\QOFFLAPForty@Gen}{4}
& \sed\GetGen{\QOFFLAPForty@Gen}{5}
& \sed\GENMean{\QOFFLAPForty@Gen} \\

& T{\'y}r & 40
& \thd\GetGen{\QOFTyrForty@Gen}{1}
& \GetGen{\QOFTyrForty@Gen}{2}
& \sed\GetGen{\QOFTyrForty@Gen}{3}
& \GetGen{\QOFTyrForty@Gen}{4}
& \thd\GetGen{\QOFTyrForty@Gen}{5}
& \thd\GENMean{\QOFTyrForty@Gen} \\

& \textbf{\paperN(Ours)} & 40 
& \fst\GetGen{\QOFOursForty@Gen}{1}
& \fst\GetGen{\QOFOursForty@Gen}{2}
& \thd\GetGen{\QOFOursForty@Gen}{3}
& \fst\GetGen{\QOFOursForty@Gen}{4}
& \fst\GetGen{\QOFOursForty@Gen}{5}
& \fst\GENMean{\QOFOursForty@Gen} \\

\midrule

\multirow{7}{*}{\shortstack[l]{Qwen2\\57B-A14B}}

& Dense & -
& \GetGen{\QWTFDense@Gen}{1}
& \GetGen{\QWTFDense@Gen}{2}
& \GetGen{\QWTFDense@Gen}{3}
& \GetGen{\QWTFDense@Gen}{4}
& \GetGen{\QWTFDense@Gen}{5}
& \GENMean{\QWTFDense@Gen} \\

\cline{2-9}

& Wanda-sp & 20
& \thd\GetGen{\QWTFWandaTwenty@Gen}{1}
& \GetGen{\QWTFWandaTwenty@Gen}{2}
& \thd\GetGen{\QWTFWandaTwenty@Gen}{3}
& \GetGen{\QWTFWandaTwenty@Gen}{4}
& \thd\GetGen{\QWTFWandaTwenty@Gen}{5}
& \thd\GENMean{\QWTFWandaTwenty@Gen} \\

& FLAP & 20
& \sed\GetGen{\QWTFFLAPTwenty@Gen}{1}
& \sed\GetGen{\QWTFFLAPTwenty@Gen}{2}
& \fst\GetGen{\QWTFFLAPTwenty@Gen}{3}
& \sed\GetGen{\QWTFFLAPTwenty@Gen}{4}
& \sed\GetGen{\QWTFFLAPTwenty@Gen}{5}
& \sed\GENMean{\QWTFFLAPTwenty@Gen} \\

& \textbf{\paperN(Ours)} & 20 
& \fst\GetGen{\QWTFOursTwenty@Gen}{1}
& \fst\GetGen{\QWTFOursTwenty@Gen}{2}
& \sed\GetGen{\QWTFOursTwenty@Gen}{3}
& \fst\GetGen{\QWTFOursTwenty@Gen}{4}
& \fst\GetGen{\QWTFOursTwenty@Gen}{5}
& \fst\GENMean{\QWTFOursTwenty@Gen} \\

\cline{2-9}

& Wanda-sp & 40
& \thd\GetGen{\QWTFWandaForty@Gen}{1}
& \GetGen{\QWTFWandaForty@Gen}{2}
& \thd\GetGen{\QWTFWandaForty@Gen}{3}
& \GetGen{\QWTFWandaForty@Gen}{4}
& \thd\GetGen{\QWTFWandaForty@Gen}{5}
& \thd\GENMean{\QWTFWandaForty@Gen} \\

& FLAP & 40
& \sed\GetGen{\QWTFFLAPForty@Gen}{1}
& \sed\GetGen{\QWTFFLAPForty@Gen}{2}
& \fst\GetGen{\QWTFFLAPForty@Gen}{3}
& \sed\GetGen{\QWTFFLAPForty@Gen}{4}
& \sed\GetGen{\QWTFFLAPForty@Gen}{5}
& \sed\GENMean{\QWTFFLAPForty@Gen} \\

& \textbf{\paperN(Ours)} & 40 
& \fst\GetGen{\QWTFOursForty@Gen}{1}
& \fst\GetGen{\QWTFOursForty@Gen}{2}
& \sed\GetGen{\QWTFOursForty@Gen}{3}
& \fst\GetGen{\QWTFOursForty@Gen}{4}
& \fst\GetGen{\QWTFOursForty@Gen}{5}
& \fst\GENMean{\QWTFOursForty@Gen} \\

\midrule

\multirow{11}{*}{\shortstack[l]{Qwen3\\30B-A3B}}

& Dense & -
& \GetGen{\QTTZDense@Gen}{1}
& \GetGen{\QTTZDense@Gen}{2}
& \GetGen{\QTTZDense@Gen}{3}
& \GetGen{\QTTZDense@Gen}{4}
& \GetGen{\QTTZDense@Gen}{5}
& \GENMean{\QTTZDense@Gen} \\

\cline{2-9}

& Wanda-sp & 20
& \thd\GetGen{\QTTZWandaTwenty@Gen}{1}
& \thd\GetGen{\QTTZWandaTwenty@Gen}{2}
& \thd\GetGen{\QTTZWandaTwenty@Gen}{3}
& \thd\GetGen{\QTTZWandaTwenty@Gen}{4}
& \thd\GetGen{\QTTZWandaTwenty@Gen}{5}
& \thd\GENMean{\QTTZWandaTwenty@Gen} \\

& FLAP & 20
& \GetGen{\QTTZFLAPTwenty@Gen}{1}
& \GetGen{\QTTZFLAPTwenty@Gen}{2}
& \GetGen{\QTTZFLAPTwenty@Gen}{3}
& \GetGen{\QTTZFLAPTwenty@Gen}{4}
& \GetGen{\QTTZFLAPTwenty@Gen}{5}
& \GENMean{\QTTZFLAPTwenty@Gen} \\

& T{\'y}r & 20
& \sed\GetGen{\QTTZTyrTwenty@Gen}{1}
& \sed\GetGen{\QTTZTyrTwenty@Gen}{2}
& \sed\GetGen{\QTTZTyrTwenty@Gen}{3}
& \sed\GetGen{\QTTZTyrTwenty@Gen}{4}
& \sed\GetGen{\QTTZTyrTwenty@Gen}{5}
& \sed\GENMean{\QTTZTyrTwenty@Gen} \\

& \textbf{\paperN(Ours)} & 20 
& \fst\GetGen{\QTTZOursTwenty@Gen}{1}
& \fst\GetGen{\QTTZOursTwenty@Gen}{2}
& \fst\GetGen{\QTTZOursTwenty@Gen}{3}
& \fst\GetGen{\QTTZOursTwenty@Gen}{4}
& \fst\GetGen{\QTTZOursTwenty@Gen}{5}
& \fst\GENMean{\QTTZOursTwenty@Gen} \\

\cline{2-9}

& Wanda-sp & 40
& \thd\GetGen{\QTTZWandaForty@Gen}{1}
& \GetGen{\QTTZWandaForty@Gen}{2}
& \thd\GetGen{\QTTZWandaForty@Gen}{3}
& \thd\GetGen{\QTTZWandaForty@Gen}{4}
& \thd\GetGen{\QTTZWandaForty@Gen}{5}
& \thd\GENMean{\QTTZWandaForty@Gen} \\

& FLAP & 40
& \GetGen{\QTTZFLAPForty@Gen}{1}
& \GetGen{\QTTZFLAPForty@Gen}{2}
& \GetGen{\QTTZFLAPForty@Gen}{3}
& \GetGen{\QTTZFLAPForty@Gen}{4}
& \GetGen{\QTTZFLAPForty@Gen}{5}
& \GENMean{\QTTZFLAPForty@Gen} \\

& T{\'y}r & 40
& \sed\GetGen{\QTTZTyrForty@Gen}{1}
& \sed\GetGen{\QTTZTyrForty@Gen}{2}
& \fst\GetGen{\QTTZTyrForty@Gen}{3}
& \sed\GetGen{\QTTZTyrForty@Gen}{4}
& \sed\GetGen{\QTTZTyrForty@Gen}{5}
& \sed\GENMean{\QTTZTyrForty@Gen} \\

& \textbf{\paperN(Ours)} & 40 
& \fst\GetGen{\QTTZOursForty@Gen}{1}
& \fst\GetGen{\QTTZOursForty@Gen}{2}
& \sed\GetGen{\QTTZOursForty@Gen}{3}
& \fst\GetGen{\QTTZOursForty@Gen}{4}
& \fst\GetGen{\QTTZOursForty@Gen}{5}
& \fst\GENMean{\QTTZOursForty@Gen} \\

\bottomrule
\end{tabular}
\end{table}

\def\QTVDense@Gen{88.38,52.21,62.43,85.48,621.79}
\def\QTVWandaTwenty@Gen{7.54,1.22,1.89,28.46,5.71}
\def\QTVFLAPTwenty@Gen{86.79,32.37,51.71,50.53,343.93}
\def\QTVOursTwenty@Gen{88.67,45.68,60.73,82.76,611.79}

\def\QTVWandaForty@Gen{0.00,0.01,0.04,0.00,0.71}
\def\QTVFLAPForty@Gen{0.00,0.00,0.01,0.05,0.00}
\def\QTVOursForty@Gen{87.99,32.06,58.13,73.14,569.64}

\def\QTFVDense@Gen{87.59,43.88,60.86,88.75,625.36}
\def\QTFVWandaTwenty@Gen{1.93,5.75,6.80,3.51,15.71}
\def\QTFVFLAPTwenty@Gen{68.42,37.27,37.16,59.63,285.00}
\def\QTFVOursTwenty@Gen{87.22,42.65,60.21,85.15,617.50}

\def\QTFVWandaForty@Gen{0.00,0.00,0.00,0.00,0.00}
\def\QTFVFLAPForty@Gen{0.00,0.00,0.00,0.00,0.00}
\def\QTFVOursForty@Gen{86.73,34.86,57.47,74.89,537.86}

\begin{table}[t]
\centering
\caption{Detailed Accuracies (\%) of Qwen2-VL \cite{Qwen2-VL} and Qwen2.5-VL \cite{Qwen2.5-VL} for 5 Visual Question Answering (VQA) tasks with different levels of structured sparsity. We used macro average as a mean value.}
\label{tab:vlm-QA}
\setlength{\tabcolsep}{4.2pt}
\renewcommand{\arraystretch}{1.1}
\footnotesize
\begin{tabular}{l|l c| ccccc c}
\toprule
Model & Method & PR(\%) & POPE & OK-VQA & GQA & ScienceQA & MME & Mean \\
\midrule

\multirow{7}{*}{Qwen2-VL}
& Dense & - 
& \GetGen{\QTVDense@Gen}{1}
& \GetGen{\QTVDense@Gen}{2}
& \GetGen{\QTVDense@Gen}{3}
& \GetGen{\QTVDense@Gen}{4}
& \GetGen{\QTVDense@Gen}{5}
& 62.14 \\

\cline{2-9}
& Wanda-sp & 20 
& \thd\GetGen{\QTVWandaTwenty@Gen}{1}
& \thd\GetGen{\QTVWandaTwenty@Gen}{2}
& \thd\GetGen{\QTVWandaTwenty@Gen}{3}
& \thd\GetGen{\QTVWandaTwenty@Gen}{4}
& \thd\GetGen{\QTVWandaTwenty@Gen}{5}
& \thd 7.86 \\

& FLAP & 20 
& \sed\GetGen{\QTVFLAPTwenty@Gen}{1}
& \sed\GetGen{\QTVFLAPTwenty@Gen}{2}
& \sed\GetGen{\QTVFLAPTwenty@Gen}{3}
& \sed\GetGen{\QTVFLAPTwenty@Gen}{4}
& \sed\GetGen{\QTVFLAPTwenty@Gen}{5}
& \sed 46.74 \\

& \textbf{\paperN (Ours)} & 20
& \fst\GetGen{\QTVOursTwenty@Gen}{1}
& \fst\GetGen{\QTVOursTwenty@Gen}{2}
& \fst\GetGen{\QTVOursTwenty@Gen}{3}
& \fst\GetGen{\QTVOursTwenty@Gen}{4}
& \fst\GetGen{\QTVOursTwenty@Gen}{5}
& \fst 59.94 \\

\cline{2-9}

& Wanda-sp & 40 
& \GetGen{\QTVWandaForty@Gen}{1}
& \sed\GetGen{\QTVWandaForty@Gen}{2}
& \sed\GetGen{\QTVWandaForty@Gen}{3}
& \GetGen{\QTVWandaForty@Gen}{4}
& \sed\GetGen{\QTVWandaForty@Gen}{5}
& \sed 0.02 \\

& FLAP & 40 
& \GetGen{\QTVFLAPForty@Gen}{1}
& \GetGen{\QTVFLAPForty@Gen}{2}
& \thd\GetGen{\QTVFLAPForty@Gen}{3}
& \sed\GetGen{\QTVFLAPForty@Gen}{4}
& \GetGen{\QTVFLAPForty@Gen}{5}
& \thd 0.01 \\

& \textbf{\paperN (Ours)} & 40 
& \fst\GetGen{\QTVOursForty@Gen}{1}
& \fst\GetGen{\QTVOursForty@Gen}{2}
& \fst\GetGen{\QTVOursForty@Gen}{3}
& \fst\GetGen{\QTVOursForty@Gen}{4}
& \fst\GetGen{\QTVOursForty@Gen}{5}
& \fst 54.33 \\

\midrule

\multirow{7}{*}{Qwen2.5-VL}

& Dense & - 
& \GetGen{\QTFVDense@Gen}{1}
& \GetGen{\QTFVDense@Gen}{2}
& \GetGen{\QTFVDense@Gen}{3}
& \GetGen{\QTFVDense@Gen}{4}
& \GetGen{\QTFVDense@Gen}{5}
& 60.68 \\

\cline{2-9}
& Wanda-sp & 20 
& \thd\GetGen{\QTFVWandaTwenty@Gen}{1}
& \thd\GetGen{\QTFVWandaTwenty@Gen}{2}
& \thd\GetGen{\QTFVWandaTwenty@Gen}{3}
& \thd\GetGen{\QTFVWandaTwenty@Gen}{4}
& \thd\GetGen{\QTFVWandaTwenty@Gen}{5}
& \thd 3.71 \\

& FLAP & 20 
& \sed\GetGen{\QTFVFLAPTwenty@Gen}{1}
& \sed\GetGen{\QTFVFLAPTwenty@Gen}{2}
& \sed\GetGen{\QTFVFLAPTwenty@Gen}{3}
& \sed\GetGen{\QTFVFLAPTwenty@Gen}{4}
& \sed\GetGen{\QTFVFLAPTwenty@Gen}{5}
& \sed 42.53 \\

& \textbf{\paperN (Ours)} & 20
& \fst\GetGen{\QTFVOursTwenty@Gen}{1}
& \fst\GetGen{\QTFVOursTwenty@Gen}{2}
& \fst\GetGen{\QTFVOursTwenty@Gen}{3}
& \fst\GetGen{\QTFVOursTwenty@Gen}{4}
& \fst\GetGen{\QTFVOursTwenty@Gen}{5}
& \fst 59.46 \\

\cline{2-9}

& Wanda-sp & 40 
& \GetGen{\QTFVWandaForty@Gen}{1}
& \GetGen{\QTFVWandaForty@Gen}{2}
& \GetGen{\QTFVWandaForty@Gen}{3}
& \GetGen{\QTFVWandaForty@Gen}{4}
& \GetGen{\QTFVWandaForty@Gen}{5}
&  0.00 \\

& FLAP & 40 
& \GetGen{\QTFVFLAPForty@Gen}{1}
& \GetGen{\QTFVFLAPForty@Gen}{2}
& \GetGen{\QTFVFLAPForty@Gen}{3}
& \GetGen{\QTFVFLAPForty@Gen}{4}
& \GetGen{\QTFVFLAPForty@Gen}{5}
& 0.00 \\

& \textbf{\paperN (Ours)} & 40 
& \fst\GetGen{\QTFVOursForty@Gen}{1}
& \fst\GetGen{\QTFVOursForty@Gen}{2}
& \fst\GetGen{\QTFVOursForty@Gen}{3}
& \fst\GetGen{\QTFVOursForty@Gen}{4}
& \fst\GetGen{\QTFVOursForty@Gen}{5}
& \fst 54.63 \\


\bottomrule
\end{tabular}
\end{table}


\def\LTOEGENPPTwenty{78.06,32.4,3.60,21.95,37.45}
\def\LTOEGENOursTwenty{79.70,39.20,3.82,23.78,40.33}
\def\LTOEGENTopkTwenty{78.10,43.2,5.60,24.39,43.44}

\def\LTOSZGENPPTwenty{75.83,15.20,7.87,6.71,11.37}
\def\LTOSZGENOursTwenty{76.46,16.20,9.78,7.93,10.24}
\def\LTOSZGENTopkTwenty{76.99,19.80,14.57,9.76,11.68}

\begin{table}[t]
\centering
\caption{Accuracies (\%) of  Llama-2~\cite{touvron2023llama} and Llama-3 \cite{dubey2024llama} for 5 Generative tasks with different pruning strategies.}
\label{tab:strategy}

\setlength{\tabcolsep}{7.5pt}
\renewcommand{\arraystretch}{1.1}
\footnotesize

\begin{tabular}{l|lc|l*{6}{c}}
\toprule
Params & Method & PR
& coqa & mbpp & nq\_open & humaneval & gsm8k
& Mean \\
\midrule

\multirow{3}{*}{Llama3.1-8B}

& Variant (1)  & 20
& \thd\GetGen{\LTOEGENPPTwenty}{1}
& \thd\GetGen{\LTOEGENPPTwenty}{2}
& \thd\GetGen{\LTOEGENPPTwenty}{3}
& \thd\GetGen{\LTOEGENPPTwenty}{4}
& \thd\GetGen{\LTOEGENPPTwenty}{5}
& \thd\GENMean{\LTOEGENPPTwenty} \\

& \textbf{\paperN(Ours)} & 20
& \fst\GetGen{\LTOEGENOursTwenty}{1}
& \sed\GetGen{\LTOEGENOursTwenty}{2}
& \sed\GetGen{\LTOEGENOursTwenty}{3}
& \sed\GetGen{\LTOEGENOursTwenty}{4}
& \sed\GetGen{\LTOEGENOursTwenty}{5}
& \sed\GENMean{\LTOEGENOursTwenty} \\

& Variant (2) & 20
& \sed\GetGen{\LTOEGENTopkTwenty}{1}
& \fst\GetGen{\LTOEGENTopkTwenty}{2}
& \fst\GetGen{\LTOEGENTopkTwenty}{3}
& \fst\GetGen{\LTOEGENTopkTwenty}{4}
& \fst\GetGen{\LTOEGENTopkTwenty}{5}
& \fst\GENMean{\LTOEGENTopkTwenty} \\
\midrule

\multirow{3}{*}{Llama2-7B}

& Variant (1)  & 20
& \thd\GetGen{\LTOSZGENPPTwenty}{1}
& \thd\GetGen{\LTOSZGENPPTwenty}{2}
& \thd\GetGen{\LTOSZGENPPTwenty}{3}
& \thd\GetGen{\LTOSZGENPPTwenty}{4}
& \sed\GetGen{\LTOSZGENPPTwenty}{5}
& \thd\GENMean{\LTOSZGENPPTwenty} \\

& \textbf{\paperN(Ours)} & 20
& \sed\GetGen{\LTOSZGENOursTwenty}{1}
& \sed\GetGen{\LTOSZGENOursTwenty}{2}
& \sed\GetGen{\LTOSZGENOursTwenty}{3}
& \sed\GetGen{\LTOSZGENOursTwenty}{4}
& \thd\GetGen{\LTOSZGENOursTwenty}{5}
& \sed\GENMean{\LTOSZGENOursTwenty} \\

& Variant (2) & 20
& \fst\GetGen{\LTOSZGENTopkTwenty}{1}
& \fst\GetGen{\LTOSZGENTopkTwenty}{2}
& \fst\GetGen{\LTOSZGENTopkTwenty}{3}
& \fst\GetGen{\LTOSZGENTopkTwenty}{4}
& \fst\GetGen{\LTOSZGENTopkTwenty}{5}
& \fst\GENMean{\LTOSZGENTopkTwenty} \\

\bottomrule
\end{tabular}
\end{table}

\def\LlamaTwoSevenQAZeroFive{71.84,63.18,52.17,66.46,70.03,38.14,28.60}
\def\LlamaTwoSevenQAOne{72.63,63.54,53.42,67.72,71.13,39.59,29.40}
\def\LlamaTwoSevenQATwo{72.29,64.26,55.06,67.80,72.31,40.02,28.40}

\def\LlamaThreeEightQAZeroFive{76.42,66.79,53.14,70.09,75.13,43.09,29.00}
\def\LlamaThreeEightQAOne{77.28,63.90,54.68,70.88,76.22,44.20,30.20}
\def\LlamaThreeEightQATwo{79.14,67.15,57.05,71.43,78.87,49.23,33.20}

\def\QwenThreeEightQAZeroFive{84.34,74.01,50.13,64.33,76.31,47.10,30.20}
\def\QwenThreeEightQAOne{84.77,74.37,51.76,64.80,77.78,48.21,28.80}
\def\QwenThreeEightQATwo{85.41,74.37,54.89,65.19,78.83,52.05,30.60}

\begin{table}[t]
\centering
\caption{
Ablation study on the effect of partition fraction in \paperN on QA tasks.
Accuracies (\%) are reported under a fixed pruning ratio.
}
\label{tab:partition-frac-qa}

\setlength{\tabcolsep}{6pt}
\renewcommand{\arraystretch}{1.1}
\footnotesize

\begin{tabular}{l|c|*{8}{c}}
\toprule
Params & PF 
& BoolQ & RTE & HS & WG & ARC-e & ARC-c & OBQA
& Mean \\
\midrule

\multirow{3}{*}{Llama2-7B}

& 0.05
& \thd\GetQA{\LlamaTwoSevenQAZeroFive}{1}
& \thd\GetQA{\LlamaTwoSevenQAZeroFive}{2}
& \thd\GetQA{\LlamaTwoSevenQAZeroFive}{3}
& \thd\GetQA{\LlamaTwoSevenQAZeroFive}{4}
& \thd\GetQA{\LlamaTwoSevenQAZeroFive}{5}
& \thd\GetQA{\LlamaTwoSevenQAZeroFive}{6}
& \sed\GetQA{\LlamaTwoSevenQAZeroFive}{7}
& \thd\QAMean{\LlamaTwoSevenQAZeroFive} \\

& 0.10
& \fst\GetQA{\LlamaTwoSevenQAOne}{1}
& \sed\GetQA{\LlamaTwoSevenQAOne}{2}
& \sed\GetQA{\LlamaTwoSevenQAOne}{3}
& \sed\GetQA{\LlamaTwoSevenQAOne}{4}
& \sed\GetQA{\LlamaTwoSevenQAOne}{5}
& \sed\GetQA{\LlamaTwoSevenQAOne}{6}
& \fst\GetQA{\LlamaTwoSevenQAOne}{7}
& \sed\QAMean{\LlamaTwoSevenQAOne} \\

& 0.20
& \sed\GetQA{\LlamaTwoSevenQATwo}{1}
& \fst\GetQA{\LlamaTwoSevenQATwo}{2}
& \fst\GetQA{\LlamaTwoSevenQATwo}{3}
& \fst\GetQA{\LlamaTwoSevenQATwo}{4}
& \fst\GetQA{\LlamaTwoSevenQATwo}{5}
& \fst\GetQA{\LlamaTwoSevenQATwo}{6}
& \thd\GetQA{\LlamaTwoSevenQATwo}{7}
& \fst\QAMean{\LlamaTwoSevenQATwo} \\

\midrule

\multirow{3}{*}{Llama3.1-8B}

& 0.05
& \thd\GetQA{\LlamaThreeEightQAZeroFive}{1}
& \sed\GetQA{\LlamaThreeEightQAZeroFive}{2}
& \thd\GetQA{\LlamaThreeEightQAZeroFive}{3}
& \thd\GetQA{\LlamaThreeEightQAZeroFive}{4}
& \thd\GetQA{\LlamaThreeEightQAZeroFive}{5}
& \thd\GetQA{\LlamaThreeEightQAZeroFive}{6}
& \thd\GetQA{\LlamaThreeEightQAZeroFive}{7}
& \thd\QAMean{\LlamaThreeEightQAZeroFive} \\

& 0.10
& \sed\GetQA{\LlamaThreeEightQAOne}{1}
& \thd\GetQA{\LlamaThreeEightQAOne}{2}
& \sed\GetQA{\LlamaThreeEightQAOne}{3}
& \sed\GetQA{\LlamaThreeEightQAOne}{4}
& \sed\GetQA{\LlamaThreeEightQAOne}{5}
& \sed\GetQA{\LlamaThreeEightQAOne}{6}
& \sed\GetQA{\LlamaThreeEightQAOne}{7}
& \sed\QAMean{\LlamaThreeEightQAOne} \\

& 0.20
& \fst\GetQA{\LlamaThreeEightQATwo}{1}
& \fst\GetQA{\LlamaThreeEightQATwo}{2}
& \fst\GetQA{\LlamaThreeEightQATwo}{3}
& \fst\GetQA{\LlamaThreeEightQATwo}{4}
& \fst\GetQA{\LlamaThreeEightQATwo}{5}
& \fst\GetQA{\LlamaThreeEightQATwo}{6}
& \fst\GetQA{\LlamaThreeEightQATwo}{7}
& \fst\QAMean{\LlamaThreeEightQATwo} \\
\midrule

\multirow{3}{*}{Qwen3-8B}

& 0.05
& \thd\GetQA{\QwenThreeEightQAZeroFive}{1}
& \sed\GetQA{\QwenThreeEightQAZeroFive}{2}
& \sed\GetQA{\QwenThreeEightQAZeroFive}{3}
& \sed\GetQA{\QwenThreeEightQAZeroFive}{4}
& \thd\GetQA{\QwenThreeEightQAZeroFive}{5}
& \thd\GetQA{\QwenThreeEightQAZeroFive}{6}
& \sed\GetQA{\QwenThreeEightQAZeroFive}{7}
& \thd\QAMean{\QwenThreeEightQAZeroFive} \\

& 0.10
& \sed\GetQA{\QwenThreeEightQAOne}{1}
& \fst\GetQA{\QwenThreeEightQAOne}{2}
& \fst\GetQA{\QwenThreeEightQAOne}{3}
& \fst\GetQA{\QwenThreeEightQAOne}{4}
& \sed\GetQA{\QwenThreeEightQAOne}{5}
& \sed\GetQA{\QwenThreeEightQAOne}{6}
& \thd\GetQA{\QwenThreeEightQAOne}{7}
& \sed\QAMean{\QwenThreeEightQAOne} \\

& 0.20
& \fst\GetQA{\QwenThreeEightQATwo}{1}
& \fst\GetQA{\QwenThreeEightQATwo}{2}
& \fst\GetQA{\QwenThreeEightQAOne}{3}
& \fst\GetQA{\QwenThreeEightQAOne}{4}
& \fst\GetQA{\QwenThreeEightQATwo}{5}
& \fst\GetQA{\QwenThreeEightQATwo}{6}
& \fst\GetQA{\QwenThreeEightQATwo}{7}
& \fst\QAMean{\QwenThreeEightQATwo} \\

\bottomrule
\end{tabular}
\end{table}

\def\LlamaTwoSevenGENZeroFive{77.14,15.00,9.00,8.54,11.22}
\def\LlamaTwoSevenGENOne{76.46,16.20,9.78,7.93,10.24}
\def\LlamaTwoSevenGENTwo{76.65,18.20,12.44,9.15,10.77}

\def\LlamaThreeEightGENZeroFive{78.47,29.00,35.40,3.42,25.00,38.59}
\def\LlamaThreeEightGENOne{79.70,39.20,3.82,23.78,40.33}
\def\LlamaThreeEightGENTwo{78.63,43.40,6.59,28.05,42.46}

\def\QwenThreeEightGENZeroFive{80.73,52.20,1.19,53.66,85.29}
\def\QwenThreeEightGENOne{81.05,56.00,2.35,56.10,85.14}
\def\QwenThreeEightGENTwo{80.01,61.80,4.32,57.32,84.91}

\begin{table}[t]
\centering
\caption{
Ablation study on the effect of partition fraction in \paperN on generative tasks.
Accuracies (\%) are reported under a fixed pruning ratio.
}
\label{tab:partition-frac-gen}

\setlength{\tabcolsep}{7pt}
\renewcommand{\arraystretch}{1.1}
\footnotesize

\begin{tabular}{l|c|*{6}{c}}
\toprule
Params & PF 
& CoQA & MBPP & NQ-open & HumanEval & GSM8K
& Mean \\
\midrule

\multirow{3}{*}{Llama2-7B}

& 0.05
& \fst\GetGen{\LlamaTwoSevenGENZeroFive}{1}
& \thd\GetGen{\LlamaTwoSevenGENZeroFive}{2}
& \thd\GetGen{\LlamaTwoSevenGENZeroFive}{3}
& \sed\GetGen{\LlamaTwoSevenGENZeroFive}{4}
& \fst\GetGen{\LlamaTwoSevenGENZeroFive}{5}
& \sed\GENMean{\LlamaTwoSevenGENZeroFive} \\

& 0.10
& \thd\GetGen{\LlamaTwoSevenGENOne}{1}
& \sed\GetGen{\LlamaTwoSevenGENOne}{2}
& \sed\GetGen{\LlamaTwoSevenGENOne}{3}
& \thd\GetGen{\LlamaTwoSevenGENOne}{4}
& \thd\GetGen{\LlamaTwoSevenGENOne}{5}
& \thd\GENMean{\LlamaTwoSevenGENOne} \\

& 0.20
& \sed\GetGen{\LlamaTwoSevenGENTwo}{1}
& \fst\GetGen{\LlamaTwoSevenGENTwo}{2}
& \fst\GetGen{\LlamaTwoSevenGENTwo}{3}
& \fst\GetGen{\LlamaTwoSevenGENTwo}{4}
& \sed\GetGen{\LlamaTwoSevenGENTwo}{5}
& \fst\GENMean{\LlamaTwoSevenGENTwo} \\

\midrule

\multirow{3}{*}{Llama3.1-8B}

& 0.05
& \thd\GetGen{\LlamaThreeEightGENZeroFive}{1}
& \thd\GetGen{\LlamaThreeEightGENZeroFive}{2}
& \fst\GetGen{\LlamaThreeEightGENZeroFive}{3}
& \thd\GetGen{\LlamaThreeEightGENZeroFive}{4}
& \thd\GetGen{\LlamaThreeEightGENZeroFive}{5}
& \thd\GENMean{\LlamaThreeEightGENZeroFive} \\

& 0.10
& \fst\GetGen{\LlamaThreeEightGENOne}{1}
& \sed\GetGen{\LlamaThreeEightGENOne}{2}
& \thd\GetGen{\LlamaThreeEightGENOne}{3}
& \sed\GetGen{\LlamaThreeEightGENOne}{4}
& \sed\GetGen{\LlamaThreeEightGENOne}{5}
& \sed\GENMean{\LlamaThreeEightGENOne} \\

& 0.20
& \sed\GetGen{\LlamaThreeEightGENTwo}{1}
& \fst\GetGen{\LlamaThreeEightGENTwo}{2}
& \sed\GetGen{\LlamaThreeEightGENTwo}{3}
& \fst\GetGen{\LlamaThreeEightGENTwo}{4}
& \fst\GetGen{\LlamaThreeEightGENTwo}{5}
& \fst\GENMean{\LlamaThreeEightGENTwo} \\

\midrule

\multirow{3}{*}{Qwen3-8B}

& 0.05
& \sed\GetGen{\QwenThreeEightGENZeroFive}{1}
& \thd\GetGen{\QwenThreeEightGENZeroFive}{2}
& \thd\GetGen{\QwenThreeEightGENZeroFive}{3}
& \thd\GetGen{\QwenThreeEightGENZeroFive}{4}
& \fst\GetGen{\QwenThreeEightGENZeroFive}{5}
& \thd\GENMean{\QwenThreeEightGENZeroFive} \\

& 0.10
& \fst\GetGen{\QwenThreeEightGENOne}{1}
& \sed\GetGen{\QwenThreeEightGENOne}{2}
& \sed\GetGen{\QwenThreeEightGENOne}{3}
& \sed\GetGen{\QwenThreeEightGENOne}{4}
& \sed\GetGen{\QwenThreeEightGENOne}{5}
& \sed\GENMean{\QwenThreeEightGENOne} \\

& 0.20
& \thd\GetGen{\QwenThreeEightGENTwo}{1}
& \fst\GetGen{\QwenThreeEightGENTwo}{2}
& \fst\GetGen{\QwenThreeEightGENTwo}{3}
& \fst\GetGen{\QwenThreeEightGENTwo}{4}
& \thd\GetGen{\QwenThreeEightGENTwo}{5}
& \fst\GENMean{\QwenThreeEightGENTwo} \\

\bottomrule
\end{tabular}
\end{table}


\end{document}